\documentclass[11pt]{article}
\usepackage{latexsym,amsfonts,amsmath,epsfig,amssymb}
\usepackage[colorlinks,linkcolor=red,anchorcolor=blue,urlcolor=red,citecolor=blue]{hyperref}
\usepackage{graphicx,times}
\usepackage{mathrsfs}
\usepackage{caption}
\usepackage{enumerate}
\usepackage{subfigure}
\usepackage{graphicx}
\usepackage{epstopdf}
\usepackage{multirow}
\usepackage{booktabs}
\usepackage{cite}
\usepackage{color}
\usepackage{algorithm}
\usepackage{algorithmicx}
\usepackage{algpseudocode}
\usepackage[numbers,sort&compress]{natbib}
\usepackage{mathtools}
\usepackage{array}
\usepackage{booktabs} 
\usepackage{cases}
\graphicspath{{figure/}}

\usepackage{array}
\newcommand{\PreserveBackslash}[1]{\let\temp=\\#1\let\\=\temp}
\newcolumntype{C}[1]{>{\PreserveBackslash\centering}p{#1}}
\newcolumntype{R}[1]{>{\PreserveBackslash\raggedleft}p{#1}}
\newcolumntype{L}[1]{>{\PreserveBackslash\raggedright}p{#1}}

\newcommand{\qed}{\hfill $\Box$}

\newtheorem{theorem}{Theorem}

\newtheorem{Prop}{Proposition}

\newtheorem{lemma}{Lemma}

\newtheorem{assumption}{Assumption}
\newtheorem{remark}{Remark}

\newtheorem{definition}{Definition}

\newcounter{nextauthor}
\def\mathrm{\mbox}

\begin{document}

\title{\Large {\bf Low-Rank Tensor Learning by Generalized Nonconvex Regularization}}

\author{Sijia Xia\footnotemark[1], \  Michael K. Ng\footnotemark[2],   \ and \
Xiongjun Zhang\footnotemark[3]
}

\renewcommand{\thefootnote}{\fnsymbol{footnote}}
\footnotetext[1]{School of Mathematics and Statistics,
Central China Normal University, Wuhan 430079, China (e-mail: sijiax@mails.ccnu.edu.cn).
}
\footnotetext[2]{Department of Mathematics, Hong Kong Baptist University,
	Kowloon Tong, Hong Kong (e-mail: michael-ng@hkbu.edu.hk).
	The research of this author was supported
	in part by the
	Hong Kong Research Grant Council GRF 
	17300021, C7004-21GF and Joint NSFC-RGC N-HKU76921.}
\footnotetext[3]{School of Mathematics and Statistics, and Key Laboratory of Nonlinear Analysis \& Applications (Ministry of Education), Central China Normal University, Wuhan 430079, China (e-mail: xjzhang@ccnu.edu.cn).
The research of this author was supported
in part by the
National Natural Science Foundation of China under Grant No. 12171189
 and the Fundamental
Research Funds for the Central Universities under Grant No. CCNU24AI002.
}

\renewcommand{\thefootnote}{\fnsymbol{footnote}}

\renewcommand{\thefootnote}{\arabic{footnote}}

\maketitle \vspace*{0mm}
\begin{center}
\begin{minipage}{5.5in}
$${\bf Abstract}$$

In this paper, we study the problem of low-rank tensor learning, where only a few of training samples are observed and the  underlying tensor has a low-rank structure.
The existing methods are based on the sum of nuclear norms of unfolding matrices of a tensor, which may be suboptimal.
In order to explore the low-rankness of the underlying tensor effectively, we propose a nonconvex model 
based on transformed tensor nuclear norm for low-rank tensor learning.
Specifically, a family of nonconvex functions are employed onto the singular values of all frontal slices of a tensor in the transformed domain to characterize the low-rankness of the underlying tensor.
An error bound between the stationary point of the nonconvex model and the underlying tensor is established under restricted strong convexity on the loss function (such as least squares loss and logistic regression) 
and suitable regularity conditions on the nonconvex penalty function.
By reformulating the nonconvex function into the difference of two convex functions,
a proximal majorization-minimization (PMM) algorithm is designed to solve the resulting model.
Then the global convergence and convergence rate  of PMM are established under very mild conditions.
Numerical experiments are conducted on tensor completion and binary classification to demonstrate the effectiveness of the proposed method over other state-of-the-art methods.
\end{minipage}
\end{center}

\begin{center}
\begin{minipage}{5.5in}

\end{minipage}
\end{center}


\section{Introduction}

Tensors, which are higher-order generalization of vectors and  matrices, have attracted much attention in the past decades and have a broad range of applications in various fields
 such as image processing \cite{Liu2013}, computer vision \cite{panagakis2021tensor}, machine learning \cite{wang2008tensor, kolda2009tensor}, and bioinformation \cite{broadbent2024deciphering}. These tensor data are generally lied in  low-dimensional subspaces in realistic scenes, which substantially  have  low-rank structure.
In turn, the low-rank tensor data lead  to efficient estimation and prediction.
 As a result, the low-rank tensor based  approaches are utilized to exploit the internal structure of a tensor efficiently.
 In this paper, we focus on the  problem of
low-rank tensor learning,
where only a few of samples are given to learn  the underlying tensor. Specifically,
the general model of low-rank tensor learning is formulated as follows:
\begin{equation}\label{genMod}
\min_{\mathcal X\in D}  f_{n,\mathcal{Y}} (\mathcal X)+\beta\cdot\text{rank}(\mathcal X),
\end{equation}
where $f_{n,\mathcal{Y}}(\cdot)$ is the loss function related to the number of samples $n$ and the observed tensor $\mathcal{Y}$, $D$ is a given constraint set,
$\beta>0$ is the regularization parameter,
and $\text{rank}(\mathcal X)$ denotes the rank function of the underlying parameter tensor $\mathcal X\in\mathbb{R}^{n_1\times n_2\times n_3}$.

Different tasks result in different loss functions in (\ref{genMod}).
Under the least squares loss, the model in (\ref{genMod}) is suitable for low-rank tensor completion with noises.
For example, Gandy et al. \cite{Gandy_2011} proposed a model composed of least squares loss and nuclear norms  of unfolding matrices of a tensor for tensor completion, where the nuclear norms  of unfolding matrices were utilized to approximate the Tucker rank.
Besides,  Qiu et al. \cite{qiu2022noisy} and 
Zhang et al. \cite{zhang2019corrected} also employed the least squares loss with different low-rank approximations for noisy tensor completion, respectively.
Under logistic regression loss, 
the model in (\ref{genMod}) can be used for binary classification, where the low-rankness with respect to learning coefficients represents that only a subspace of feature space is utilized for classification.
For high-dimensional tensor regression to classification problems, Lian   \cite{lian2021learning} proposed a  support tensor machine model with hinge loss and low-rank regularization, 
and also established the convergence rate of the estimator.
Tan et al. \cite{tan2013logistic} proposed a logistic tensor regression model
 for classification of high-dimensional data
with structural information, which employed the CANDECOMP/PARAFAC (CP) decomposition \cite{carroll1970analysis} to measure the low-rankness of a tensor.
Wimalawarne et al. \cite{wimalawarne2016theoretical} proposed a regularization method combined the logistic regression  loss function and tensor norms  based on the unfolding matrices of a tensor along each mode for binary classification.
Furthermore, when the order of the tensor is second, low-rank tensor learning reduces to low-rank matrix learning, 
which has attracted much attention in the past few decades. For instance, when the loss function is the least squares loss, it is low-rank matrix completion with noisy observations.
There exist many work for matrix completion in the literature, see \cite{candes2009exact, candes2010matrix,  recht2010guaranteed} and references therein.
Besides, for logistic regression, Yin et al.  \cite{yin2018robust} proposed a low-rank plus sparse method to detect the intra-sample outliers via training data.

\subsection{Low-Rank Tensor Learning}

For low-rank tensor learning, the key issue is the definition of the rank of a tensor, which is acquired by tensor decomposition.
Some widely used tensor decompositions include CP decomposition \cite{carroll1970analysis}, Tucker decomposition \cite{tucker1966some}, tensor singular value decomposition (SVD) \cite{Kilmer2011}, tensor train decomposition \cite{oseledets2011tensor},  tensor ring decomposition \cite{zhao2016tensor}, and fully-connected tensor network decomposition \cite{zheng2021fully}.
However, computing the CP rank of a tensor is NP-hard in general \cite{hillar2013most}.
Although the Tucker rank of a tensor can be computed via the SVD of the unfolding matrices easily,
the  Tucker rank minimization is also NP-hard due to the difficulty of matrix rank minimization \cite{candes2009exact}.
Some convex approximation methods were proposed and studied via the sum of nuclear norms (SNN) of unfolding matrices of a tensor for tensor completion \cite{Liu2013,Gandy_2011}.
However, the SNN is not the convex envelope of the sum of entries of Tucker rank of  a tensor \cite{romera2013new}.

For tensor train rank minimization,
which  constitutes of ranks of matrices formed by a well-balanced matricization scheme,
Bengua et al.  \cite{bengua2017efficient} proposed two approaches for low-rank tensor completion via tensor train nuclear norm and its parallel matrix factorization,
which were capable of capturing
the global correlation of the tensor entries.
However,  the above work may destroy the intrinsic structure of a tensor
since  a low-order tensor should be represented into a
higher-order tensor by the  ket augmentation technique.
Based on tensor ring rank minimization,
Yuan et al. \cite{yuan2019tensor} proposed a tensor ring nuclear norm
method  for noisy tensor completion to reduce the computational complexity by using the low-rank assumption on tensor factors instead
of on the original tensor,
where the tensor ring nuclear norm was defined as  the nuclear norms of the unfolding matrices of all factor tensors in tensor ring decomposition.
While the tensor train and tensor ring methods  only established a connection between adjacent two factors, rather
than any two factors.
In order to overcome the limitation of tensor train and tensor ring decomposition,
Zheng et al. \cite{zheng2021fully} proposed a fully-connected tensor network decomposition and presented a novel method for tensor completion via this kind of decomposition, which established an operation between any two factor  tensors.
However, the fully-connected tensor network decomposition may be resulted in large computational burden due to the multiple connections of factor tensors.

\subsection{Tensor SVD}

The tensor SVD was proposed and studied based on tensor-tensor product for third-order tensors in  \cite{Kilmer2011}, which was further extended to higher-order tensors \cite{qin2022low, martin2013order}.
Moreover, the optimal
representation and compression about tensor SVD were studied in
\cite{kilmer2021tensor}.
For the problem of low-rank tensor minimization, the tensor nuclear norm was proposed to approximate the tensor tubal rank in \cite{Semerci2014}.
And the tensor nuclear norm based methods were further presented to low-rank tensor learning problems, see \cite{lu2020tensor,wang2022} and references therein.
However, the tensor nuclear norm is based on Fourier transform, which may be challenged
since the periodicity is assumed. Recently, Song et al. \cite{Song_2020} proposed a transformed tensor SVD via any unitary transform instead of Fourier transform.
Then the transformed tensor nuclear norm (TTNN) was proposed to approximate the transformed multi-rank of a tensor for robust tensor completion, which was capable of  acquiring a lower rank tensor under suitable unitary transformations.
Moreover, the TTNN was applied to other low-rank tensor optimization problems successfully, see \cite{Qiu_2021,qiu2021nonlocal, zhang2022low, song2023tensor} and references therein.
However, the TTNN is just the convex envelope of  sum of each entry of transformed multi-rank of a tensor under the unit ball of tensor spectral norm, which may be suboptimal.
Furthermore, the tensor nuclear norms based on non-invertible
 transformations \cite{jiang2020framelet} and nonlinear transformations \cite{li2022nonlinear}  were proposed and studied for  tensor completion.

On the other hand, some nonconvex approximations were also  proposed and studied for low-rank tensor learning, which can generally approximate the tensor rank better than the convex relaxation methods.
For example, based on SNN for tensor completion,
Zhang proposed a nonconvex method by folding a tenor into a square matrix to approximate the Tucker rank minimization \cite{zhang2019nonconvex}.
Xu et al.  \cite{Xu2015} proposed a low-rank factorization model for tensor completion,
where the unfolding matrices of  a tensor along each mode were  factorized into the product of two smaller matrices.
Besides, Yao et al.  \cite{Yao2020} proposed a nonconvex model based on SNN for low-rank tensor learning
and established the statistical performance on the tensor completion problem,
where the nonconvex regularization was employed onto the singular values of unfolding matrices of a tensor along each mode.
However, the  error bound between the stationary  point of the nonconvex model and the underlying tensor was only established  for low-rank tensor learning with least squares loss.
Moreover, the nonconvex approximation based on SNN may be suboptimal since the SNN is not the tightest convex relaxation of Tucker rank minimization \cite{mu2014square}.

Based on tensor SVD in low-rank tensor learning, Wang et al. \cite{wang2022} proposed a nonconvex model via using the nonconvex function onto the singular values of all frontal slices in the Fourier domain
for tensor completion without noise.
Furthermore,
Qiu et al. \cite{qiu2021nonlocal, qiu2024robust} proposed a nonconvex approach via using nonconvex regularization for robust tensor completion,
where a family of nonconvex functions were employed onto the low-rank and sparse components, respectively.
Moreover, Zhao et al. \cite{Zhao2022} proposed a nonconvex tensor surrogate function for robust tensor completion
via equivalent nonconvex surrogates with difference of
convex functions structures.
While they only established the recovery error bound between the estimator of an approximate convex model and the underlying tensor,
and did not analyzed the error bound of the nonconvex model.
In addition, Gao et al. \cite{gao2023tensor} proposed an $\ell_p$ $(0<p<1)$ method
by employing the $\ell_p$ function to each singular value of the low-rank tensor
in the Fourier domain and each entry of the sparse tensor for tensor robust principle component analysis,
where the $\ell_p$ norm was used to approximate tensor fibered rank and measure sparsity, respectively.
Recently, Zhang et al. \cite{zhang2022sparse} proposed a sparse tensor factorization method based on tensor-tensor product under general observations.
In the previous two  work, the error bounds between the global minimizer of the nonconvex model  and the underlying tensor  were established under some conditions.
However, there is a significant gap between the theory and practice computation
since it is difficult to achieve the global minimizers of the these nonconvex models in numerical algorithms.

\subsection{The Contribution}

In this paper, we propose a nonconvex approach for low-rank tensor learning.
Specifically,
a general loss function is utilized between  given samples  and the underlying tensor to fit the observed data.
Moreover, in order to explore the global low-rankness of the underlying tensor,
a family of nonconvex functions are employed onto the singular values of all frontal slices of a tensor in the transformed domain.
Compared with TTNN, the nonconvex method is capable of acquiring a lower rank tensor, which is preferred for low-rank tensor learning.
Besides, an error bound between any stationary point of the proposed nonconvex  model and the underlying tensor
is established under  the restricted strong convexity (RSC) condition on the loss function and suitable regularity conditions on the nonconvex penalty,
which is smaller than that of the Tucker based method in \cite{Yao2020}.
In particular, our model covers the least squares loss  for tensor completion and logistic regression for binary classification.
Moreover, by reformulating the nonconvex regularization into a difference of  convex (DC) functions,
a proximal majorization-minimization (PMM) algorithm is designed to solve the proposed nonconvex model,
where the loss function and one convex function in the DC structure are linearized at the current point of iterations.
Moreover, we show that
the PMM algorithm globally converges to a stationary point of the proposed nonconvex model  under the Kurdyka-$\L$ojasiewicz (KL) assumption on the nonconvex penalty, where the convergence rate of PMM is also established.
 And an alternating direction method of multipliers (ADMM)  is presented to solve the resulting subproblem in PMM.
Numerical examples on tensor completion and binary classification are conducted to demonstrate the superiority of the proposed method compared with  several state-of-the-art methods.

The remaining parts of this paper are organized as follows.
Next, we give some notations and notions about tensors and transformed tensor SVD.
In  Section \ref{chap3}, we propose a nonconvex model for low-rank tensor learning based on TTNN.
The error bound of any stationary point   of the proposed model is established under some conditions.
A PMM algorithm is designed to solve the proposed nonconvex model
and the ADMM is applied to solve the resulting subproblem in Section \ref{PMMAl}, where the global convergence and convergence rate of PMM are also established.
In Section \ref{NumericEx}, some numerical experiments are conducted on tensor completion and logistical regression  to  illustrate
the advantage of our method over other existing approaches.
We conclude this paper in Section \ref{Conc}. Finally, all the
technical proofs are deferred to the Appendix.

\subsection{Preliminaries}\label{chap2}

Some notations  used throughout this paper are summarized  in Table \ref{table1}, where the size of a tensor is  $n_1\times n_2\times n_3$.

Now we give the definition of subdifferential of a function.
\begin{definition}\label{defin1}\cite[Definition 8.3]{R.TyrrellRockafellar1998}
Consider a function $f:\mathbb{R}^n\rightarrow (-\infty,+\infty] $ and a point $\mathbf{x}\in \mathbb{R}^n$ with the finite $f(\mathbf{x})$.
The regular subdifferential of $f$ at $\mathbf{x}$ is defined by

$$
\widehat{\partial} f(\mathbf{x}):=\left\{\mathbf{y} \in \mathbb{R}^n: \liminf _{\mathbf{z} \rightarrow \mathbf{x}, \mathbf{z} \neq \mathbf{x}} \frac{f(\mathbf{z})-f(\mathbf{x})-\langle \mathbf{y}, \mathbf{z}-\mathbf{x}\rangle}{\|\mathbf{z}-\mathbf{x}\|_2} \geq 0\right\} .
$$
The (limiting) subdifferential of the function $f$ at $\mathbf{x}$ is defined by
$$
\partial f(\mathbf{x}):=\left\{\mathbf{y} \in \mathbb{R}^n: \  \exists \mathbf{x}^k \stackrel{f}{\rightarrow} \mathbf{x}, \mathbf{y}^k \rightarrow \mathbf{y} \  \text{with} \ \mathbf{y}^k \in \widehat{\partial} f(\mathbf{x}^k) \text { for each } k\right\},
$$
where $\mathbf{x}^k \stackrel{f}{\rightarrow} \mathbf{x}$ means $\mathbf{x}^k \rightarrow \mathbf{x}$ with $f(\mathbf{x}^k) \rightarrow f(\mathbf{x})$.
\end{definition}

\begin{table}[!t]
	\caption{Notations}\label{table1}
	\footnotesize
	\centering
	\begin{tabular}{l|c}
		\toprule
		Notations & Description \\
		\midrule
		$a$/$\bf a$/$\bf A$/$\mathcal A$ & Scalars/Vectors/Matrices/Tensors \\
		$\mathcal A_{ijk}$ &  The $(i,j,k)$-th element of $\mathcal A$\\
		$\cdot^{T}$ & The conjugate transpose operator \\
		$\textup{Tr}(\cdot)$ & The trace of a matrix \\
		$\mathcal A^{\langle i\rangle}$ & The $i$-th frontal slice of $\mathcal A$ \\
		$\mathcal A_{(i)}$ &  The mode-$i$ unfolding of $\mathcal A$ \\
		$\text{Fold}_i(\cdot)$ & The inverse operator of mode-$i$ unfolding, i.e., $\text{Fold}_i(\mathcal A_{(i)})=\mathcal A$ \\
		$\|{\bf a}\|_2$ &  The $\ell_2$ norm of ${\bf a}$ \\
		$\operatorname{\text{Diag}}(\bf a)$ & A diagonal matrix with the $i$-th diagonal element being the $i$-th component of $\bf a$ \\	
		$\sigma_j({\bf A})$ & The $j$-th largest singular value of $\bf A$
		\\
		$\|{\bf A}\|_{*}$ & The nuclear norm of ${\bf A}$
		defined as $\|{\bf A}\|_*:=\sum_{j=1}^{\min\{n_1,n_2\}}\sigma_j({\bf A})$	\\
		$\|{\bf A}\|$ & The spectral norm of ${\bf A}$ defined as $\|{\bf A}\|:= \sigma_1({\bf A})$
		\\
		$\|{\bf A}\|_F$ &  Frobenius norm of $\bf A$ defined as $\|{\bf A}\|_F:=\sqrt{\textup{Tr}({\bf A}^T\bf A)}$ \\
		$\mathbf U$  &  A unitary matrix  satisfying
		$\mathbf U\mathbf U^{T}=\mathbf U^{T}\mathbf U=\mathbf I_{n_3}$,
		where $\mathbf I_{n_3}$ is the $n_3\times n_3$ identity matrix \\
		$	\widehat{\mathcal X}_\mathbf U$ or $\mathbf U[\mathcal {X}]$	&	  $
		\widehat{\mathcal X}_\mathbf U=\mathbf U[\mathcal {X}]:=\text{Fold}_3(\mathbf U \mathcal X_{(3)})
		$ \\
		$\langle \mathcal{A},\mathcal{B} \rangle$
		& The inner product of two tensors defined as $\langle \mathcal{A},\mathcal{B} \rangle:=\sum_{i=1}^{n_3}\textup{Tr}((\mathcal A^{\langle i\rangle})^T\mathcal B^{\langle i\rangle})$ \\
		$\operatorname{vec}(\mathcal A)$  & Vectorizing a tensor $\mathcal A$ into a vector\\
		$\|\mathcal A\|_\infty$ & Tensor $\ell_\infty$ norm of $\mathcal A$ defined as $\|\mathcal A\|_{\infty}:=\max|\mathcal A_{ijk}|$ \\
		$\|\mathcal{A}\|_F$ &  Tensor Frobenius norm of $\mathcal A$ defined as $\|\mathcal{A}\|_F:=\sqrt{\langle\mathcal{A},\mathcal{A} \rangle }$ \\
		$\delta_D(\cdot)$ & The indicator function of a set $D$ with $\delta_D(a)=0$ if $a\in D$, otherwise $+\infty$ \\
	$	\mathop{\mathrm{dist}}(\mathbf{a},D)$ & The distance from
	$\mathbf{a}$ to $D$ and is defined as $	\mathop{\mathrm{dist}}(\mathbf{a},D):=\inf\{\|\mathbf{y}-\mathbf{a}\|_2, \mathbf{y}\in D\}$ \\
		\bottomrule
	\end{tabular}
\end{table}

The KL  function plays a vital role for the convergence analysis in our algorithm, and we list the definition of KL function in the following definition.

\begin{definition}\label{DeKLF}
	\cite[Definition 3]{Bolte2013} 
	Let $f:\mathbb R^n\rightarrow \mathbb R\cup\{+\infty\}$ be a proper lower semicontinuous function.
	We say that $f$ has the Kurdyka-$\L$ojasiewicz (KL)
	property at point $\mathbf{x}^{*}\in\text{dom}(\partial f)$,
	if there exist a neighborhood $U$ of $\mathbf{x}^{*}$, $\eta\in(0,+\infty]$ and
	a continuous concave function $\varphi : [0,\eta)\rightarrow \mathbb R_{+}$ such that:\par
	(i)   $\varphi(0)=0$,\par
	(ii)   $\varphi$ is $C^1$ on $(0,\eta)$,\par
	(iii)   for all $s\in(0,\eta)$, $\varphi{'}(s)>0$,\par
	(iv)  for all $\mathbf{x} \text{ in } U \cap [f(\mathbf{x}^{*})<f<f(\mathbf{x}^{*})+\eta]$, the KL inequality holds:
	\begin{equation}\label{test316KL}
		\varphi{'}(f(\mathbf{x})-f(\mathbf{x}^{*}))\mathop{\mathrm{dist}}(0,\partial f(\mathbf{x})) \geq 1.
	\end{equation}
If $f$ satisfy the KL property at each point of $\text{dom}(\partial f):=\{\mathbf{v}\in \mathbb R^n : \partial f(\mathbf{v})\neq\emptyset\}$,  then $f$ is called a KL function.
\end{definition}

A function is said to have the KL property at $\mathbf{x}^{*}$ with an exponent $\alpha$ if the function $\varphi$ in  Definition \ref{DeKLF} takes the form of $\varphi(s)=\mu_1 s^{1-\alpha}$ with $\mu_1>0$ and $\alpha \in[0,1)$.
The proper closed semi-algebraic functions are KL functions with exponent $\alpha \in[0,1)$
\cite{Pong2018}.


Next, we review the definition  of transformed tensor SVD for third-order tensors,
see \cite{Song_2020} for more details.
We denote $\overline{\mathcal X}$ (also denoted by $\text{bdiag}({\widehat{\mathcal X}}_{\mathbf U})$) as a block diagonal matrix,
where the $i$-th block is the matrix $\widehat{\mathcal X}_\mathbf U^{\langle i\rangle}$, $i=1,\ldots, n_3$, i.e.,
$$
\overline{\mathcal X}=\text{bdiag}({\widehat{\mathcal X}}_{\mathbf U})
:=\begin{bmatrix}
    {\widehat{\mathcal X}}_\mathbf U^{\langle 1\rangle}   & &     \\
       &\ddots &  \\
      & & \widehat{\mathcal X}_\mathbf U^{\langle n_3\rangle}
\end{bmatrix}.
$$
The corresponding inverse operator, denoted by $``\text{fold}_{3}(\cdot)"$, is defined as
\begin{equation}\label{Fold3}
\text{fold}_{3}(\text{bdiag}({\widehat{\mathcal X}}_{\mathbf U}))=\widehat{\mathcal X}_{\mathbf U}.
\end{equation}

\begin{definition}\cite[Definition 1]{Song_2020}
	The $\mathbf U$-product of arbitrary
	two tensors $\mathcal {A}\in \mathbb {C}^{n_1\times n_2\times n_3}$ and  $\mathcal {B}\in \mathbb{C}^{n_2\times l\times n_3}$ is defined as
	$
	\mathcal A\diamond_{\mathbf U}\mathcal B=\mathbf U^{T}[\text{\rm{{fold}}}_{3}(\text{\rm{bdiag}}(\widehat {\mathcal A}_\mathbf U)\cdot \text{\rm{bdiag}}(\widehat
	{\mathcal B}_\mathbf U)) ]\in\mathbb {C}^{n_1\times l\times n_3}.
	$
\end{definition}

\begin{definition}\label{definition1}\cite[Definition 7]{Song_2020}
The transformed tensor nuclear norm  of  $\mathcal X \in \mathbb {C}^{n_1\times n_2\times n_3}$
is defined as
$
\|\mathcal X\|_{\operatorname{TTNN}}=\sum_{i=1}^{n_3}\|\widehat{\mathcal X}_\mathbf U^{\langle i\rangle}\|_*.
$
\end{definition}

It can be easily verified that $\|\mathcal X\|_{\text{TTNN}}=\|\overline{\mathcal X}\|_*.$

\begin{definition}\label{definition2}\cite{Song_2020}
The tensor spectral norm with respect to $\mathbf U$, denoted by $\|\mathcal X\|_{\mathbf U}$, is defined as
$\|\mathcal X\|_{\mathbf U}=\|\overline{\mathcal X}\|$.
\end{definition}

Now we recall the definitions of the diagonal tensor, conjugate transpose, and unitary tensor \cite{Kilmer2011, Song_2020}. A third-order  tensor is called to be diagonal if each frontal slice is a diagonal matrix.
The conjugate transpose of $\mathcal X \in \mathbb {C}^{n_1\times n_2\times n_3}$ with respect to $\mathbf U$, denoted by $\mathcal X^T$,
is defined as $\mathcal X^T=\mathbf U^T[\text{fold}_{3}((\text{bdiag}({\widehat{\mathcal X}}_{\mathbf U}))^T)]\in\mathbb {C}^{n_2\times n_1\times n_3}$.
 A tensor $\mathcal U$ is called to be unitary if $\mathcal U\diamond_{\mathbf U}\mathcal U^T=\mathcal U^T\diamond_{\mathbf U}\mathcal U=\mathcal{I}_{\mathbf U}$, where $\mathcal{I}_{\mathbf U}$ denotes the identity tensor and is defined as each frontal slice of $\mathbf{U}[\mathcal{I}_{\mathbf U}]$ being the identity matrix. Next we give the definition of transformed tensor SVD of a third-order tensor.

\begin{theorem}\cite[Theorem 5.1]{Kernfeld2015}
The transformed tensor singular value decomposition of $\mathcal X \in \mathbb {C}^{n_1\times n_2\times n_3}$ is given by
	$\mathcal X= \mathcal U \diamond_{\mathbf U}\Sigma\diamond_{\mathbf U}\mathcal V^T,$
where $\Sigma \in \mathbb {C}^{n_1\times n_2\times n_3}$ is a diagonal tensor,
$\mathcal U \in \mathbb {C}^{n_1\times n_1\times n_3}$ and $\mathcal V \in \mathbb {C}^{n_2\times n_2\times n_3}$ are unitary tensors with respect to $\mathbf U$-product.
\end{theorem}

\begin{definition}\cite[Definition 6]{Song_2020}
The transformed multi-rank of  $\mathcal X \in \mathbb {C}^{n_1\times n_2\times n_3}$ is
a vector ${\bf r} = (r_1,r_2,...,r_{n_3})$
with $r_i=\textup{rank}({\widehat{\mathcal X}}_\mathbf U^{\langle i\rangle}),i=1,2,\ldots,n_3$,
where $\textup{rank}({\widehat{\mathcal X}}_\mathbf U^{(i)})$ denotes the rank of the matrix ${\widehat{\mathcal X}}_\mathbf U^{\langle i\rangle}$.
\end{definition}

\section{Nonconvex Model for Low-Rank Tensor Learning}\label{chap3}

In this section, we present the framework of our approach for  the problem of low-rank tensor learning.
Given $n$ samples, we consider a general loss function $f_{n,\mathcal{Y}} (\mathcal X)$,
which is used to fit the parameter tensor $\mathcal{X}\in\mathbb{R}^{n_1\times n_2\times  n_3}$ and the observed data $\mathcal{Y}$.
Suppose that the parameter tensor $\mathcal{X}$ is low-rank, which we aim to estimate.
Now a nonconvex model based on TTNN  with a general loss function  is presented for low-rank tensor learning:
\begin{equation}\label{test1}
\begin{aligned}
\min _\mathcal X & \ f_{n,\mathcal{Y}} (\mathcal X)+ \beta G_\lambda(\mathcal X) \\
\text{s.t.}& \  \|\mathcal X\|_{\infty} \leq c,
\end{aligned}
\end{equation}
where $f_{n,\mathcal{Y}} (\mathcal X)$ represents a differentiable loss function with $n$ samples, $G_\lambda(\mathcal X)$ represents the nonconvex regularization defined as
\begin{equation}\label{Glmbade}
G_\lambda(\mathcal X):=\sum_{i=1}^{n_3} \sum_{j=1}^{\min\{n_1,n_2\}} g_\lambda(\sigma_j(\widehat{\mathcal X}_\mathbf U^{\langle i\rangle})),
\end{equation}
$\beta>0$ is a penalty parameter, and $c>0$ is a given constant.
Here  $g_\lambda(\cdot)$ is a nonconvex function with respect to the parameter $\lambda$.
In addition, the tensor $\ell_{\infty}$ norm constraint  is effective to exclude over-spiky tensors.

Throughout this paper, the nonconvex function $g_\lambda(\cdot)$  should satisfy the following assumptions.

\begin{assumption}\label{assum1}
	The  function $g_\lambda\left(x\right): \mathbb{R}\rightarrow \mathbb{R}_+$ is symmetric and  has the following properties:
	
	(i) $g_\lambda(x)$ is concave  and  non-decreasing for $x\geq0$ with $g_\lambda\left(0\right)=0$.
	
	(ii) The function $\frac{g_\lambda\left(x\right)}{x}$ is non-increasing for $x>0$.
	
	(iii) $g_\lambda(x)$ is differentiable for any $x\neq0$ and $\lim\limits_{x \rightarrow 0^+}g_\lambda^{\prime}(x)=\lambda k_0$, where $k_0>0$ is a  constant and $\lambda k_0$ is an upper bound of $g_\lambda^{\prime}(\cdot)$ on $(0,+\infty)$.
	
	(iv) There exists a  parameter $\mu>0$ such that $g_\lambda(x)+\frac{\mu}{2}x^2$ is convex on $(0,+\infty)$.
\end{assumption}

A lot of nonconvex functions satisfy Assumption \ref{assum1} in the literature, such as smoothly clipped absolute deviation (SCAD) \cite{fan2001variable}, minimax concave penalty (MCP) \cite{zhang2010nearly}, logarithmic function \cite{candes2008enhancing}, which will be given in Table \ref{table2}.
These nonconvex function can approximate the $\ell_0$ norm better than the $\ell_1$ norm, which can yield a sparser solution in statistical learning \cite{Gong2013}.
In model (\ref{test1}), we propose to employ a family of nonconvex functions based on TTNN to explore the global low-rankness of the underlying tensor in low-rank tensor learning.
Although the TTNN can get a lower rank tensor compared with tensor nuclear norm under suitable unitary transformations \cite{Song_2020},
the TTNN is just the $\ell_1$ norm  of all singular values vectors of the frontal slices of a tensor in the transformed domain, which also suffers from the drawback of $\ell_1$ norm.
 Recently, some nonconvex surrogate functions  have been demonstrated
 more efficiently than the convex relaxation in various fields such as 
 statistical learning \cite{zhang2010nearly,fan2001variable}, compressed sensing  \cite{candes2008enhancing}, and tensor completion \cite{zhang2019nonconvex, qiu2021nonlocal}.
Moreover, the nonconvex relaxation based on TTNN can help to achieve a lower rank tensor than TTNN for robust tensor completion \cite{qiu2021nonlocal}.
In low-rank tensor learning, we employ a family of nonconvex functions onto each singular value of the frontal slices of  the underlying tensor in the transformed domain.
Compared with the TTNN,
the main advantage of the proposed method is that  the nonconvex functions can  approximate the sum of  entries  of the transformed multi-rank of a tensor better.

For the general loss function $f_{n,\mathcal{Y}}$ in model (\ref{test1}), we just need it to be differentiable.
In various real-world applications, we can specify the loss function $f_{n,\mathcal{Y}}$.
In particular, when $f_{n,\mathcal{Y}}$ is the least squares loss function, model (\ref{test1}) can be used for noisy tensor completion.
When $f_{n,\mathcal{Y}}$ is the logistic regression loss function,
model (\ref{test1}) can be utilized for binary classification in machine learning.

\begin{remark}
	When $f_{n,\mathcal{Y}}(\mathcal{X})=\frac{1}{2}\|\mathcal{P}_{\Omega}(\mathcal{Y})-\mathcal{P}_{\Omega}(\mathcal{X})\|_F^2$,  model  (\ref{test1}) can be used for tensor completion with noisy observations.
Recently, some nonconvex surrogates were proposed and studied for tensor completion \cite{zhang2019nonconvex, wang2022, qiu2021nonlocal},
where a family of nonconvex functions were employed onto the low-rank component of a tensor.
For example, Wang et al. \cite{wang2022} proposed to utilize a general nonconvex surrogate of the tensor tubal rank for tensor completion and a least squares loss function for the data-fitting term,
 where  the tensor nuclear norm was used in the Fourier domain.
And Zhang \cite{zhang2019nonconvex} used a family of nonconvex function onto the singular values of the square matrice via matricizing a tensor to approximate the Tucker rank of  a tensor.
However, they do not analyze the statistical performance of their proposed models.
\end{remark}

\begin{remark}
Based on the overlapped nuclear norm of a tensor,
Yao et al.  \cite{Yao2020}
proposed a nonconvex approach for low-rank tensor learning, where the nonconvex functions were employed onto each singular value of unfolding matrices of a tensor.
However, the unfolding based methods are challenged due to its suboptimality,
 where the overlapped nuclear norm was not the convex envelope of the sum of Tucker rank of a tensor \cite{romera2013new}.
Moreover,  they only analyzed the statistical performance of their model for  the least squares  loss function.
We will establish the error bound between any stationary point of  model (\ref{test1}) and the underlying tenor for a general loss function  in the next subsection.
\end{remark}

\subsection{Statistical Guarantee}\label{StaGua}

In this subsection, we first introduce the definition of the  RSC condition for a general loss function  and then establish the error bound between any stationary point of the proposed model and the underlying tensor.


The RSC  condition of a differentiable function in the tensor case
 plays a vital role in establishing the error bound of the proposed model.
The foundational works on the RSC condition in the vector case are due to \cite{2012A,loh15a}.
For any tensor $\tilde{\mathcal V}\in\mathbb {R}^{n_1\times n_2\times n_3}$,
we say that
the function $f_{n,\mathcal{Y}}(\mathcal{X}):\mathbb {R}^{n_1\times n_2\times n_3}\rightarrow \mathbb {R}$ satisfies
the RSC condition  if
\begin{equation}\label{test0}
	\langle\nabla f_{n,\mathcal{Y}}(\mathcal X^*+\tilde{\mathcal V})
	-\nabla f_{n,\mathcal{Y}}(\mathcal X^*), \tilde{\mathcal V}\rangle
	\geq\left\{\begin{array}{ll}
		\alpha_1\|\tilde{\mathcal V}\|_F^2-\tau_1 \frac{\log d}{n}\|\tilde{\mathcal V}\|_{\text{TTNN}}^2, & \text {if }\|\tilde{\mathcal V}\|_F \leq 1 ,\\
		\alpha_2\|\tilde{\mathcal V}\|_F-\tau_2 \sqrt{\frac{\log d}{n}}\|\tilde{\mathcal V}\|_{\text{TTNN}}, & \text {otherwise},
	\end{array}\right.
\end{equation}
where $\mathcal X^*$ denotes the ground-truth tensor, $d:=n_1n_2n_3$, $n$ is the number of samples, and  $\alpha_1,\alpha_2>0$, $\tau_1,\tau_2\geq0$ are given constants.

The RSC condition involves a lower bound on the remainder in the
first-order Taylor expansion of $f_{n,\mathcal{Y}}$, where we only require $f_{n,\mathcal{Y}}$ to be differentiable.
If $f_{n,\mathcal{Y}}$  is convex, the    left hand in (\ref{test0}) is always nonnegative, and then (\ref{test0}) holds trivially for $\frac{\|\tilde{\mathcal V}\|_{\textup{TTNN}}}{\|\tilde{\mathcal V}\|_F}\geq\sqrt{\frac{\alpha_1 n}{\tau_1\log d}}$ and $\frac{\|\tilde{\mathcal V}\|_{\textup{TTNN}}}{\|\tilde{\mathcal V}\|_F}\geq \frac{\alpha_2 }{\tau_2}\sqrt{\frac{n}{\log d}}$.
As a result, the inequality in (\ref{test0}) only enforces a type of strong convexity condition
over the cone
$\left\{\frac{\|\tilde{\mathcal V}\|_{\textup{TTNN}}}{\|\tilde{\mathcal V}\|_F}\leq \tilde{c} \sqrt{\frac{n}{\log d}}\right\}$, where $\tilde{c}>0$ is a constant.
In particular, we will show the least squares loss and logistic regression loss satisfy the RSC condition in Appendix A and B.
More discussions about the RSC condition in the vector case can be referred to  \cite{loh15a}.

\begin{remark}
The  RSC condition has been widely studied for statistical learning  in the  literature.
	For example, the least squares loss for  linear regression  in the  matrix case  satisfies the RSC condition \cite{gui2016towards}.
Moreover, the loss functions of the generalized linear model  and  corrected linear model satisfy the RSC condition by  selecting  appropriate parameters \cite{2012A,loh15a, Loh2012}, where the RSC condition of these loss functions is employed in the vector case.
\end{remark}


Let
$\tilde{\mathcal X}$  be a stationary point of problem (\ref{test1}).
In the following analysis of the statistical performance guarantee of the proposed model,
we enforce the additional constraint $\|\mathcal X\|_{\operatorname{TTNN}}\leq t$
for any  $\mathcal X \in \mathbb {R}^{n_1\times n_2\times n_3}$ in model (\ref{test1}), where $t>0$ is a given constant.
The main result is described in the following theorem.

\begin{theorem}\label{theo1}
Suppose that the regularizer $g_{\lambda}(\cdot)$ satisfies Assumption \ref{assum1} and
the loss function $f_{n,\mathcal{Y}}$ satisfies the RSC condition (\ref{test0})
with $\frac{3}{4} \beta\mu<\alpha_1$, where $\mu$ is defined in Assumption \ref{assum1}(iv).
Consider the parameter $\lambda$ with
\begin{equation}\label{test20}
\begin{aligned}
\frac{4}{\beta k_0} \max \left\{\| \nabla f_{n,\mathcal{Y}}(\mathcal X^*)\|_\mathbf U, \alpha_2 \sqrt{\frac{\operatorname{log} d}{n}}\right\} \leq \lambda \leq \frac{\alpha_2}{6  t \beta k_0},
\end{aligned}
\end{equation}
and
\begin{equation}\label{test27}
n \geq \frac{16 t^2 \max \{\tau_1^2,\tau_2^2\}}{\alpha_2^2} \log d,
\end{equation}
where $d:=n_1n_2n_3$.
Then any stationary point $\tilde{\mathcal X}$ of model (\ref{test1}) satisfies
$$
\|\tilde{\mathcal X}-\mathcal X^*\|_F\leq
\frac{6\beta\lambda k_0\sqrt{\sum_{i=1}^{n_3}r_i}}
{4\alpha_1-3\beta\mu},
$$
where $r_i$ denotes the rank of matrix 
$(\widehat{\mathcal X^*})_\mathbf U^{\langle i\rangle}, i=1,\ldots, n_3.$
\end{theorem}

The proof of Theorem \ref{theo1} is left to Appendix E.
From Theorem \ref{theo1}, we know that the upper bound is related to the sum of each entry of the transformed  multi-rank of the underlying  tensor,
which will be small under suitable unitary transformations \cite{Song_2020}.
More details about the  choice of unitary transformations in TTNN can be referred to \cite{Song_2020, song2023tensor}.
Note that the parameters  $k_0,\mu, \lambda$  are related to the nonconvex function $g_\lambda$, which are given in Assumption \ref{assum1}.
Moreover, we need the loss function $f_{n,\mathcal{Y}}$ to satisfy  the RSC condition (\ref{test0}) in Theorem \ref{theo1}.
We will show the least squares  loss for tensor completion and logistic regression for binary classification in low-rank tensor learning satisfy the RSC condition in Appendix A and B,
where the two loss functions are widely used in practice.

In particular, the error bound in Theorem \ref{theo1} can reduce to that of the TTNN model,
where the nonconvex regularization term of model (\ref{test1}) is replaced by TTNN.
Notice that the error bound in Theorem \ref{theo1} is just the tensor Frobenius
norm of the difference between any stationary point of model (\ref{test1})  and the underlying tensor,
which is the worst case  for the error bound of the nonconvex model in (\ref{test1}).
In numerical computation, we can design an efficient algorithm to obtain a stationary point of the nonconvex model,
which will be shown in the next section.
Now we compare with the error bound of the model in  \cite{Yao2020},
which utilized the least squares loss and the nonconvex regularization based on the nuclear norms of all unfolding matrices of a tensor for low-rank tensor learning.
Note that $\frac{r_1+\cdots +r_{n_3}}{n_3}\leq \max\{r_1,\ldots, r_{n_3}\}\leq \textup{rank}(\mathcal X_{(1)}^*)$. This demonstrates that a tensor
with low Tucker rank has low average transformed multi-rank.
Compared with the error bound in \cite{Yao2020},
which used the Tucker rank in the model and was on the order of $O(\sum_{i=1}^d\sqrt{ \textup{rank}(\mathcal X_{(i)}^*)})$,
the error bound in Theorem  \ref{theo1} is smaller
when $n_3$ is not too large or the rank of unfolding matrices of the underlying tensor is large.

\section{Optimization Algorithm}\label{PMMAl}

In this section, we first design a proximal majorization-minimization (PMM) algorithm \cite{Zhao2022, Tang2020} to solve  problem (\ref{test1}), and then establish the global convergence  and convergence rate of PMM under  very mild conditions.
Finally, an ADMM based algorithm is utilized to solve the resulting subproblem in PMM.

\subsection{PMM Algorithm}
Note that problem (\ref{test1}) is nonconvex and nonsmooth since $G_\lambda(\mathcal X)$  is nonconvex and nonsmooth.
The nonconvexity of the objective function results in great challenges to numerical computation and theoretical analysis.
By the special structure of some nonconvex functions,
they can be written as the difference of two convex functions,
which has a variety of applications in statistical and  machine learning \cite{Hartman1959,Gong2013,Thi2015,Ahn2017}.
In particular, we assume that the nonconvex function $g_\lambda(x)$ in our model  can be written as
\begin{equation}\label{test53}
	g_\lambda(x)=s_1(x)-s_2(x),
\end{equation}
where $s_1$ is convex  and $s_2$ satisfies Assumption
\ref{assum5}.

\begin{assumption}\label{assum5}
The function $s_2:\mathbb{R}\rightarrow\mathbb{R}$ satisfies the following assumptions.
	
	(a) $s_2$ is convex and  symmetric, i.e., $s_2(-x)=s_2(x)$.
	
	(b) $s_2$ is differentiable and its derivative $s_2'$ is  locally Lipschitz continuous.
\end{assumption}

A lot of nonconvex functions satisfying  Assumption \ref{assum1} can be expressed in the  decomposition formulation (\ref{test53}), which are summarized
in Table \ref{table2}. Besides,  $s_2$ in Table \ref{table2} also satisfies Assumption \ref{assum5}.

\begin{table}[htpb]
	\caption{Examples of the nonconvex function $g_\lambda(x)$ and its corresponding difference of two convex functions (i.e., $g_\lambda(x)=s_1(x) -s_2(x)$), where $\gamma>0$ for Logarithm and MCP, and $\gamma>1$ for SCAD.}\label{table2}
	\centering
	\begin{tabular}{c|c|c|c}
		\toprule
		Nonconvex function & $g_\lambda(x) (x \geq 0, \lambda>0)$ & $s_1(x)$ & $s_2(x)$ \\
		\midrule  Logarithm \cite{candes2008enhancing}  & $\lambda\log (\frac{x}{\gamma} +1) $
		& $\lambda x$
		& $\lambda x-\lambda\log (\frac{x}{\gamma} +1), x>0 $ \\
		\midrule MCP \cite{zhang2010nearly} & $\begin{cases}\lambda x-\frac{x^2}{2 \gamma}, & x \leq \gamma \lambda, \\
			\frac{1}{2} \gamma \lambda^2, & x>\gamma \lambda .\end{cases} $ &
		$\lambda x$ &
		$\begin{cases}\frac{x^2}{2\gamma}, & x \leq \gamma \lambda, \\
			\lambda x- \frac{\gamma \lambda^2}{2}, & x>\gamma \lambda .\end{cases}$\\
		\midrule  SCAD \cite{fan2001variable} & $\begin{cases}\lambda x, & x<\lambda . \\
			\frac{-x^2+2 \gamma \lambda x-\lambda^2}{2(\gamma-1)}, & \lambda \leq x< \gamma \lambda, \\
			\frac{\lambda^2(\gamma+1)}{2}, & x \geq \lambda\gamma.\end{cases}$
		& $\lambda x$
		& $\begin{cases}0, & x<\lambda, \\
			\frac{x^2-2\lambda x+\lambda^2}{2(\gamma-1)}, & \lambda \leq x<\gamma \lambda, \\
			\lambda x- \frac{(\gamma+1)\lambda^2}{2}, & x \geq \gamma \lambda .\end{cases}$\\
		\bottomrule
	\end{tabular}
\end{table}

Based on (\ref{test53}), we can rewritten $G_\lambda(\mathcal X)$ in (\ref{test1}) as follows:
\begin{equation}\label{eq22}
	\begin{aligned}
		G_\lambda(\mathcal X)=S_1(\mathcal X)-S_2(\mathcal X),
	\end{aligned}
\end{equation}
where
$
S_1(\mathcal X)=\sum_{i=1}^{n_3} \sum_{j=1}^{\min\{n_1,n_2\}} s_1(\sigma_j(\widehat{\mathcal X}_\mathbf U^{\langle i\rangle}))
$
and
\begin{equation}\label{S2Def}
S_2(\mathcal X)=\sum_{i=1}^{n_3} \sum_{j=1}^{\min\{n_1,n_2\}} s_2(\sigma_j(\widehat{\mathcal X}_\mathbf U^{\langle i\rangle})).
\end{equation}
Consequently,
problem (\ref{test1}) can be reformulated equivalently as  follows:
\begin{equation}\label{test42}
	\min _\mathcal X f_{n,\mathcal{Y}} (\mathcal X)+\beta S_1(\mathcal X)-\beta S_2(\mathcal X)+\delta_D(\mathcal X), 
\end{equation}
where $\delta_D(\mathcal X)$ is the indicator function of the set $D$ with
$D:=\{\mathcal X: \|\mathcal X\|_{\infty} \leq c\}$.

We adopt the PMM algorithm  to solve  problem (\ref{test42}),
whose main idea  is to linearize the concave function $-S_2(\mathcal X)$ and the smooth loss function $f_{n,\mathcal{Y}}(\mathcal X)$ of the objective function in (\ref{test42}) at the current iteration point $\mathcal X^t$.
Specifically,
given $\mathcal X^t\in\mathbb {R}^{n_1\times n_2\times n_3}$,
we consider to solve the following  problem:
\begin{equation}\label{test44}
	\begin{aligned}
		\min _\mathcal X & \ f_{n,\mathcal{Y}} (\mathcal X^t)+\langle \nabla f_{n,\mathcal{Y}}(\mathcal X^t),\mathcal X-\mathcal X^t \rangle
		+\frac{\rho}{2}\|\mathcal{X}-\mathcal{X}^{t}\|_{F}^{2}
		+\beta S_1(\mathcal X)-\beta S_2(\mathcal X^t)\\
		& \ -\beta\langle \nabla S_2(\mathcal X^t),\mathcal X-\mathcal X^t \rangle
		+\delta_D(\mathcal X), 
	\end{aligned}
\end{equation}
where  $\rho>0$ is a given constant.
Notice that problem (\ref{test44}) is equivalent to
\begin{equation}\label{test46}
	\begin{aligned}
		\min_\mathcal X \beta S_1(\mathcal X)
		+\langle \nabla f_{n,\mathcal{Y}}(\mathcal X^t)-\beta\nabla S_2(\mathcal X^t),\mathcal X-\mathcal X^t \rangle
		+
		\frac{\rho}{2}\|\mathcal{X}-\mathcal{X}^{t}\|_{F}^{2}
		+
		\delta_D(\mathcal X). 
	\end{aligned}
\end{equation}

Denote
\begin{equation}\label{eq3}
	H(\mathcal{X}):=f_{n,\mathcal{Y}}(\mathcal{X})+\beta G_\lambda(\mathcal{X})+\delta_D(\mathcal X)
	=f_{n,\mathcal{Y}}(\mathcal{X})+\beta S_1(\mathcal X)-\beta S_2(\mathcal X)+\delta_D(\mathcal X),
\end{equation}
and
\begin{equation}\label{eq4}
	\begin{aligned}
		Q(\mathcal{X},\mathcal X^t)
		:=& \ f_{n,\mathcal{Y}}(\mathcal X^t)
		+\langle \nabla f_{n,\mathcal{Y}}(\mathcal X^t),\mathcal X-\mathcal X^t \rangle
		+\frac{\rho}{2}\|\mathcal{X}-\mathcal{X}^{t}\|_{F}^{2}
		+\beta S_1(\mathcal X)-\beta S_2(\mathcal X^t)\\
		& \ -\beta\langle \nabla S_2(\mathcal X^t),\mathcal X-\mathcal X^t \rangle
		+\delta_D(\mathcal X).
	\end{aligned}
\end{equation}
It can be easily verified that $Q(\mathcal X^t,\mathcal X^t)=H(\mathcal X^t).$

However, it is difficult to compute the exact solution of problem (\ref{test46}) in practice. We consider to solve an inexact solution at each iteration of PMM.
In particular, we propose  an inexact version of  PMM algorithm for solving problem (\ref{test46}).
The error criteria of each iteration should satisfy the following condition:
Find $\mathcal{W}^{t+1}\in\mathbb{R}^{n_1\times n_2\times n_3}$ such that
\begin{equation}\label{eq9}
	\mathcal{W}^{t+1}\in \partial Q(\mathcal{X}^{t+1},\mathcal X^t) \ \textup{and} \
	\|\mathcal{W}^{t+1}\|_F\leq \xi\rho\|\mathcal{X}^{t+1}-\mathcal{X}^{t}\|_{F},
\end{equation}
where $\xi\in(0,\frac{1}{2})$ is a constant.

Now, we state the PMM  in Algorithm \ref{alg:algorithm1}.
\begin{algorithm}[htb]
	\caption{A PMM Algorithm for Solving  Problem (\ref{test42})}
	\label{alg:algorithm1}
	\begin{algorithmic}[1]
		\State \textbf{Initialization:} Given
		parameter $\rho,\lambda,\gamma,\beta>0 $ and $\mathcal{W}^{0}$.
		For $t=0,1,2,\ldots$
		\Repeat
		
		Find $\mathcal{W}^{t+1}$ such that
		$\mathcal{W}^{t+1}\in \partial Q(\mathcal{X}^{t+1},\mathcal X^t)$ and
		$ \|\mathcal{W}^{t+1}\|_F\leq \xi\rho\|\mathcal{X}^{t+1}-\mathcal{X}^{t}\|_{F}.
		$
		\Until A stopping condition is satisfied.
	\end{algorithmic}
\end{algorithm}

\begin{remark}
In Algorithm \ref{alg:algorithm1}, we propose an inexact PMM algorithm for solving problem (\ref{test42}), where the condition in (\ref{eq9}) should be satisfied.
In particular, when $\mathcal{X}^{t+1}$ is the optimal solution of (\ref{test46}), we just choose $\mathcal{W}^{t+1}$ to be a zero tensor.
	
\end{remark}

\subsection{Convergence Analysis}

In this subsection,  the global convergence and convergence rate of  Algorithm \ref{alg:algorithm1} are established.
First, the convergent result  of Algorithm \ref{alg:algorithm1} is presented in the following theorem.

\begin{theorem}\label{ConvRe}
Let $\{\mathcal X^t\}$ be the sequence generated by Algorithm \ref{alg:algorithm1}.
Suppose that Assumption \ref{assum5} hold, where  the locally Lipschitz constant of $s_2'$ is set as $L_0$.
Assume that $\nabla f_{n,\mathcal{Y}}$ is Lipschitz continuous with Lipschitz constant $L$, and  $f_{n,\mathcal{Y}}, g_\lambda(x)$ are KL functions.
Then for any $\rho>\frac{L}{1-2\xi}$ with $\xi\in(0,\frac{1}{2})$, the sequence $\{\mathcal X^t\}$ converges to a stationary point  $\tilde{\mathcal X}$  of (\ref{test1}) as $t$ goes to infinity.
\end{theorem}

The proof of Theorem \ref{ConvRe} is left to Appendix G.
	The assumptions in Theorem \ref{ConvRe} are very mild.
	The nonconvex functions in Table \ref{table2} are KL functions \cite{Wen2017, qiu2021nonlocal}.
	Moreover,
	when  $f_{n,\mathcal{Y}}$ is the logistic loss function (e.g., see (\ref{eq61})) or the least squares loss function (e.g., see (\ref{eq30})),
it is a KL function \cite{Wen2017, Bolte2013}.
Besides, when $f_{n,\mathcal{Y}}$  is the least squares loss, the Lipschitz constant of $\nabla f_{n,\mathcal{Y}}$  is $1$. And when $f_{n,\mathcal{Y}}$ is the logistic loss function in (\ref{eq61}), it follows from Lemma \ref{lem22} that
 the Lipschitz constant of $\nabla f_{n,\mathcal{Y}}$  is $\frac{1}{4n} \sum_{i=1}^n\|\mathcal{Z}_i\|_F^2$ .

\begin{remark}
		The assumption about the locally Lipschitz continuity of $s_2'$ is very mild.
		In fact, the derivatives of these functions in Table \ref{table2} are Lipschitz continuous.
In particular,	it is evident from \cite{Wen2017,Ahn2017} that for Logarithm function, the Lipschitz constant of $ s_2'$ is $\frac{\lambda}{\gamma^2}$.
	For SCAD and MCP, the Lipschitz constants of $ s_2'$ are $\frac{1}{\gamma-1}$ and $\frac{1}{\gamma}$, respectively.
\end{remark}

Theorem \ref{ConvRe}
shows that
the sequence $\{\mathcal X^t\}$ generated by Algorithm \ref{alg:algorithm1}
converges to $\tilde{\mathcal X}$ globally as $t$ tends to infinity, i.e. $\lim_{t\rightarrow \infty}\mathcal X^t=\tilde{\mathcal X}$.
Now we also give the convergence rate of Algorithm  \ref{alg:algorithm1}, which is stated in the following theorem.

\begin{theorem}\label{thm3}
Let $\{\mathcal X^t\}$ be the sequence generated by Algorithm \ref{alg:algorithm1}.
Suppose that the assumptions in Theorem \ref{ConvRe} hold  and $H({\mathcal X})$ defined in (\ref{eq3}) satisfies the KL property at
$\tilde{\mathcal X}$ with an exponent $\alpha\in[0,1)$.
Then, we have the following results:\par
(i) If $\alpha=0$, then the sequence $\left\{\mathcal X^t\right\}$ converges in a finite number of steps.\par
(ii) If $0<\alpha \leq \frac{1}{2}$, then the sequence $\left\{\mathcal X^t\right\}$ converges $R$-linearly, i.e., there exist $w>0$ and $\vartheta\in[0,1)$ such that $\|\mathcal X^t-\tilde{\mathcal X}\|_F \leq w\vartheta^t$.\par
(iii) If $\frac{1}{2}<\alpha<1$, then the sequence $\left\{\mathcal X^t\right\}$ converges $R$-sublinearly, i.e., there exists $w>0$ such that $\|\mathcal X^t-\tilde{\mathcal X}\|_F \leq w t^{-\frac{1-\alpha}{2 \alpha-1}}$.
 \end{theorem}

The proof of Theorem \ref{thm3} is left to Appendix H.
In particular, if we know the KL exponent of $H({\mathcal X})$, the detailed convergence rate of PMM can be determined.

\subsection{ADMM  for Solving the Subproblem}

Now we consider to solve the subproblem (\ref{test46}) by applying ADMM \cite{glowinski1975approximation,Fazel2013}.
Let $\mathcal X=\mathcal M$, then problem (\ref{test46}) is equivalent to
\begin{equation}\label{test47}
\begin{aligned}
\min _{\mathcal X,\mathcal M}& \
\beta S_1(\mathcal M)
+\langle \nabla  f_{n,\mathcal{Y}} (\mathcal X^t)-\beta\nabla S_2(\mathcal X^t),\mathcal X-\mathcal X^t \rangle
+
\frac{\rho}{2}\|\mathcal{X}-\mathcal{X}^{t}\|_{F}^{2}
+\delta_D(\mathcal X)\\
\text{s.t.} & \ \mathcal X =\mathcal M.
\end{aligned}
\end{equation}
The augmented Lagrangian function associated with (\ref{test47}) is given by
$$
\begin{aligned}
L(\mathcal X,\mathcal M, \mathcal Z) =
& \ \beta S_1(\mathcal M)
+\langle \nabla  f_{n,\mathcal{Y}} (\mathcal X^t)- \beta \nabla S_2(\mathcal X^t),\mathcal X-\mathcal X^t \rangle
+
\frac{\rho}{2}\|\mathcal{X}-\mathcal{X}^{t}\|_{F}^{2}
+
\delta_D(\mathcal X) \\
& \ + \langle \mathcal Z,\mathcal X-\mathcal M\rangle
+\frac{\eta}{2}\|\mathcal X-\mathcal M\|_{F}^{2},
\end{aligned}
$$
where $\eta>0$ is the penalty parameter and $\mathcal Z\in \mathbb{R}^{n_1\times n_2\times n_3}$ is the Lagrangian  multiplier.
Then the iteration of  ADMM is given as follows:
\begin{align}
\mathcal M^{k+1}&= \mathop{\arg\min}\limits_{\mathcal M} L(\mathcal X^{k},\mathcal M, \mathcal Z^k),\label{test70}\\
\mathcal X^{k+1}&= \mathop{\arg\min}\limits_{\mathcal X}L(\mathcal X,\mathcal M^{k+1}, \mathcal Z^k),\label{test71}\\
\mathcal Z^{k+1}&= \mathcal Z^{k}+\tau\eta(\mathcal X^{k+1}-\mathcal M^{k+1}), \label{test72}
\end{align}
where $\tau\in(0,\frac{1+\sqrt{5}}{2})$ is the step size.

Now we consider to solve the subproblems in (\ref{test70}) and (\ref{test71}) in detail.
Note that problem (\ref{test70}) can be equivalently written as
\begin{equation}\label{Mk1}
\begin{aligned}
	\mathcal M^{k+1}
	&=\mathop{\arg\min}\limits_{\mathcal M}  \beta S_1(\mathcal{M})
	+\frac{\eta}{2}\left\|\mathcal M -\left(\mathcal X^{k}+\frac{1}{\eta}\mathcal Z^{k}\right)\right\|_{F}^{2}\\
	&=\textup{Prox}_{\frac{ \beta }{\eta}S_1}\left(\mathcal X^{k}+\frac{1}{\eta}\mathcal Z^{k}\right).
\end{aligned}
\end{equation}

Problem (\ref{test71}) can be rewritten as
$$
\begin{aligned}
\mathcal X^{k+1}
= \ &\mathop{\arg\min}\limits_{\mathcal X}\delta_D(\mathcal X)+
\langle \nabla  f_{n,\mathcal{Y} }(\mathcal X^t)- \beta \nabla S_2(\mathcal X^t),\mathcal X-\mathcal X^t \rangle
+ \langle \mathcal Z^k,\mathcal X-\mathcal M^{k+1}\rangle\\
& \quad\quad\quad \ \ +\frac{\eta}{2}\|\mathcal X-\mathcal M^{k+1}\|_{F}^{2}
+\frac{\rho}{2}\|\mathcal{X}-\mathcal{X}^{t}\|_{F}^{2}\\
=& \ \mathop{\arg\min}\limits_{\mathcal X}\delta_D(\mathcal X)
+\frac{\rho+\eta}{2}\left\|\mathcal{X}-\frac{1}{\rho+\eta}\mathcal H^{k+1}\right\|_{F}^{2},
\end{aligned}
$$
where $\mathcal H^{k+1}=\rho\mathcal X^t-\nabla  f_{n,\mathcal{Y}}(\mathcal X^t)+ \beta \nabla S_2(\mathcal X^t)+\eta \mathcal M^{k+1}-\mathcal Z^k$.
A simple computation leads to
\begin{equation}\label{test48}
\mathcal X^{k+1}=\mathcal P_{D}\left(\frac{1}{\rho+\eta}\mathcal H^{k+1}\right),
\end{equation}
where $\mathcal P_{D}(\cdot)$ is the projection operator  onto the set $D$ given by
$\mathcal P_{D}(\mathcal Y)=\text{max}\{\text{min}\{\mathcal Y_{ijk},c\},-c\}$.

Then the ADMM for solving  problem (\ref{test47}) is stated in Algorithm \ref{alg:algorithm2}.

\begin{algorithm}[htb]
    \caption{An ADMM Algorithm for Solving   Problem (\ref{test47})}
    \label{alg:algorithm2}
    \begin{algorithmic}[1]
        \State \textbf{Initialization:} Given initial value ${\mathcal X}^0,{\mathcal M}^0,\mathcal Z^0, {\mathcal X}^t$,
         parameters
         $\eta>0, \tau\in(0,\frac{1+\sqrt{5}}{2})$.
        \Repeat
  		\State \textbf{Step 1}. Compute ${\mathcal M}^{k+1}$ by (\ref{Mk1}).
  		\State \textbf{Step 2}. Compute ${\mathcal X}^{k+1}$ by (\ref{test48}).
  		\State \textbf{Step 3}. Update ${\mathcal Z}^{k+1}$ by (\ref{test72}).
        \Until A stopping condition is satisfied.
	\end{algorithmic}
\end{algorithm}

\begin{remark}
In Algorithm \ref{alg:algorithm2}, we need to compute the proximal mapping of $S_1$.
In the experiments, the function $s_1(x)$ is taken as $s_1(x)=\lambda x$ and then $S_1(\mathcal X)=\lambda\|\mathcal X\|_{\textup{TTNN}}$.
As a result, problem (\ref{Mk1}) can be equivalently written as
$$
\begin{aligned}
	\mathcal M^{k+1}
	&=\textup{Prox}_{\frac{ \beta\lambda }{\eta}\|\cdot\|_{\textup{TTNN}}}\left(\mathcal X^{k}+\frac{1}{\eta}\mathcal Z^{k}\right).
\end{aligned}
$$
It follows from \cite[Theorem 3]{Song_2020}  that
$$
	\mathcal M^{k+1}=\mathcal U \diamond_{\mathbf U}\Sigma_{\lambda}\diamond_{\mathbf U}\mathcal V^T,
$$
where $\mathcal X^{k}+\frac{1}{\eta}\mathcal Z^{k}=\mathcal U \diamond_{\mathbf U}\Sigma\diamond_{\mathbf U}\mathcal V^T$, $\Sigma_{\lambda}=\mathbf U^T[\widehat{\Sigma}_{\lambda}]$ and
$
\widehat{\Sigma}_{\lambda}=\max\{\widehat{\Sigma}_{\mathbf U}-\frac{\beta\lambda}{\eta},0\}.
$
\end{remark}

\begin{remark}
For (\ref{test48}), we need to compute $\nabla S_2(\mathcal X^t)$ in order to get $\mathcal H^{k+1}$.
Note that $s_2$ is differentiable.
By utilizing the differentiable case in Lemma \ref{lem3}, we obtain that
$$
\nabla S_2(\mathcal X^t)=\mathcal U^t \diamond_{\mathbf U}\mathcal D^t\diamond_{\mathbf U}(\mathcal V^t)^T,
$$
where $\mathcal X^t=\mathcal U^t \diamond_{\mathbf U}\Sigma^t\diamond_{\mathbf U}(\mathcal V^t)^T$
and $(\widehat{\mathcal D^t})_\mathbf U^{\langle i\rangle}=\textup{Diag}( s_2'((\widehat{\Sigma^t}_\mathbf U^{\langle i\rangle})_{11}), \ldots,  s_2'((\widehat{\Sigma^t}_\mathbf U^{\langle i\rangle})_{mm}))$, $i=1,\ldots, n_3$.
Here $(\widehat{\Sigma^t}_\mathbf U^{\langle i\rangle})_{jj}$ represents the $(j,j)$-th entry of a matrix and $m=\min\{n_1,n_2\}$.
\end{remark}

Note that problem (\ref{test47}) is convex, and
Algorithm \ref{alg:algorithm2} is  just the classical two-block ADMM, whose convergence has been established in  \cite{glowinski1975approximation, Fazel2013}.
For brevity, we omit the details of convergence  of Algorithm \ref{alg:algorithm2} here.

Now we give the  computational cost of ADMM for solving the subproblem (\ref{test46}) based on the nonconvex functions in Table \ref{table2} and the least squares loss or logistic regression, which is the main iteration of PMM.
Note that $S_1(\mathcal X)=\lambda\|\mathcal X\|_{\textup{TTNN}}$ in Table \ref{table2}.
In this case, the computational cost of $\mathcal M^{k+1}$ is $O(n_{(1)}n_{(2)}^2n_3+n_1n_2n_3^2)$ \cite[Section 4.1]{Song_2020}, where $n_{(1)}=\max\{n_1,n_2\}$ and $n_{(2)}=\min\{n_1,n_2\}$.
The main cost of $\mathcal X^{k+1}$ is to compute $\nabla S_2(\mathcal X^t)$, which is $O(n_{(1)}n_{(2)}^2n_3+n_1n_2n_3^2)$ and only computes one time in ADMM.
Therefore, the computational cost of ADMM in each iteration is $O(n_{(1)}n_{(2)}^2n_3+n_1n_2n_3^2)$.
Furthermore, the computational complexity of PMM in each iteration is  $O((n_{(1)}n_{(2)}^2n_3+n_1n_2n_3^2)k_m)$, where $k_m$ represents the number of iterations of ADMM.

\section{Numerical Experiments}\label{NumericEx}

In this section, some experiments are conducted  to demonstrate the effectiveness of the proposed method.
Our model combines the loss function and nonconvex regularization (called LFNR for short),
where the least squares loss and  logistic regression loss are used for the loss function in model (\ref{test1}).
In this case, the corresponding problems are  tensor completion and binary classification, respectively.
For the nonconvex function $g_\lambda(\cdot)$ in model (\ref{test1}), we use the MCP in all experiments for simplicity.
We remark that the  performance of other nonconvex functions is similar to that of MCP.
All experiments are conducted in MATLAB R2020b with an Intel Core i7-10750H 2.6GHz and 16GB RAM.

For the unitary matrix in TTNN, the choice is given as follows:
First, an initial estimator $\mathcal{X}_1$ is obtained by discrete cosine transform in (\ref{test42}).
Afterwards,  we unfold $\mathcal{X}_1$  into a matrix $(\mathcal{X}_1)_{(3)}$ along the third-dimension
and take the SVD of  $(\mathcal{X}_1)_{(3)}$ as  $(\mathcal{X}_1)_{(3)}=\mathbf{U}\Sigma \mathbf{V}^T$.
Then $\mathbf{U}^T$ is the desirable unitary matrix in TTNN.
More details about the choice of unitary transform can be referred to \cite{Song_2020}.

\subsection{Stopping Criterion}

 Algorithm \ref{alg:algorithm1} will be terminated if
$\frac{\|\mathcal X^{t+1}-\mathcal X^{t}\|_F}{\|\mathcal X^{t}\|_F}\leq 5\times 10^{-4}$
or the  number of iterations reaches $100$.

The Karush-Kuhn-Tucker (KKT) condition of (\ref{test47}) is given as follows:
\begin{equation}
\begin{aligned}
&\mathcal Z\in \partial (\beta S_1(\mathcal M)), \
\mathcal M =\mathcal X, \
\rho(\mathcal X^t-\mathcal X)-\nabla f_{n,\mathcal{Y}}(\mathcal X^t)+\beta\nabla S_2(\mathcal X^t)-\mathcal Z\in\partial\delta_D(\mathcal X).\nonumber
\end{aligned}
\end{equation}
The relative KKT residual is employed  to evaluate the accuracy of Algorithm \ref{alg:algorithm2}:
\begin{equation}\label{test54}
\eta_{res}:=\text{max}\{\eta_{e},\eta_{d},\eta_{p}\},
\end{equation}
where
$$
\begin{aligned}
\eta_{e}:=\frac{\|\mathcal M-\mathcal X\|_F}{1+\|\mathcal M\|_F+\|\mathcal X\|_F}, \
\eta_{d}:=\frac{\|\mathcal M-\text{Prox}_{\beta S_1}(\mathcal M+\mathcal Z)\|_F}{1+\|\mathcal M\|_F+\|\mathcal Z\|_F},\\
\eta_{p}:=\frac{\|\mathcal X-\text{Prox}_{\rho^{-1}\delta_D(\cdot)}(\mathcal X^t-\rho^{-1}(\nabla f_{n,\mathcal{Y}}(\mathcal X^t)-\beta\nabla S_2(\mathcal X^t)+\mathcal Z))\|_F}
{1+\|\rho^{-1}\mathcal Z\|_F+\|\mathcal X^t\|_F+\|\rho^{-1}\nabla f_{n,\mathcal{Y}}(\mathcal X^t)\|_F+\|\rho^{-1}\beta\nabla S_2(\mathcal X^t)\|_F}.
\end{aligned}
$$
Then Algorithm \ref{alg:algorithm2} will be terminated if $\eta_{res}\leq 3\times 10^{-3}$
or the number of iterations exceeds 100.
\subsection{Tensor Completion}
In this subsection, we present  numerical experiments for tensor completion, where the least squares loss is utilized for $f_{n,\mathcal{Y}}$.
In this case, model (\ref{test1}) reduces to the least squares loss function with nonconvex regularization given by
\begin{equation}\label{eq30}
	\begin{aligned}
		\min _\mathcal X& \ \frac{1}{2p}\|\mathcal{P}_{\Omega}(\mathcal{X}-\mathcal{Y})\|_F^2 + \beta G_\lambda(\mathcal X) \\
		\text{s.t.} & \  \|\mathcal X\|_{\infty} \leq \  c,
	\end{aligned}
\end{equation}
where  $\mathcal{Y}\in\mathbb{R}^{n_1\times n_2\times n_3}$ is the observed tensor only known its entries in $\Omega$, $\Omega$ is the index set,  $p:=\frac{n}{n_1n_2n_3}$  denotes the probability of each element to be observed, and $\mathcal{P}_{\Omega}$ is the projection operator onto $\Omega$ such that the entry maintaining the same for the index in $\Omega$ and zero outside $\Omega$.
By choosing  the regularization term $G_\lambda(\mathcal X)$
 as TTNN, model (\ref{eq30}) is convex and called TTNN for short.
We also compare with the following three methods: sum of nuclear norms of unfolding matrices of a tensor (SNN) \cite{Gandy_2011}, parallel matrix factorization for tensor completion (TMac)\footnote{\footnotesize \url{https://xu-yangyang.github.io/TMac/}} \cite{Xu2015}, nonconvex regularized tensor algorithm (NORT)\footnote{\footnotesize \url{https://github.com/quanmingyao/FasTer}} \cite{Yao2020}.

\begin{table}[!t]
	\centering
	\caption{PSNR and SSIM values of different methods for the Balloons dataset with different $\sigma$ and SRs.}
	\begin{tabular}{cccccccc}
		\toprule
		& $\sigma$ & SR    & SNN   & TMac  & NORT  & TTNN  & LFNR \\
		\midrule
		\multirow{12}[4]{*}{PSNR} & \multirow{6}[2]{*}{0.005} & 0.05  & 24.50 & \underline{35.23} & 32.79 & 33.91 & \textbf{35.65} \\
		&       & 0.10   & 31.13 & 37.74 & 37.22 & \underline{38.75} & \textbf{40.01} \\
		&       & 0.15  & 34.41 & 39.05 & 39.44 & \underline{41.76} & \textbf{42.80} \\
		&       & 0.20   & 36.76 & 40.20 & 40.38 & \underline{43.72} & \textbf{44.57} \\
		&       & 0.25  & 38.57 & 41.25 & 41.59 & \underline{45.46} & \textbf{46.07} \\
		&       & 0.30   & 39.82 & 41.88 & 42.48 & \underline{46.43} & \textbf{47.21} \\
		\cmidrule{2-8}          & \multirow{6}[2]{*}{0.01} & 0.05  & 23.96 & 33.41 & \underline{33.60} & 33.24 & \textbf{35.15} \\
		&       & 0.10   & 30.50 & 35.96 & 37.22 & \underline{37.56} & \textbf{38.78} \\
		&       & 0.15  & 33.38 & 37.65 & 38.79 & \underline{39.24} & \textbf{41.08} \\
		&       & 0.20   & 35.36 & 39.17 & 39.26 & \underline{41.10} & \textbf{42.47} \\
		&       & 0.25  & 36.69 & 40.26 & 39.96 & \underline{42.35} & \textbf{43.51} \\
		&       & 0.30   & 37.73 & 40.83 & 40.96 & \underline{43.32} & \textbf{44.34} \\
		\midrule
		\multirow{12}[4]{*}{SSIM} & \multirow{6}[2]{*}{0.005} & 0.05  & 0.8342 & 0.9053 & 0.8989 & \underline{0.9057} & \textbf{0.9281} \\
		&       & 0.10   & 0.9211 & 0.9447 & 0.9417 & \underline{0.9615} & \textbf{0.9693} \\
		&       & 0.15  & 0.9493 & 0.9604 & 0.9655 & \underline{0.9768} & \textbf{0.9819} \\
		&       & 0.20   & 0.9640 & 0.9692 & 0.9715 & \underline{0.9838} & \textbf{0.9868} \\
		&       & 0.25  & 0.9720 & 0.9747 & 0.9776 & \underline{0.9875} & \textbf{0.9899} \\
		&       & 0.30   & 0.9763 & 0.9777 & 0.9810 & \underline{0.9898} & \textbf{0.9913} \\
		\cmidrule{2-8}          & \multirow{6}[2]{*}{0.01} & 0.05  & 0.8218 & 0.8385 & \underline{0.9015} & 0.8854 & \textbf{0.9058} \\
		&       & 0.10   & 0.9029 & 0.9064 & 0.9417 & \underline{0.9421} & \textbf{0.9541} \\
		&       & 0.15  & 0.9300 & 0.9384 & 0.9577 & \underline{0.9593} & \textbf{0.9661} \\
		&       & 0.20   & 0.9417 & 0.9545 & 0.9612 & \underline{0.9640} & \textbf{0.9756} \\
		&       & 0.25  & 0.9469 & 0.9621 & 0.9615 & \underline{0.9752} & \textbf{0.9790} \\
		&       & 0.30   & 0.9493 & 0.9667 & 0.9721 & \underline{0.9787} & \textbf{0.9828} \\
		\bottomrule
	\end{tabular}%
	\label{tab:addlabel}%
\end{table}%

\begin{table}[!t]
	\centering
	\caption{PSNR and SSIM values of different methods for the Lemons dataset with different $\sigma$ and SRs.}	\label{tab:addlabe2}%
	\begin{tabular}{cccccccc}
		\toprule
		& $\sigma$ & SR    & SNN   & TMac  & NORT  & TTNN  & LFNR \\
		\midrule
		\multirow{12}[4]{*}{PSNR} & \multirow{6}[2]{*}{0.005} & 0.05  & 30.17 & 37.76 & 36.38 & \underline{37.81} & \textbf{39.98} \\
		&       & 0.10   & 35.60 & 40.46 & 39.53 & \underline{42.24} & \textbf{44.25} \\
		&       & 0.15  & 38.34 & 42.24 & 42.68 & \underline{44.68} & \textbf{46.69} \\
		&       & 0.20   & 40.19 & 43.56 & 44.24 & \underline{46.37} & \textbf{47.90} \\
		&       & 0.25  & 41.67 & 44.60 & 44.92 & \underline{47.68} & \textbf{49.15} \\
		&       & 0.30   & 42.87 & 45.46 & 45.67 & \underline{48.76} & \textbf{50.14} \\
		\cmidrule{2-8}          & \multirow{6}[2]{*}{0.01} & 0.05  & 30.04 & 36.62 & 37.07 & \underline{37.33} & \textbf{38.99} \\
		&       & 0.10   & 34.54 & 39.93 & 40.57 & \underline{41.25} & \textbf{42.64} \\
		&       & 0.15  & 36.81 & 41.44 & 41.69 & \underline{43.27} & \textbf{44.44} \\
		&       & 0.20   & 38.34 & 42.58 & 42.21 & \underline{44.53} & \textbf{45.61} \\
		&       & 0.25  & 39.29 & 43.08 & 42.87 & \underline{44.70} & \textbf{46.40} \\
		&       & 0.30   & 40.02 & 43.45 & 43.66 & \underline{44.90} & \textbf{47.00} \\
		\midrule
		\multirow{12}[3]{*}{SSIM} & \multirow{6}[2]{*}{0.005} & 0.05  & 0.8775 & 0.9194 & 0.9137 & \underline{0.9265} & \textbf{0.9554} \\
		&       & 0.10   & 0.9410 & 0.9541 & 0.9529 & \underline{0.9697} & \textbf{0.9804} \\
		&       & 0.15  & 0.9608 & 0.9703 & 0.9742 & \underline{0.9819} & \textbf{0.9877} \\
		&       & 0.20   & 0.9699 & 0.9783 & 0.9806 & \underline{0.9869} & \textbf{0.9903} \\
		&       & 0.25  & 0.9756 & 0.9820 & 0.9830 & \underline{0.9898} & \textbf{0.9925} \\
		&       & 0.30   & 0.9787 & 0.9845 & 0.9852 & \underline{0.9916} & \textbf{0.9937} \\
		\cmidrule{2-8}          & \multirow{6}[1]{*}{0.01} & 0.05  & 0.8760 & 0.9004 & 0.9210 & \underline{0.9236} & \textbf{0.9383} \\
		&       & 0.10   & 0.9231 & 0.9501 & 0.9591 & \underline{0.9600} & \textbf{0.9674} \\
		&       & 0.15  & 0.9390 & 0.9626 & 0.9657 & \underline{0.9726} & \textbf{0.9766} \\
		&       & 0.20   & 0.9464 & 0.9691 & 0.9687 & \underline{0.9773} & \textbf{0.9810} \\
		&       & 0.25  & 0.9490 & 0.9717 & 0.9710 & \underline{0.9757} & \textbf{0.9845} \\
		&       & 0.30   & 0.9498 & 0.9735 & 0.9738 & \underline{0.9800} & \textbf{0.9867} \\
		\bottomrule
	\end{tabular}%
\end{table}%

\begin{figure}[!t]
	\centering
	\subfigure[Balloons]{
		\begin{minipage}[t]{0.46\textwidth}
			\centering
			\raisebox{0.1cm}{\includegraphics[width=7.5cm,height=5.5cm]{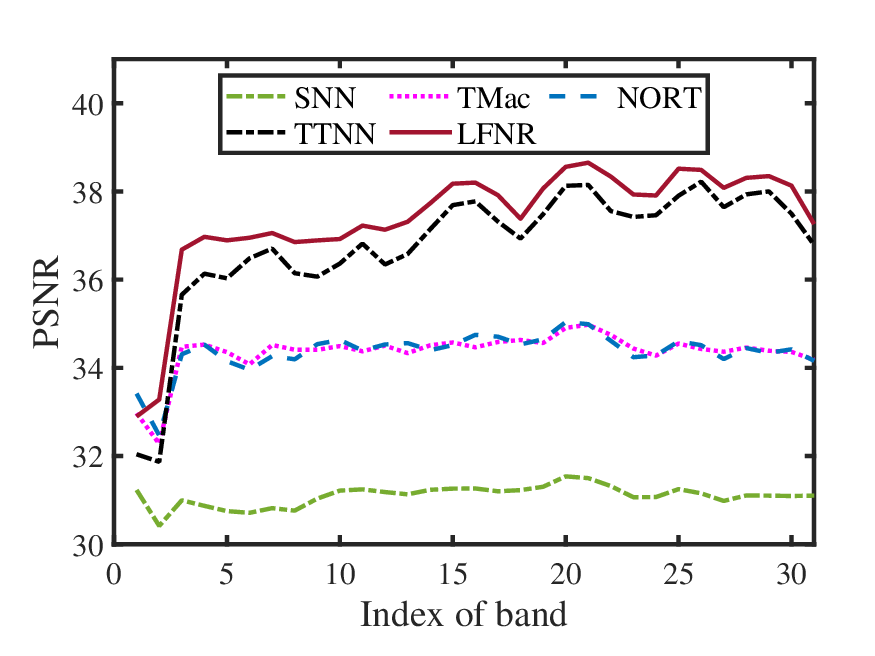}}
		\end{minipage}
	}\hspace{3mm}
	\subfigure[Lemons]{
		\begin{minipage}[t]{0.46\textwidth}
			\centering
			\raisebox{0.1cm}{\includegraphics[width=7.5cm,height=5.5cm]{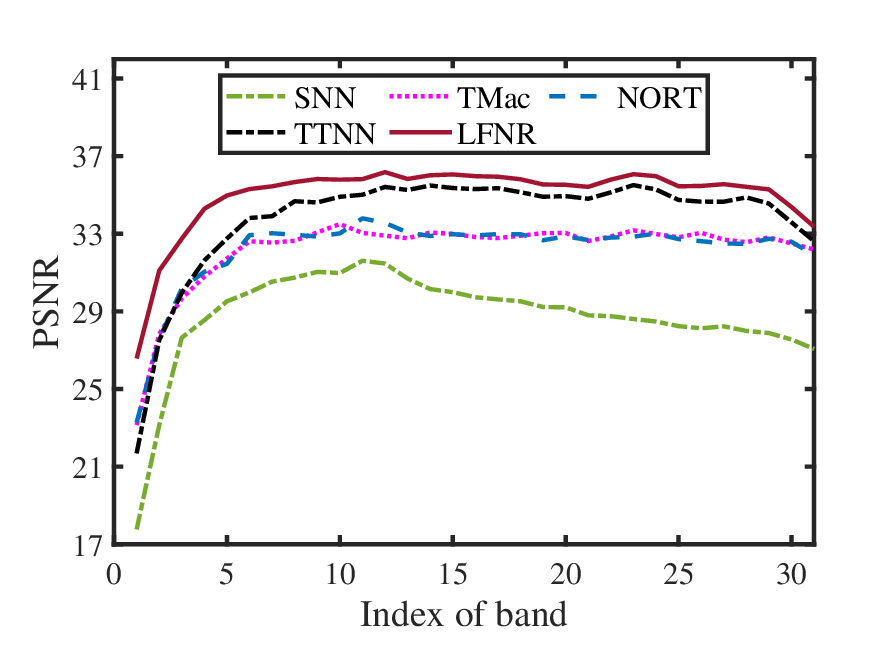}}
		\end{minipage}
	}
	\caption{PSNR values versus index of band of different methods for the Balloons and Lemons
		datasets. (a) Balloons dataset, where $\textup{SR} = 0.5$ and $\sigma=0.05$. (b) Lemons
		dataset, where  $\textup{SR}= 0.4$ and $\sigma=0.1$.
	}\label{Fig: multispectral}	
\end{figure}

\begin{figure}[!t]
	\centering
	\subfigure[Original]{
		\begin{minipage}[t]{0.2\textwidth}
			\centering
			\raisebox{0.1cm}{\includegraphics[width=3.5cm,height=3.5cm]{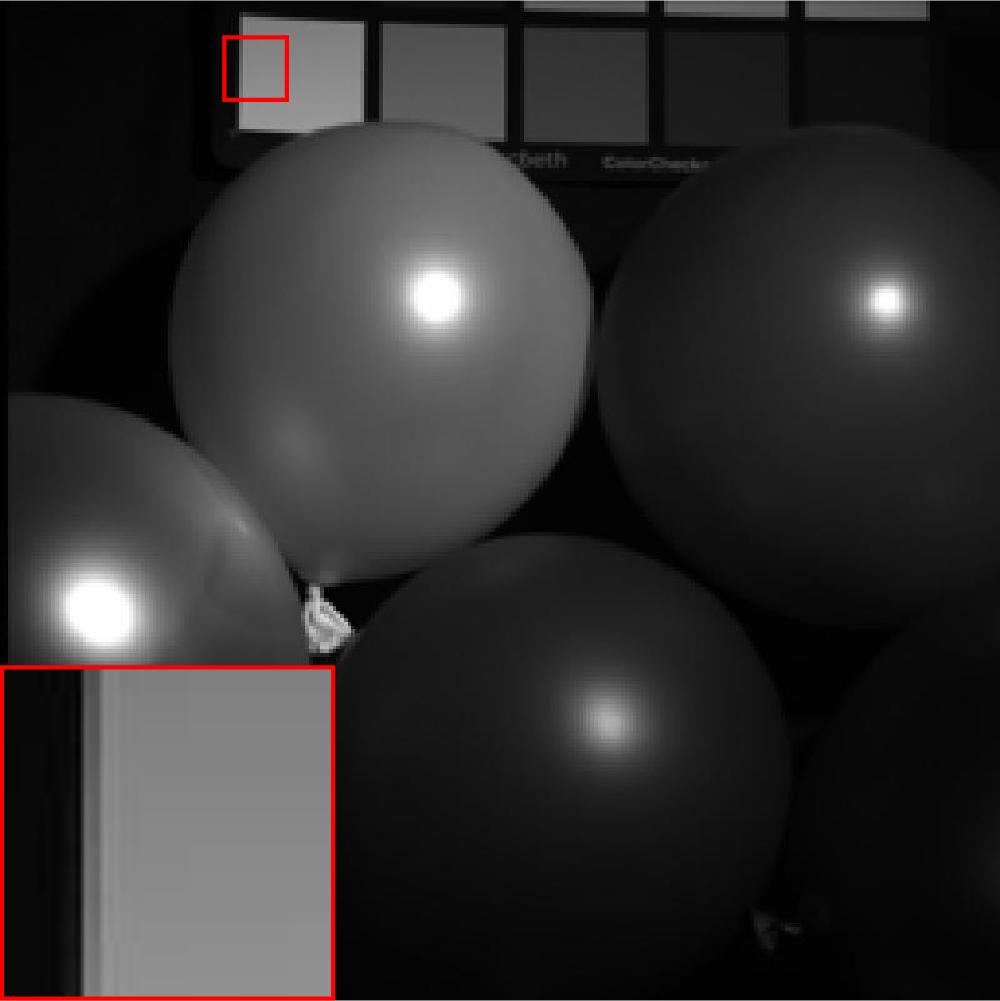}}
		\end{minipage}
	}\hspace{3mm}
	\subfigure[Observation]{
		\begin{minipage}[t]{0.2\textwidth}
			\centering
			\raisebox{0.1cm}{\includegraphics[width=3.5cm,height=3.5cm]{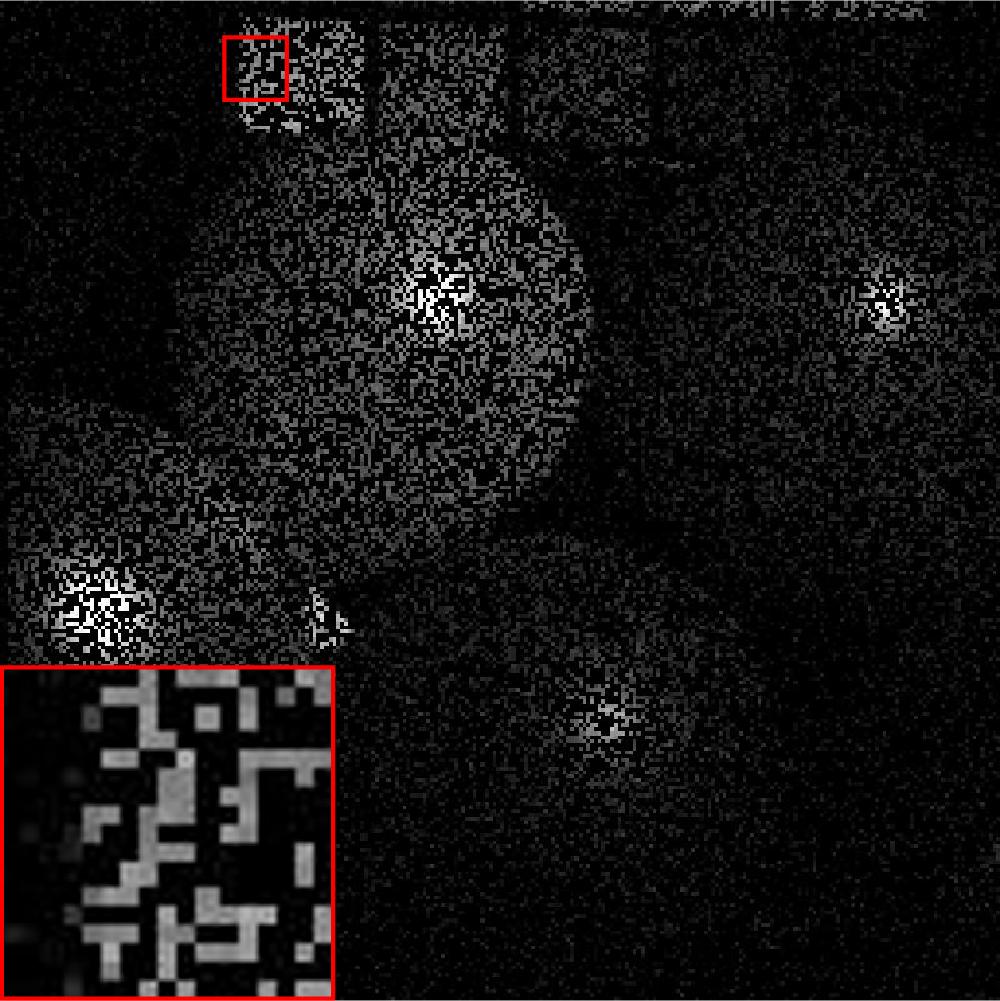}}
		\end{minipage}
	}\hspace{3mm}
	\subfigure[SNN: 30.48]{
		\begin{minipage}[t]{0.2\textwidth}
			\centering
			\raisebox{0.1cm}{\includegraphics[width=3.5cm,height=3.5cm]{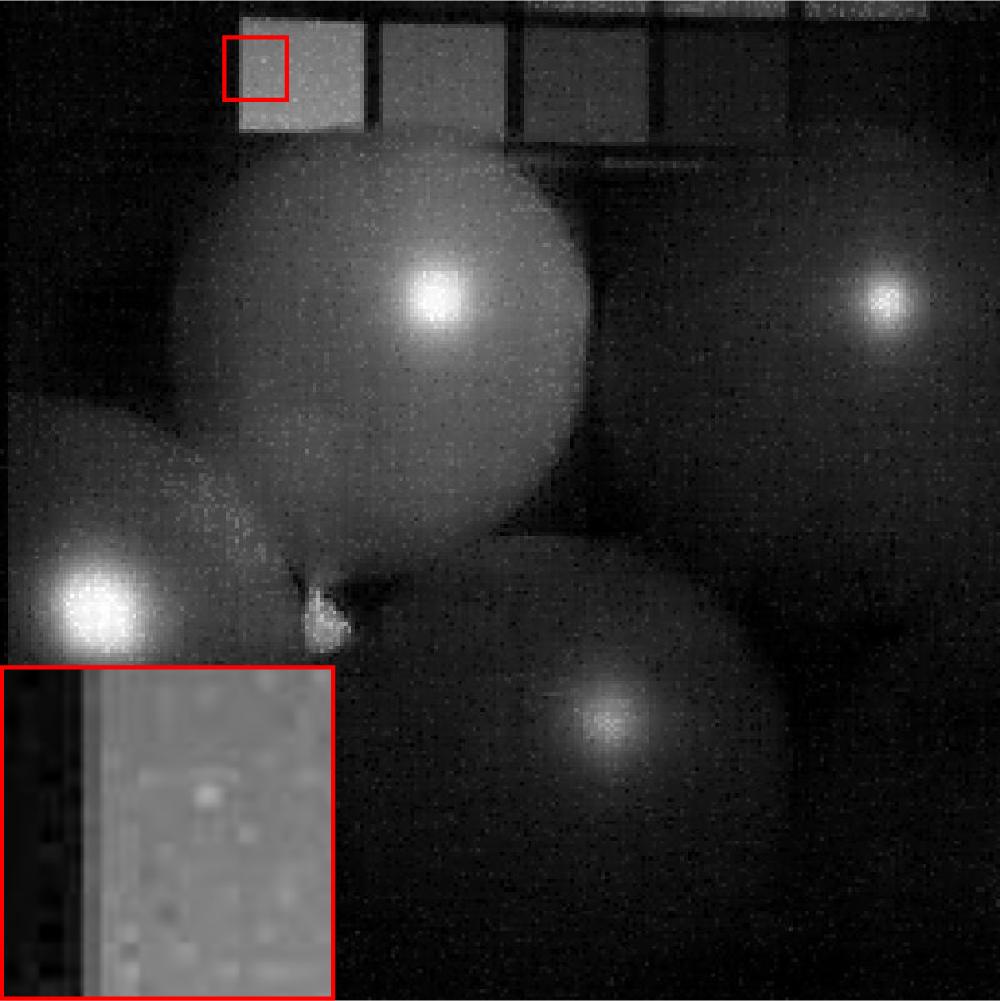}}
		\end{minipage}
	}\hspace{3mm}
	\subfigure[TMac: 34.24]{
		\begin{minipage}[t]{0.2\textwidth}
			\centering
			\raisebox{0.1cm}{\includegraphics[width=3.5cm,height=3.5cm]{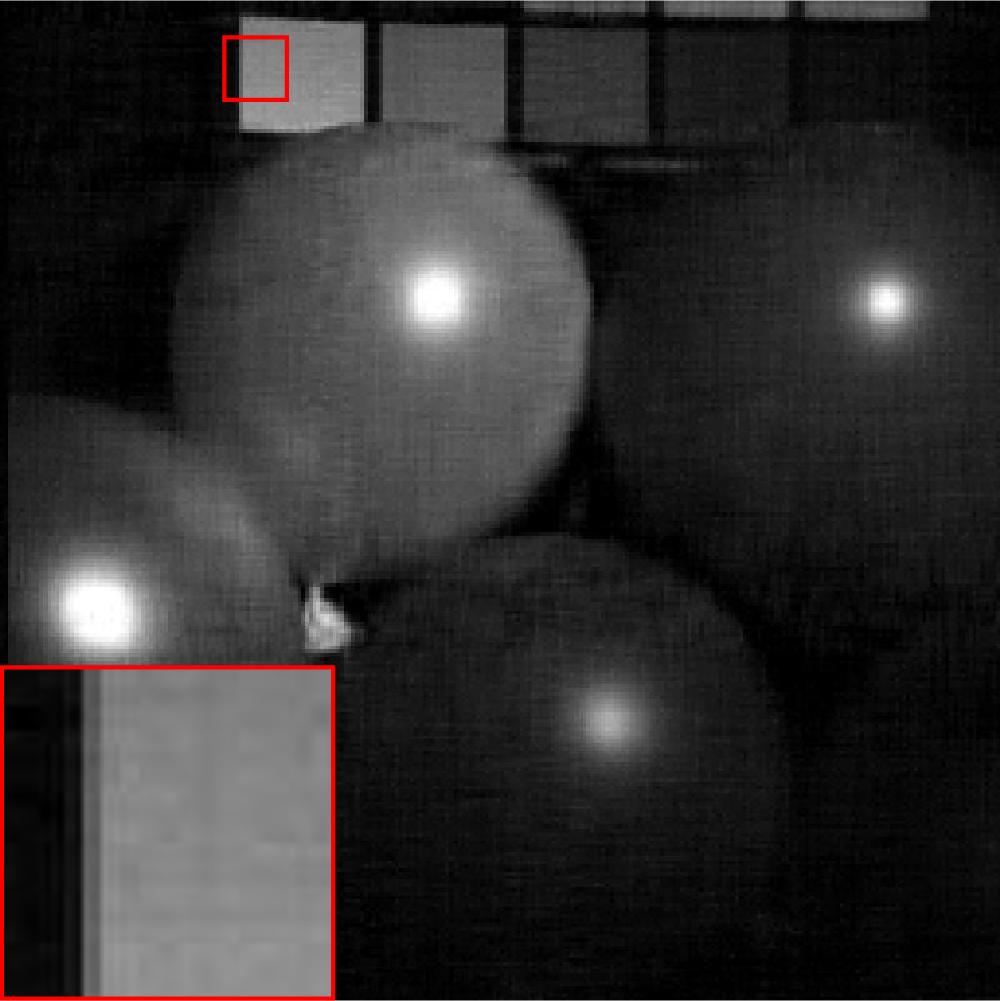}}
		\end{minipage}
	}\hspace{3mm}
	\subfigure[NORT: 34.51]{
		\begin{minipage}[t]{0.2\textwidth}
			\centering
			\raisebox{0.1cm}{\includegraphics[width=3.5cm,height=3.5cm]{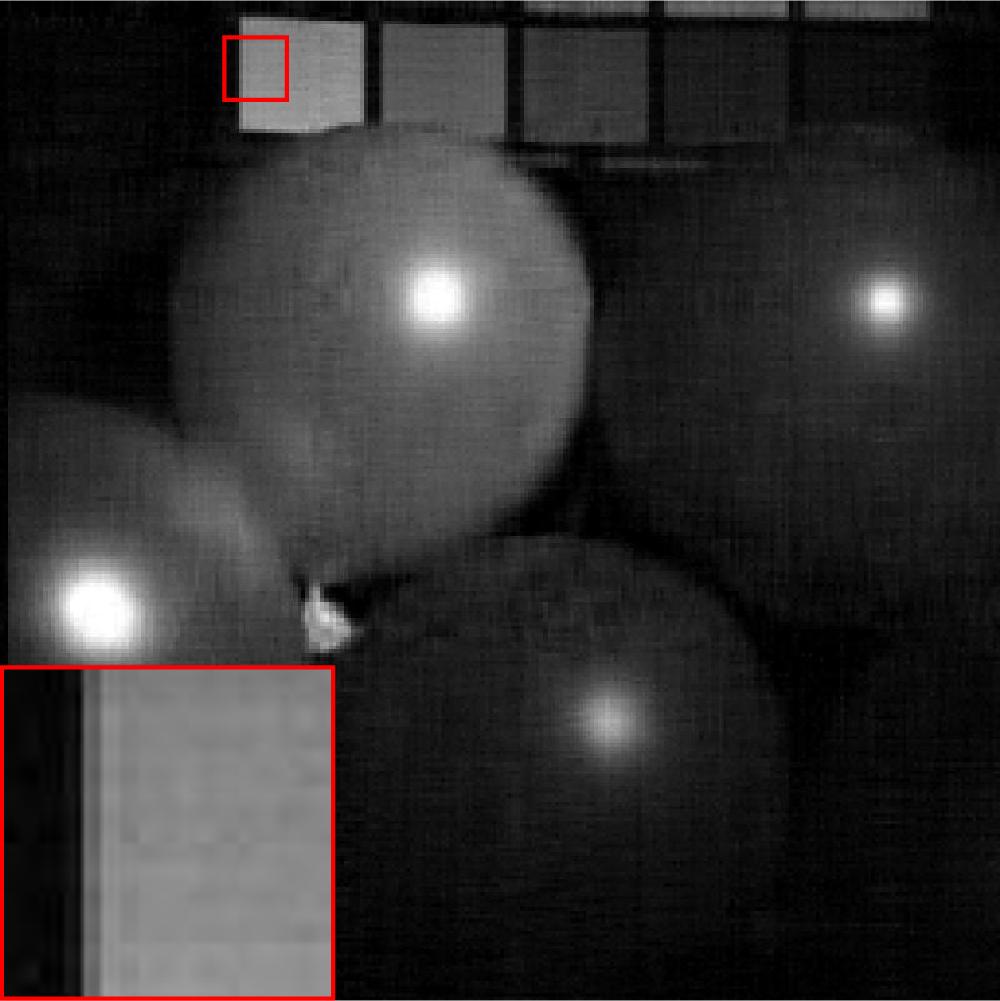}}
		\end{minipage}
	}\hspace{3mm}
	\subfigure[TTNN: 35.57]{
		\begin{minipage}[t]{0.2\textwidth}
			\centering
			\raisebox{0.1cm}{\includegraphics[width=3.5cm,height=3.5cm]{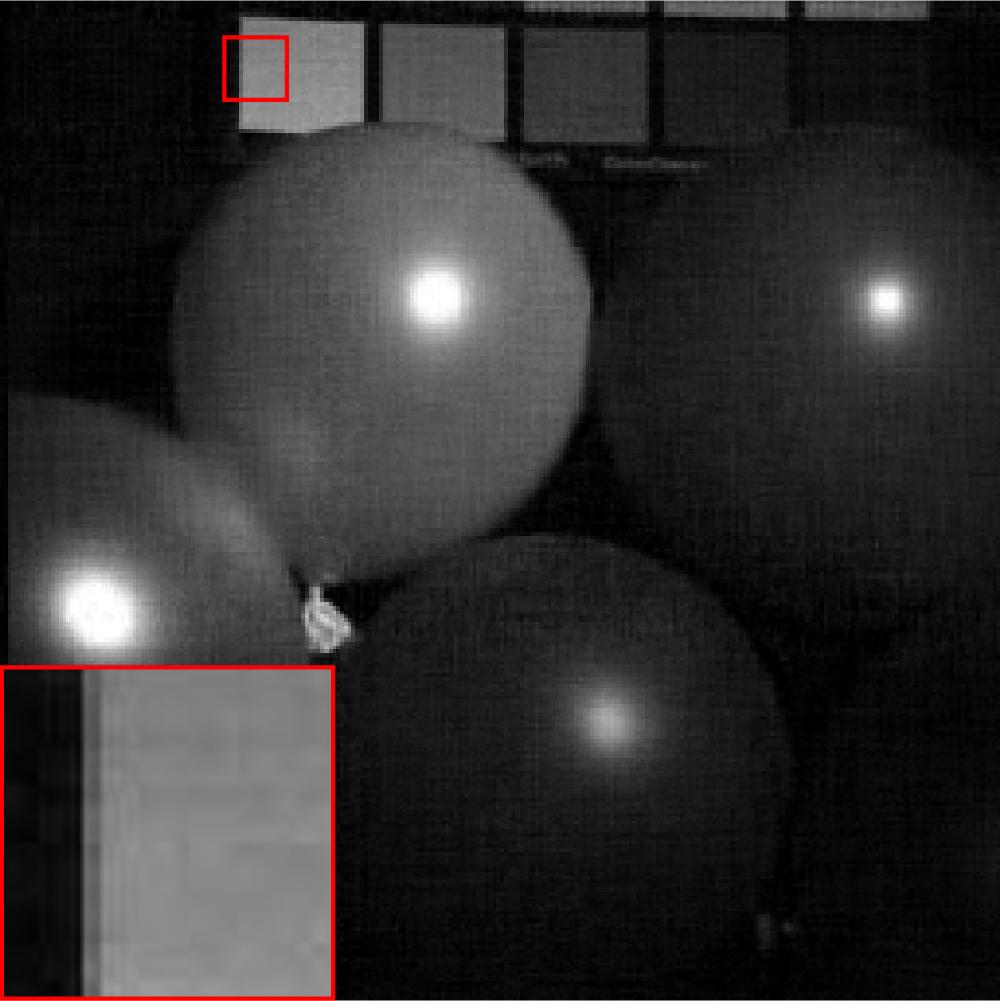}}
		\end{minipage}
	}\hspace{3mm}
	\subfigure[LFNR: 37.24]{
		\begin{minipage}[t]{0.2\textwidth}
			\centering
			\raisebox{0.1cm}{\includegraphics[width=3.5cm,height=3.5cm]{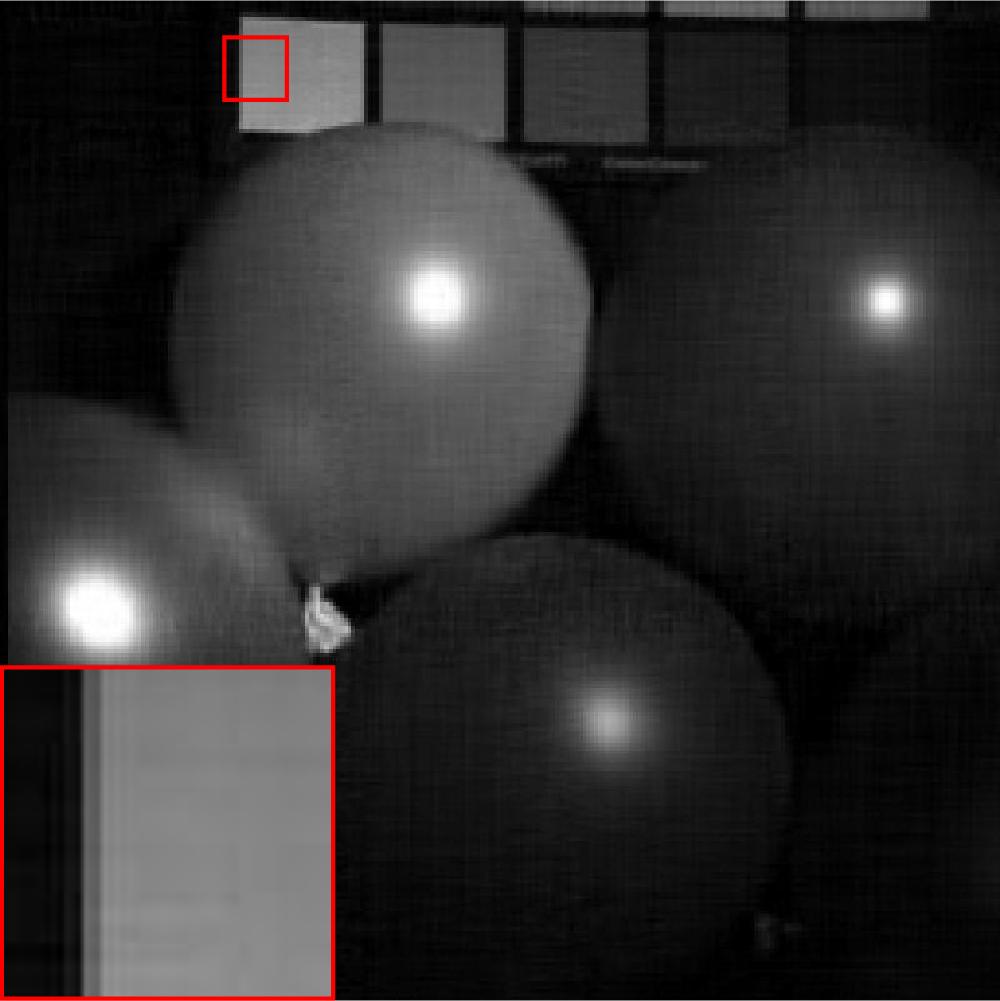}}
		\end{minipage}
	}
	\caption{The recovered images (with PSNR values) and zoomed regions  of different methods for the 30th band of the Balloons dataset, where  $\textup{SR}= 0.4$ and $\sigma=0.05$.
	}\label{Fig: Balloons}	
\end{figure}

\begin{figure}[!t]
\centering
	\subfigure[Original]{
		\begin{minipage}[t]{0.2\textwidth}
			\centering
		\raisebox{0.1cm}{\includegraphics[width=3.5cm,height=3.5cm]{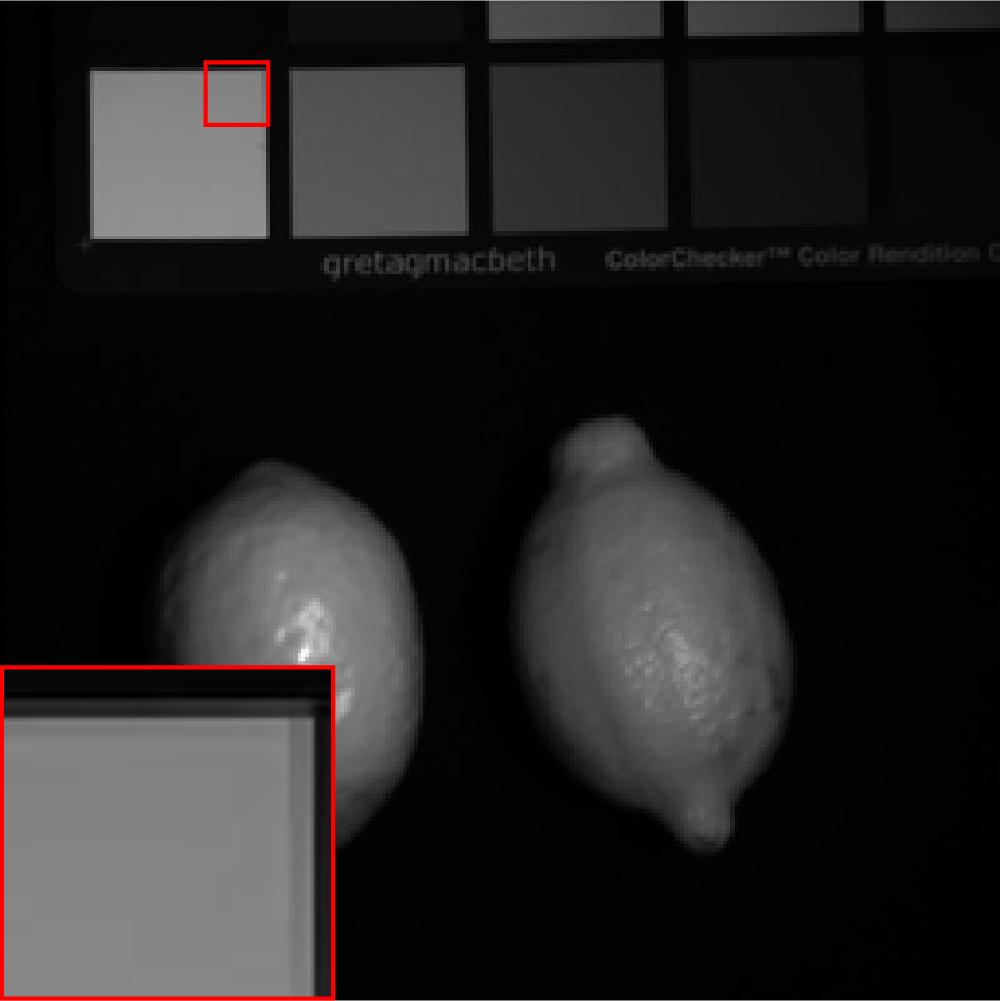}}
		\end{minipage}
	}\hspace{3mm}
	\subfigure[Observation]{
		\begin{minipage}[t]{0.2\textwidth}
			\centering
		\raisebox{0.1cm}{\includegraphics[width=3.5cm,height=3.5cm]{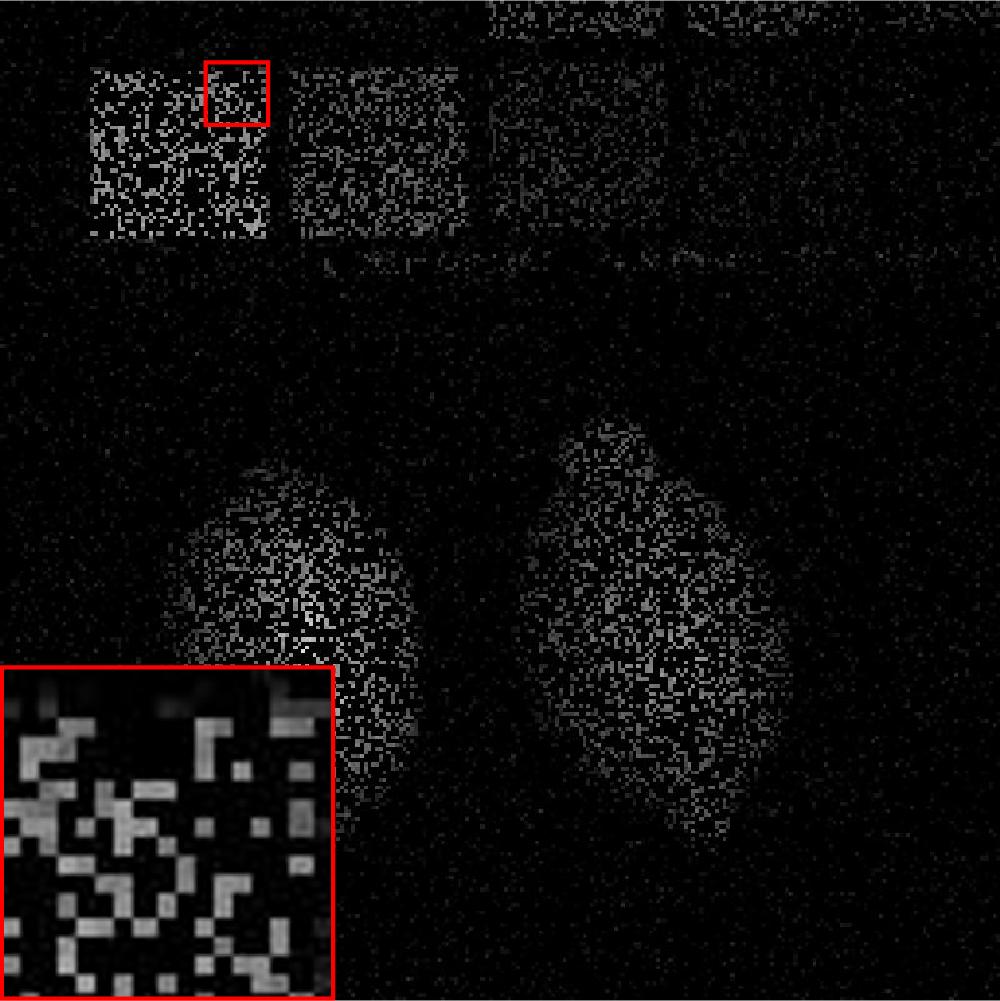}}
		\end{minipage}
	}\hspace{3mm}
   \subfigure[SNN: 29.86]{
		\begin{minipage}[t]{0.2\textwidth}
			\centering
			\raisebox{0.1cm}{\includegraphics[width=3.5cm,height=3.5cm]{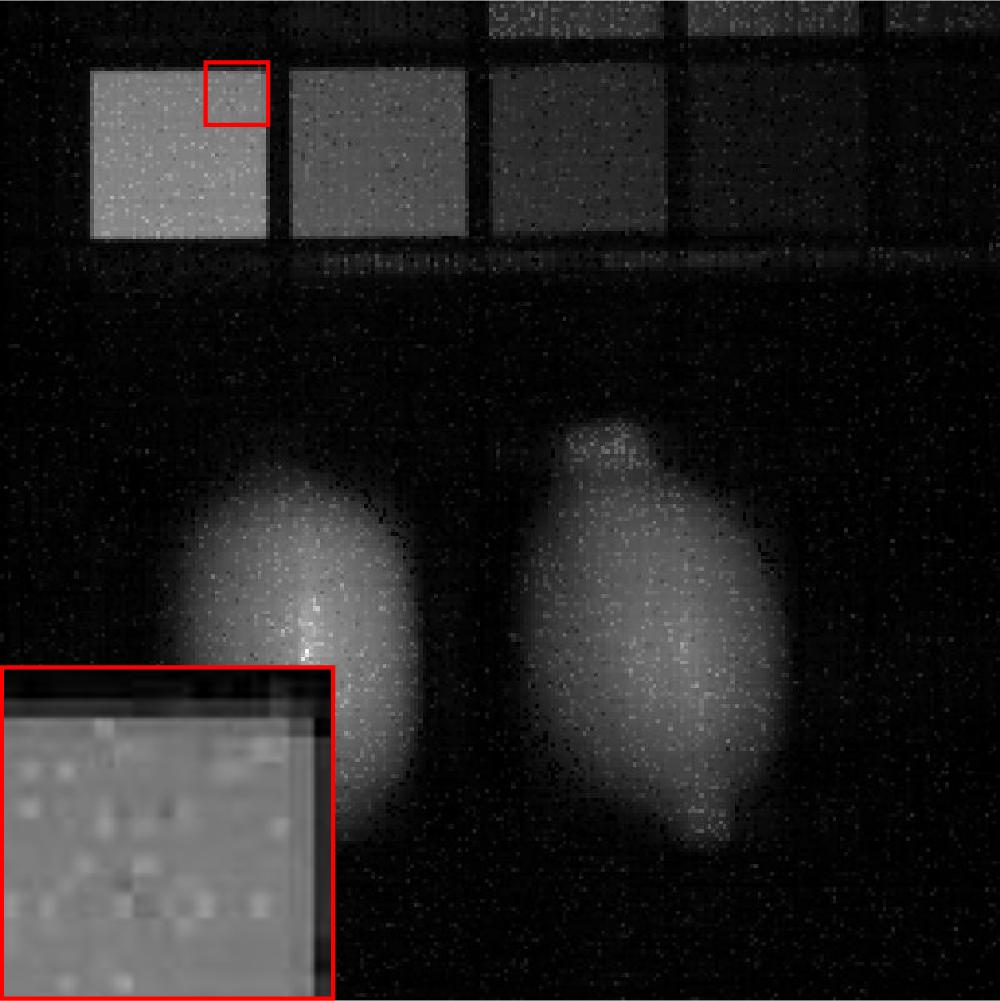}}
		\end{minipage}
	}\hspace{3mm}
    \subfigure[TMac: 36.36]{
		\begin{minipage}[t]{0.2\textwidth}
			\centering
			\raisebox{0.1cm}{\includegraphics[width=3.5cm,height=3.5cm]{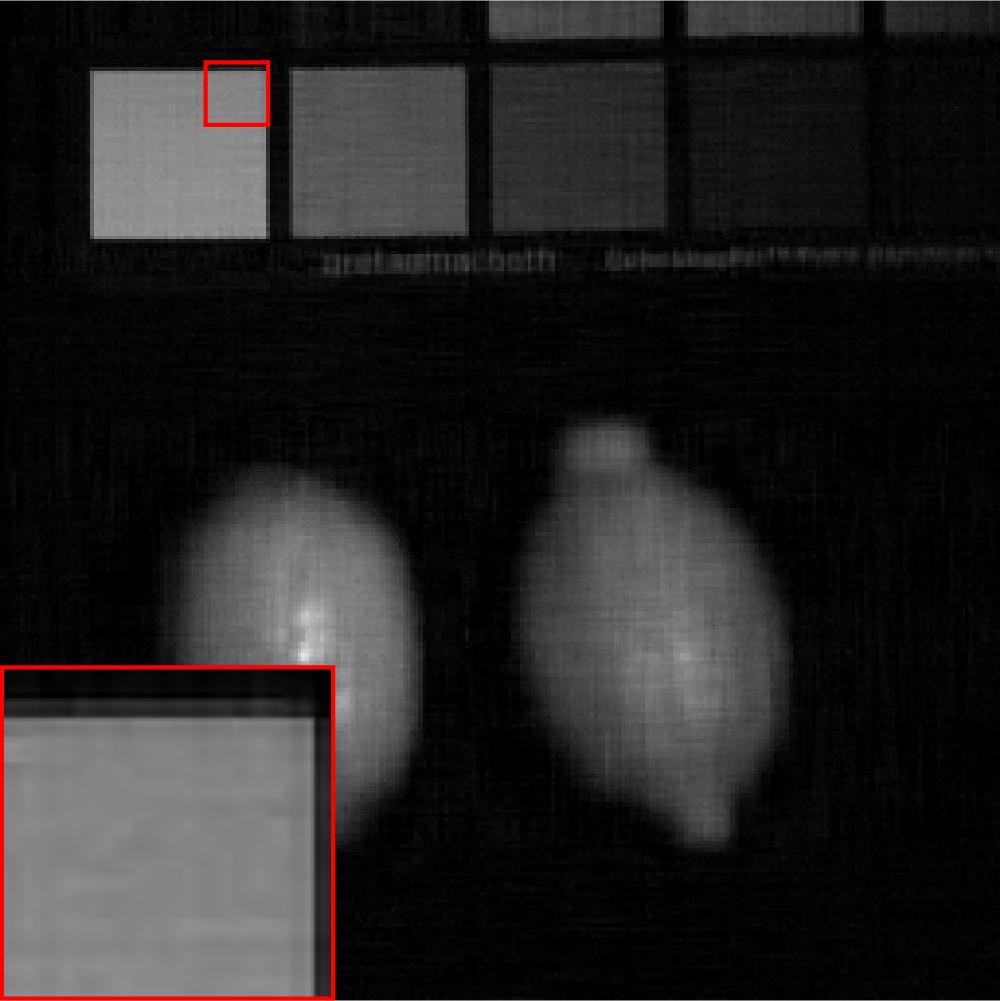}}
		\end{minipage}
	}\hspace{3mm}
     \subfigure[NORT: 36.29]{
		\begin{minipage}[t]{0.2\textwidth}
			\centering
			\raisebox{0.1cm}{\includegraphics[width=3.5cm,height=3.5cm]{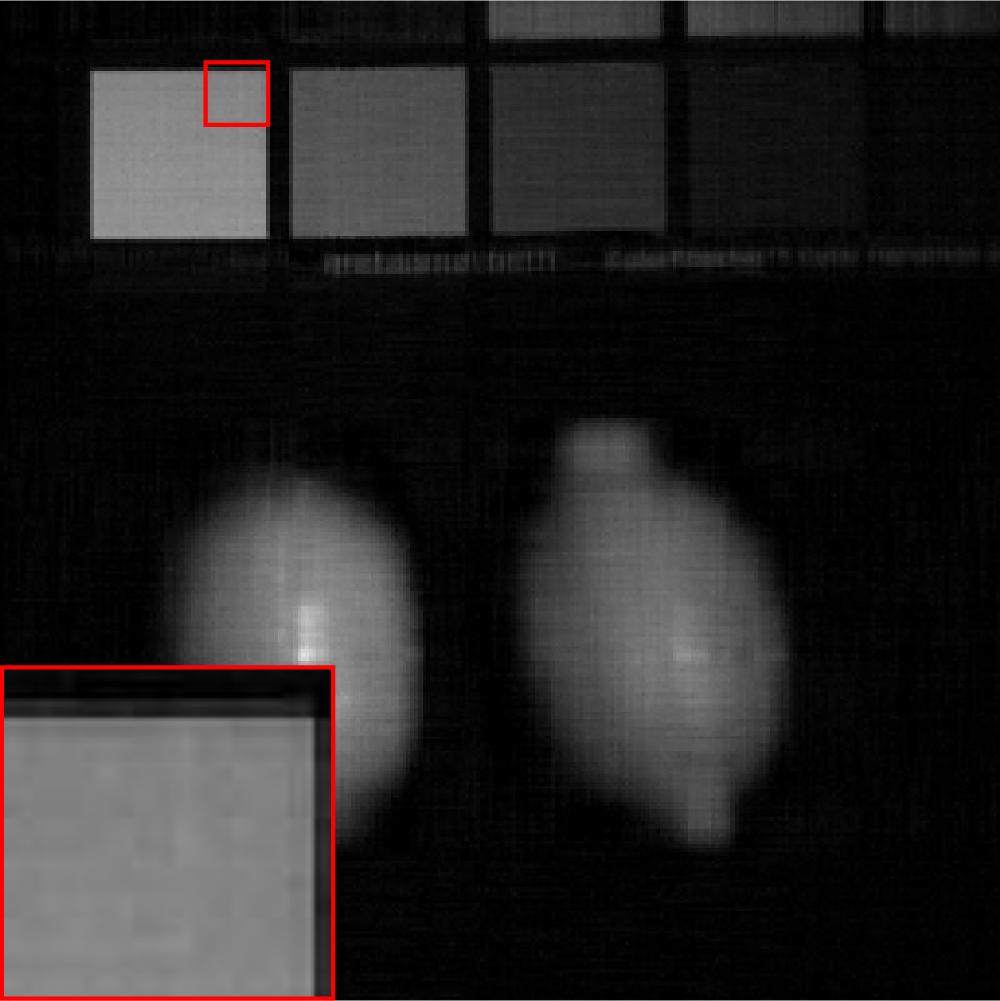}}
		\end{minipage}
	}\hspace{3mm}
	 \subfigure[TTNN: 37.58]{
		\begin{minipage}[t]{0.2\textwidth}
			\centering
			\raisebox{0.1cm}{\includegraphics[width=3.5cm,height=3.5cm]{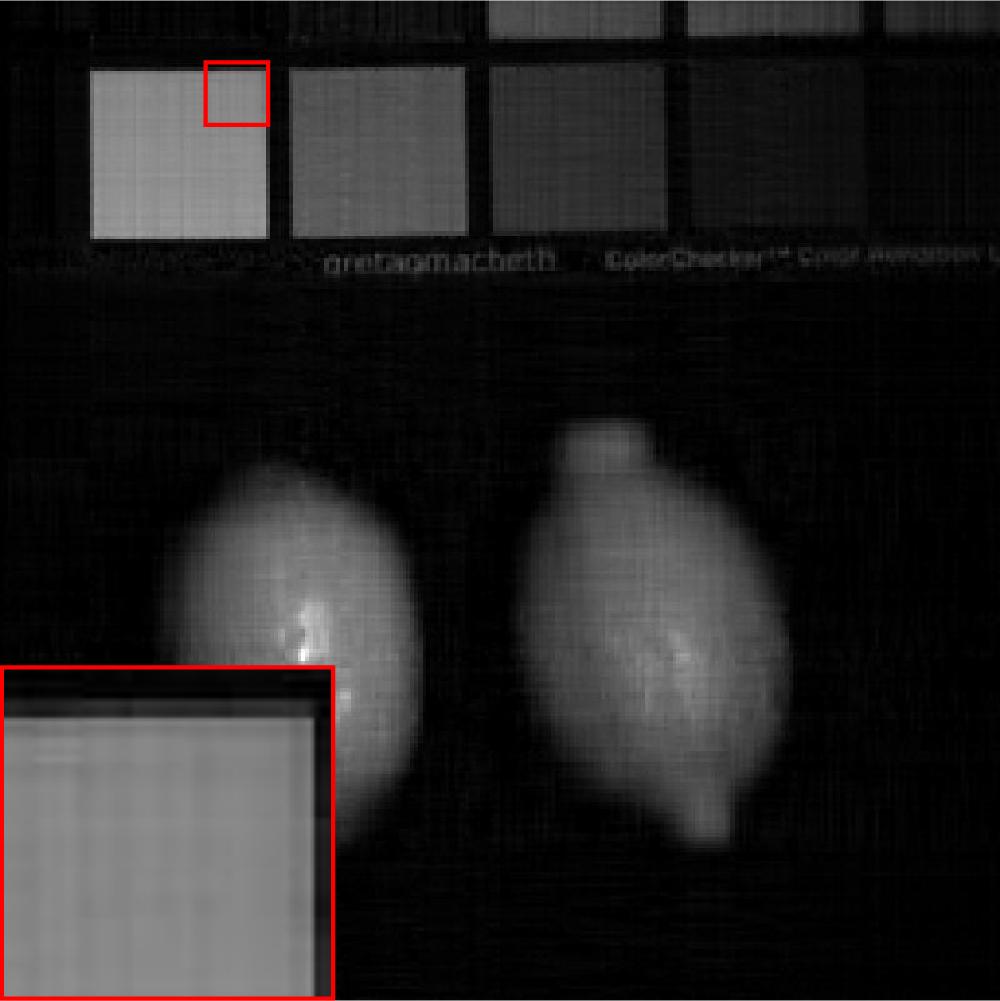}}
		\end{minipage}
	}\hspace{3mm}
	\subfigure[LFNR: 38.61]{
		\begin{minipage}[t]{0.2\textwidth}
			\centering
			\raisebox{0.1cm}{\includegraphics[width=3.5cm,height=3.5cm]{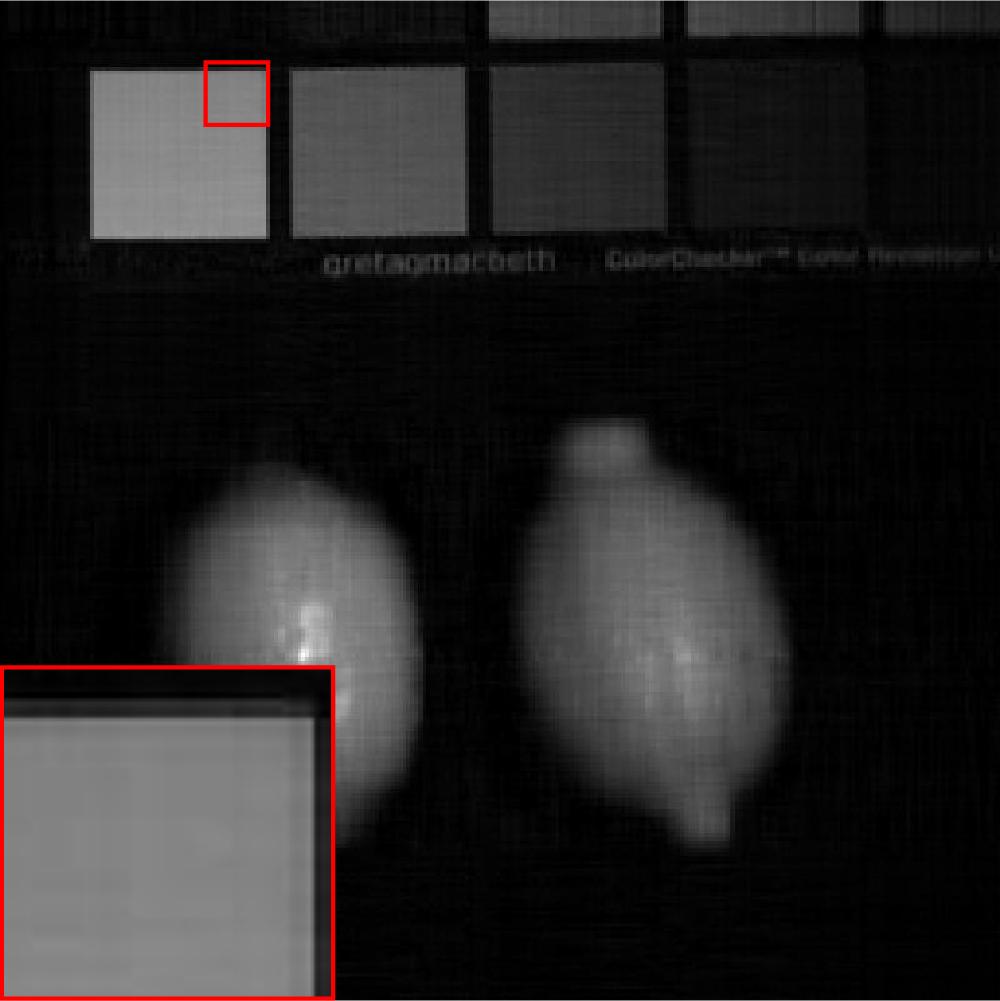}}
		\end{minipage}
	}
	\caption{The recovered images  (with PSNR values) and zoomed regions of different methods for the 30th band of the Lemons dataset,  where $\textup{SR}= 0.3$ and $\sigma=0.05$.}
	\label{Fig: lemons}	
\end{figure}

The observed tensor is constructed as follows: For an $n_1\times n_2 \times n_3$ tensor $\mathcal{X}$, we first add the zero mean Gaussian noise with standard deviation $\sigma$, which is denoted by $\mathcal{Y}$. Then the index set $\Omega$ is uniformly generated at random  and we  get  $\mathcal{P}_{\Omega}(\mathcal{Y})$, where the  sample ratio (SR) is defined as $\textup{SR}:=\frac{|\Omega|}{n_1 n_2  n_3}$.
Here $|\Omega|$ denotes the cardinality of $\Omega$.

The peak signal-to-noise ratio (PSNR) and structural similarity (SSIM) index \cite{Wang2004}
are adopted to measure the recovery performance for real-world data.
Specifically, the PSNR is defined as
$$
\text { PSNR } =10 \log _{10} \frac{n_1 n_2 n_3(\mathcal{X}_{\operatorname{max}}-\mathcal{X}_{\operatorname{min}})^2}
{\|\widetilde{\mathcal{X}}-\mathcal{X}\|_F^2},
$$
where $\mathcal{X}_{\operatorname{max}}$ and $\mathcal{X}_{\operatorname{min}}$ denote the maximum and minimum entries of $\mathcal{X}$, respectively, and $\widetilde{\mathcal{X}}$ and $\mathcal{X}$ are the recovered and underlying tensors, respectively. The SSIM is defined as
$$
\mathrm{SSIM} =\frac{\left(2 \mu_{x} \mu_{y}+c_1\right)\left(2 \sigma_{x y}+ c_2 \right )}{\left(\mu_{x}^2+\mu_{y}^2+c_1\right)+\left(\sigma_{x}^2+\sigma_{y}^2+c_2\right)},
$$
where $\mu_{x}$ and  $\sigma_{x}$ represent the mean intensity and standard deviation of the original image, respectively, $\mu_{y}$ and $\sigma_{y}$ represent the mean intensity and standard deviation of the recovered image, respectively, $\sigma_{x y}$ denotes the covariance between the original and recovered images, and $c_1,c_2>0$ are constants.
For multi-dimensional images, the SSIM denotes the mean value of SSIM   of all images.

\subsubsection{Parameter Settings}

For the parameters $\eta$ and $\tau$ in Algorithm \ref{alg:algorithm2},
we set $\eta=10$ and $\tau=1.618$ in the experiments for simplicity.
For the parameter $\rho$ in (\ref{test46}), it is set to $10$.
$\beta$ is sensitive to the results for different cases and we  select it from the set
$\{0.2,0.3,0.4,0.6,0.7,0.8,0.9,1.2,1.3,1.4,1.5,2\}$ to obtain the best recovery performance for tensor completion.
For the two parameters $\gamma$ and $\lambda$ in MCP, we simply set $\gamma=2.7$ and  choose $\lambda$ from
the set
$\{0.2,0.3,0.4,0.5,0.6,0.7,0.8,0.9,1.1,1.2,1.5,1.6,1.8,2,2.5\}$
to achieve the best performance in the testing cases.

\subsubsection{Multispectral Images}

For tensor completion, we test two multispectral images to demonstrate the effectiveness of the proposed method, including  Balloons and Lemons\footnote{\url{https://www.cs.columbia.edu/CAVE/databases/multispectral/}}, whose sizes are both $256\times256\times31$.
In Tables \ref{tab:addlabel} and \ref{tab:addlabe2}, we
show the PSNR and SSIM values of different methods for the Balloons and Lemons datasets, respectively, where $\sigma=0.005, 0.01$ and $\textup{SR}=0.05,0.10,0.15,0.20,0.25,0.30$.
The best results are highlighted in bold and the second best results are highlighted in underline.
We can see that the PSNR and SSIM obtained by LFNR are higher than those obtained by other methods.
In particular, the LFNR outperforms NORT in terms of PSNR and SSIM values,
which demonstrates that the nonconvex regularization based on TTNN is better than that based on the sum of nuclear norms of unfolding matrices of a tensor.
Besides, the TTNN performs better than SNN, TMac, and NORT for most cases, especially for $\sigma=0.01$.

\begin{table}[!t]
	\centering
	\caption{PSNR and SSIM values of different methods for the MRI dataset with different SRs.}
	\begin{tabular}{ccccccc}
		\toprule
		Index & SR    & SNN   & TMac  & NORT  & TTNN  & LFNR \\
		\midrule
		\multirow{10}[2]{*}{PSNR} & 0.05  & 14.22 & \underline{17.41} & 17.30 & 17.28 & \textbf{17.88} \\
		& 0.10  & 15.58 & \underline{19.85} & 19.51 & 19.13 & \textbf{20.02} \\
		& 0.15  & 16.81 & \underline{20.81} & 20.80 & 20.59 & \textbf{21.77} \\
		& 0.20  & 17.99 & 21.4  & \underline{22.03} & 21.84 & \textbf{23.24} \\
		& 0.25  & 19.12 & 21.88 & \underline{23.56} & 23.09 & \textbf{24.50} \\
		& 0.30  & 20.21 & 22.31 & \underline{24.26} & 24.17 & \textbf{25.60} \\
		& 0.35  & 21.31 & 22.70 & 24.76 & \underline{25.33} & \textbf{26.71} \\
		& 0.40  & 22.44 & 23.08 & 25.23 & \underline{26.36} & \textbf{27.75} \\
		& 0.45  & 23.56 & 24.80 & 25.67 & \underline{27.31} & \textbf{28.72} \\
		& 0.50  & 24.71 & 25.31 & 26.11 & \underline{28.29} & \textbf{29.68} \\
		\midrule
		\multirow{10}[2]{*}{SSIM} & 0.05  & 0.2485 & \underline{0.3447} & 0.3104 & 0.3107 & \textbf{0.3526} \\
		& 0.10  & 0.3346 & \underline{0.5143} & 0.4955 & 0.4567 & \textbf{0.5136} \\
		& 0.15  & 0.4282 & \underline{0.5805} & 0.5786 & 0.6218 & \textbf{0.6268} \\
		& 0.20  & 0.5174 & 0.6193 & \underline{0.6572} & 0.6189 & \textbf{0.6987} \\
		& 0.25  & 0.5966 & 0.6527 &\underline{0.7398} & 0.6913 & \textbf{0.7521} \\
		& 0.30  & 0.6651 & 0.6803 & \underline{0.7644} & 0.7455 & \textbf{0.7914} \\
		& 0.35  & 0.7239 & 0.7018 & 0.7838 & \underline{0.7841} & \textbf{0.8261} \\
		& 0.40  & 0.7740 & 0.7229 & 0.8019 & \underline{0.8189} & \textbf{0.8601} \\
		& 0.45  & 0.8155 & 0.8025 & 0.8192 & \underline{0.8378} & \textbf{0.8823} \\
		& 0.50  & 0.8500 & 0.8214 & 0.8359 & \underline{0.8607} & \textbf{0.9002} \\
		\bottomrule
	\end{tabular}%
	\label{label5}%
\end{table}%

Figure \ref{Fig: multispectral} displays the PSNR values versus index of band of different methods for the Balloons and Lemons datasets, where
$\textup{SR} = 0.5, \sigma=0.05$ for the Balloons dataset, and   $\textup{SR}= 0.4, \sigma=0.1$ for the Lemons
dataset.
As can be seen from the two figures that the LFNR performs best compared with SNN, TMac, NORT, and TTNN for almost all bands.
And the TTNN outperforms SNN, TMac, and NORT  in terms of PSNR values for most bands.

Figures \ref{Fig: Balloons} and \ref{Fig: lemons}	 present the recovered images and zoomed regions of the 30th band of different methods
for the  Balloons and Lemons datasets, where $\textup{SR} = 0.4, \sigma=0.05$ for the Balloons  datasets, and
$\textup{SR}= 0.3, \sigma=0.05$ for the Lemons dataset. We can see that the images recovered by
LFNR are better than other comparison methods in term of visual quality, especially from the zoomed regions.
In fact, the LFNR can preserve the details and edges of images better than SNN, TMac, NORT, and TTNN.

\begin{figure}[!t]
	\centering
	\subfigure[]{
		\begin{minipage}[t]{0.46\textwidth}
			\centering
			\raisebox{0.1cm}{\includegraphics[width=7.5cm,height=5.5cm]{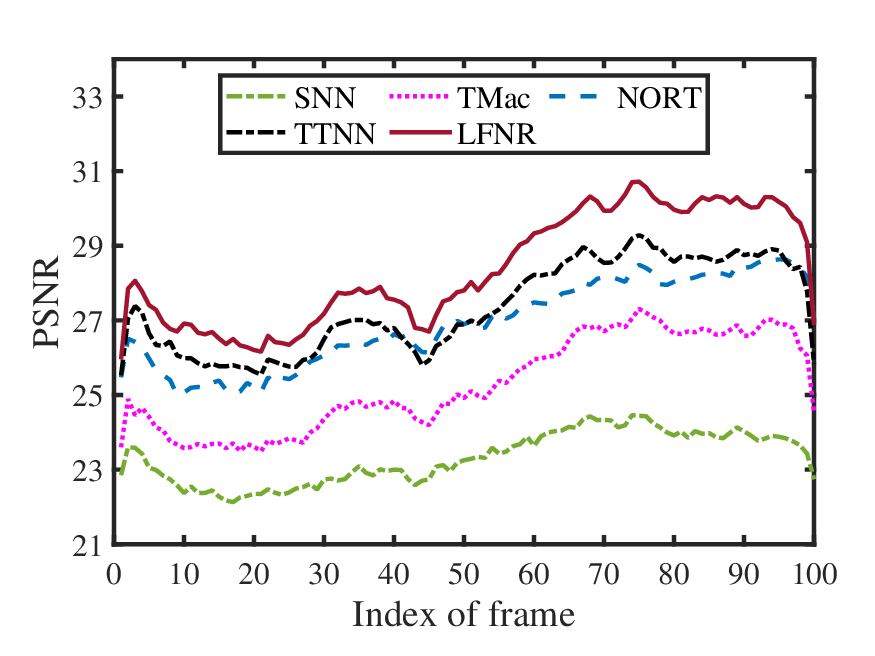}}
		\end{minipage}
	}\hspace{3mm}
	\subfigure[]{
		\begin{minipage}[t]{0.46\textwidth}
			\centering
			\raisebox{0.1cm}{\includegraphics[width=7.5cm,height=5.5cm]{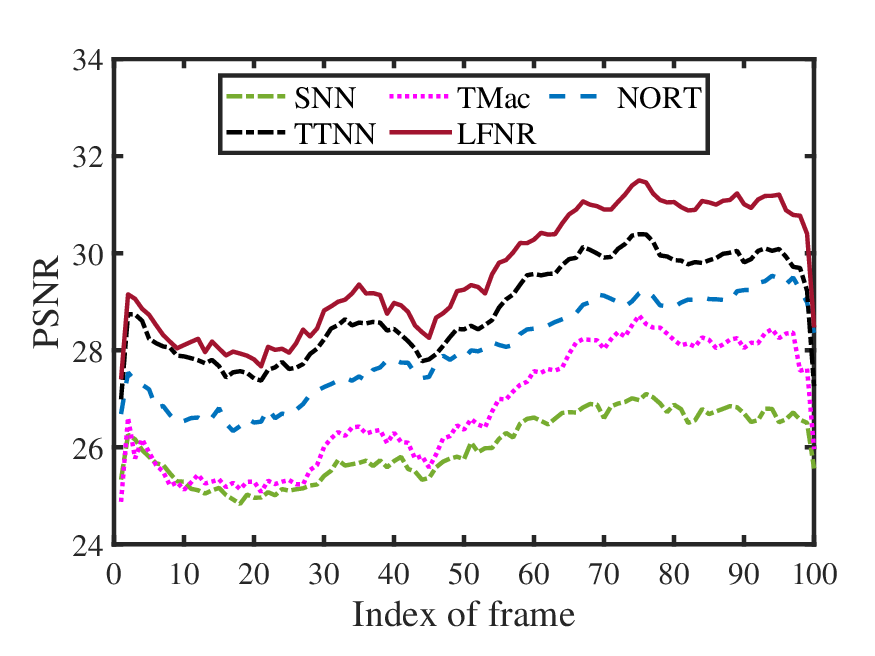}}
		\end{minipage}
	}
	\caption{PSNR values versus index of frame of different methods for the MRI
		dataset. (a) $\textup{SR} = 0.45$ and $\sigma=0.005$. (b) $\textup{SR}= 0.6$ and $\sigma=0.02$.
	}\label{Fig: MRI}	
\end{figure}

\begin{figure}[!t]
	\centering
	\subfigure[Original]{
		\begin{minipage}[t]{0.19\textwidth}
			\centering
			\raisebox{0.1cm}{\includegraphics[width=3.4cm,height=3.4cm]{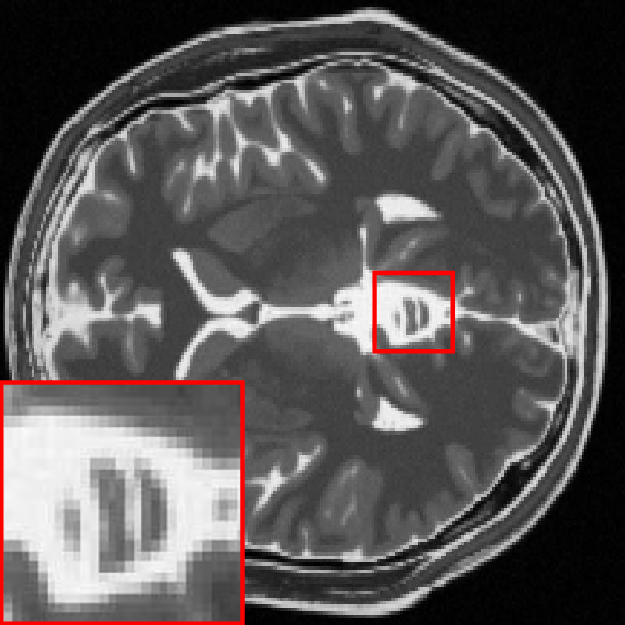}}
		\end{minipage}
	}\hspace{3mm}
	\subfigure[Observation]{
		\begin{minipage}[t]{0.19\textwidth}
			\centering
			\raisebox{0.1cm}{\includegraphics[width=3.4cm,height=3.4cm]{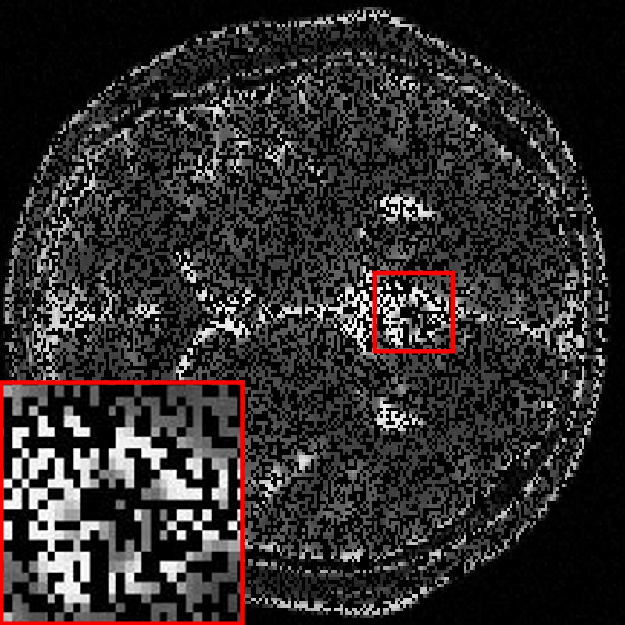}}
		\end{minipage}
	}\hspace{3mm}
	\subfigure[SNN: 25.77]{
		\begin{minipage}[t]{0.19\textwidth}
			\centering
			\raisebox{0.1cm}{\includegraphics[width=3.4cm,height=3.4cm]{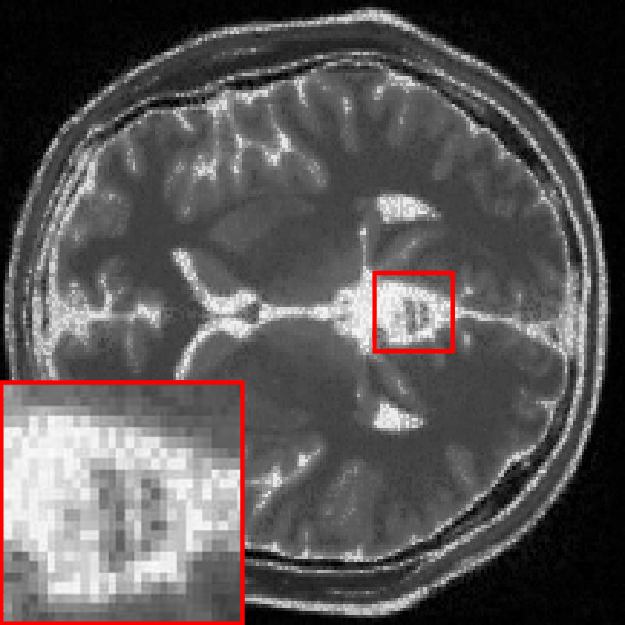}}
		\end{minipage}
	}\hspace{3mm}
	\subfigure[TMac: 27.73]{
		\begin{minipage}[t]{0.19\textwidth}
			\centering
			\raisebox{0.1cm}{\includegraphics[width=3.4cm,height=3.4cm]{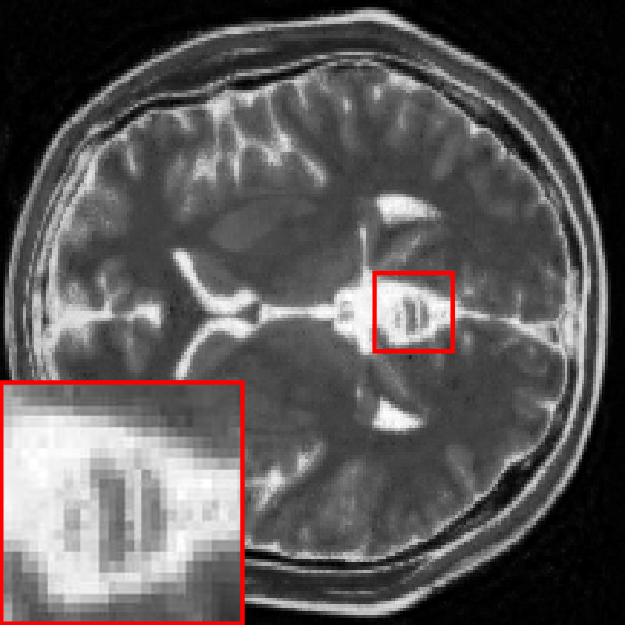}}
		\end{minipage}
	}\hspace{3mm}
	\subfigure[NORT: 28.55]{
		\begin{minipage}[t]{0.19\textwidth}
			\centering
			\raisebox{0.1cm}{\includegraphics[width=3.4cm,height=3.4cm]{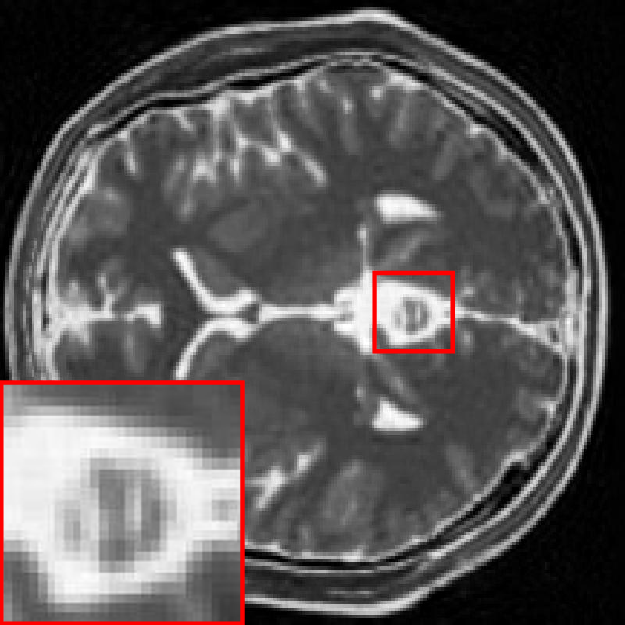}}
		\end{minipage}
	}\hspace{3mm}
	\subfigure[TTNN: 30.27]{
		\begin{minipage}[t]{0.19\textwidth}
			\centering
			\raisebox{0.1cm}{\includegraphics[width=3.4cm,height=3.4cm]{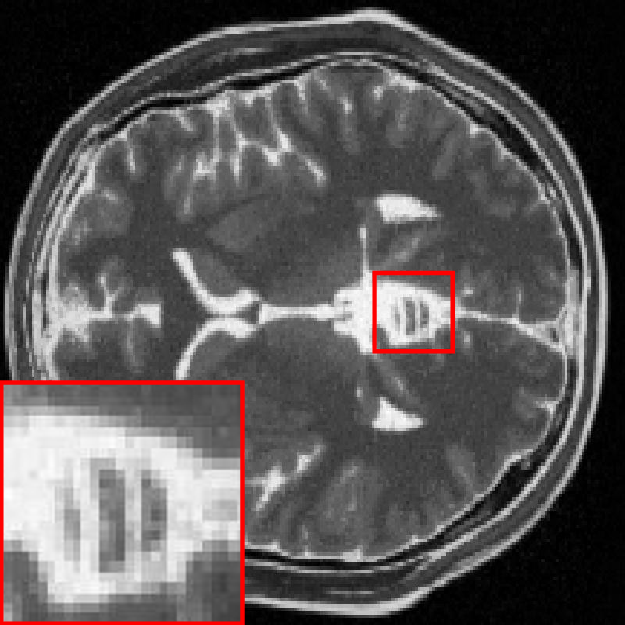}}
		\end{minipage}
	}\hspace{3mm}
	\subfigure[LFNR: 31.51]{
		\begin{minipage}[t]{0.19\textwidth}
			\centering
			\raisebox{0.1cm}{\includegraphics[width=3.4cm,height=3.4cm]{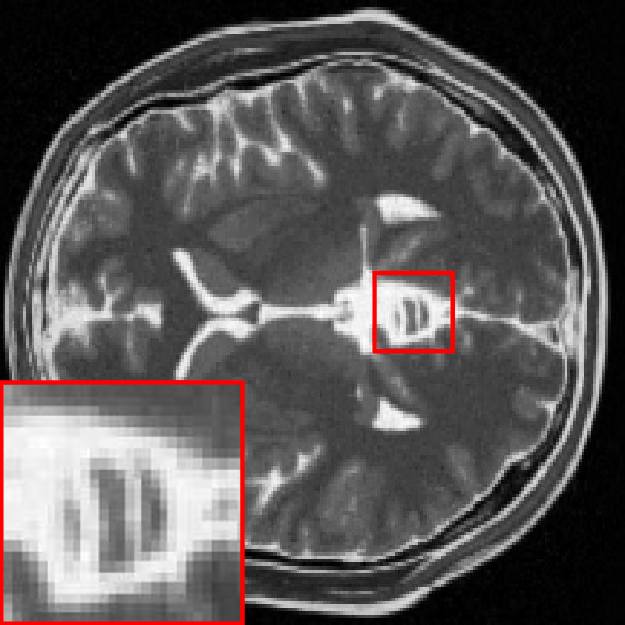}}
		\end{minipage}
	}
	\caption{The recovered images  (with PSNR values) and zoomed regions of different methods for the 74th frontal slice of the MRI dataset,  where $\textup{SR}= 0.5$ and $\sigma=0.005$.}
	\label{Fig:MRIE}	
\end{figure}

\subsubsection{MRI Dataset}

In this subsection, we test a magnetic resonance imaging (MRI) dataset from
Brainweb\footnote{\footnotesize{\url{https://brainweb.bic.mni.mcgill.ca/brainweb/}}} to show the good performance of LFNR,
where the size of MRI is $181\times217\times100$
and the noise level $\sigma$ is set to be $0.01$.
Table \ref{label5} lists the PSNR and SSIM values of different methods for the MRI dataset at different SRs.
where the best and second results are highlighted in bold and in underline, respectively.
It can be observed from
Table \ref{label5} that the LFNR, TMac, NORT and TTNN always reach higher
PSNR and SSIM values than the SNN model.
In particular, the proposed LFNR method obtains
the best recovery performance on both PSNR and SSIM values compared with other approaches.
And the LFNR method has at least 1dB improvement in terms of  PSNR values in comparison with the TTNN model.

In Figure \ref{Fig: MRI}, we show the PSNR values of each frame of different methods for the MRI dataset,
where $\textup{SR} = 0.45$ and $\sigma=0.005$ in Figure \ref{Fig: MRI}(a), and  $\textup{SR}= 0.6, \sigma=0.02$ in Figure \ref{Fig: MRI}(b).
We can see that LFNR achieves  higher PSNR values  than other  methods for all frames.
Besides, the TTNN outperforms SNN, TMac, NORT in most frames.
Figure \ref{Fig:MRIE} exhibits the visual results of different methods on the 74th frontal slice of
the MRI dataset, where $\textup{SR}=0.5, \sigma=0.005$.
Compared with SNN, TMac, NORT and TTNN, the images obtained by LFNR is visually closest to the original image, especially for the zoomed regions.
In fact, the LFNR can preserve the details and texture better than other methods.

\subsection{Logistic Regression for Binary Classification}

In this subsection, we investigate the logistic regression loss for binary classification in problem (\ref{test1}).
Specifically, we consider the case in which
the observation pairs $\{(\mathcal Z_i, y_i)\}_{i=1}^n$ are drawn independent and identically distributed (i.i.d.) from a distribution
of the form
\begin{equation}\label{LosP}
	\mathbb{P}(y_i| \mathcal Z_i,\mathcal X)=\exp\{y_i\langle\mathcal Z_i, \mathcal X\rangle-\log \left(1+\exp (\left\langle\mathcal{Z}_i, \mathcal X \right\rangle)\right)\}.
\end{equation}
Here the response $y_i$ takes binary values $\{0,1\}$.
In addition, $\mathcal X$ is an unknown parameter tensor, which need to be estimated.
By maximum likelihood estimate,
the   loss function based on the distribution of the observations in (\ref{LosP}) can be written as
\begin{equation}\label{eq61}
	f_{n,y}(\mathcal{X}):=\frac{1}{n} \sum_{i=1}^n[\log (1+\exp (\langle\mathcal{Z}_i, \mathcal X \rangle))-y_i\langle \mathcal{Z}_i,\mathcal X\rangle].
\end{equation}
Then problem (\ref{test1}) reduces to
\begin{equation}\label{LRMN}
\widetilde{\mathcal X}=\underset{\|\mathcal X\|_{\infty} \leq c}{\operatorname{argmin}}\left\{
\frac{1}{n} \sum_{i=1}^n[\log \left(1+\exp (\langle\mathcal{Z}_i, \mathcal X \rangle)\right)-y_i\langle \mathcal{Z}_i,\mathcal X\rangle]+\beta G_\lambda(\mathcal X) \right\}.
\end{equation}
Model (\ref{LRMN}) combines the logistic regression loss with nonconvex regularization.
In particular,  when $G_\lambda$ is the TTNN, model (\ref{LRMN}) is also called TTNN for brevity.

We compare our methods with
the support vector machine (SVM), which comes from the {\bf fitcsvm} function in MATLAB, the dual structure preserving kernels approach (DuSK)\footnote{\footnotesize \url{https://github.com/LifangHe/SDM14_DuSK}} \cite{DuSK2014},
 support tensor train machine by employing  tensor train as the parameter
 model (STTM)\footnote{\footnotesize{\url{https://github.com/git2cchen/STTM}}} \cite{2019STTM}, logistic loss with overlapped trace norm (LOTN) \cite{wimalawarne2016theoretical}.
 The CIFAR-10 dataset\footnote{\footnotesize{\url{https://www.cs.toronto.edu/~kriz/cifar.html}}}  is tested in the experiments for binary classification, where there are $60000$ color images ($32 \times 32\times3$) from $10$ classes, including $50000$ training images and $10000$ testing images.
 In the following experiments,  each case run $10$ times to reduce the impact of randomness, and the final results are reported by the average value of all results of ten times.
 In particular,  $\mathcal Z_i$ (with size $32\times 32\times 3$) are    the training images and $y_i$ are the labels of the corresponding training images in our experiments for image classification, $i=1,\ldots, n$. Here $n$ is the number of training samples.

 Let $\tilde{m}$ be the number of the testing images. Then the response of the testing images is defined as
 $$
 y_j^\text{testp}=\frac{1}{1+\exp (-\langle\mathcal{Z}_j^{\text{test}}, \widetilde{\mathcal X} \rangle)}, \quad j=1,2,\ldots,\tilde{m},
 $$
  where $\mathcal{Z}_j^{\text{test}}$ denotes the $j$-th testing image  and $\widetilde{\mathcal X}$ is the trained parameter in (\ref{LRMN}).
 If $y_j^\text{testp}>0.5$, set $y_j^\text{testp}=1$, otherwise, $y_j^\text{testp}=0$.
 The testing accuracy (TAc) of classification is defined as
 $$
 \text{TAc}=1-\frac{1}{\tilde{m}}\sum_{j=1}^{\tilde{m}}|y_j^\text{testp}-y_j^\text{test}|,
 $$
where $y_j^\text{test}$ denotes the true category of the $j$-th testing image.

\subsubsection{Parameter Settings}

For the experiments of image classification, we set $\rho=100$, and choose $\eta$ from the set $\{10,100\}$ to get the best classification results.
For the parameters $\gamma$ and $\lambda$ in the MCP function, we choose  $\gamma$  from the set $\{0.2,0.3,0.4,0.5,0.6,0.8,0.9,1.5,2.5,3.4,5,6\}$
and $\lambda$ from the set
$\{0.1,0.2,0.3,0.4,0.5,0.6,0.8\}$ to obtain the best classification accuracy.
In addition, the penalty parameter $\beta$ is chosen from the set
$\{0.1,0.2,0.3,0.4,0.6,0.7\}$ to achieve the best results.

\begin{table}[!t]
	\centering
	\caption{Classification  accuracy of different methods for the CIFAR-10 dataset.}
	\begin{tabular}{cccccccc}
		\toprule
		Class 1 & Class 2 & SVM   & DuSK  & STTM & LOTN & TTNN  & LFNR \\
		\midrule
		dog  & frog   & 0.675 & 0.738 & 0.721 &0.696 & \underline{0.747} & \textbf{0.753} \\
		automobile    & bird & 0.787 & 0.839 & 0.826 & 0.820 & \underline{0.857} & \textbf{0.861} \\
		bird  & deer   & 0.625 & \underline{0.719} & 0.676 & 0.649 & 0.715  & \textbf{0.737} \\
		cat   & bird    & 0.637 & 0.712 & 0.716 & 0.700   & \underline{0.738} & \textbf{0.742} \\
		bird  & dog   & 0.632 & \underline{0.719} & 0.703 & 0.647 & 0.715 & \textbf{0.730} \\
		truck & automobile & 0.598 & 0.668 & 0.666 & 0.624 & \underline{0.680} & \textbf{0.686} \\
		bird  & horse & 0.685 & \underline{0.727} & 0.695 & 0.714 & 0.723 & \textbf{0.750} \\
		deer  & frog  & 0.665 & 0.735 & 0.705 & 0.712 & \underline{0.747} & \textbf{0.752} \\
		frog  & automobile & 0.836 & \underline{0.879} & 0.855 & 0.853 & 0.868 & \textbf{0.882} \\
		bird  & airplane & 0.623 & 0.780 & 0.752 & 0.689 & \underline{0.796} & \textbf{0.812} \\
		\bottomrule
	\end{tabular}%
	\label{Cifar-10}%
\end{table}%

\begin{figure}[!t]
	\centering
	\includegraphics[width=8.3cm]{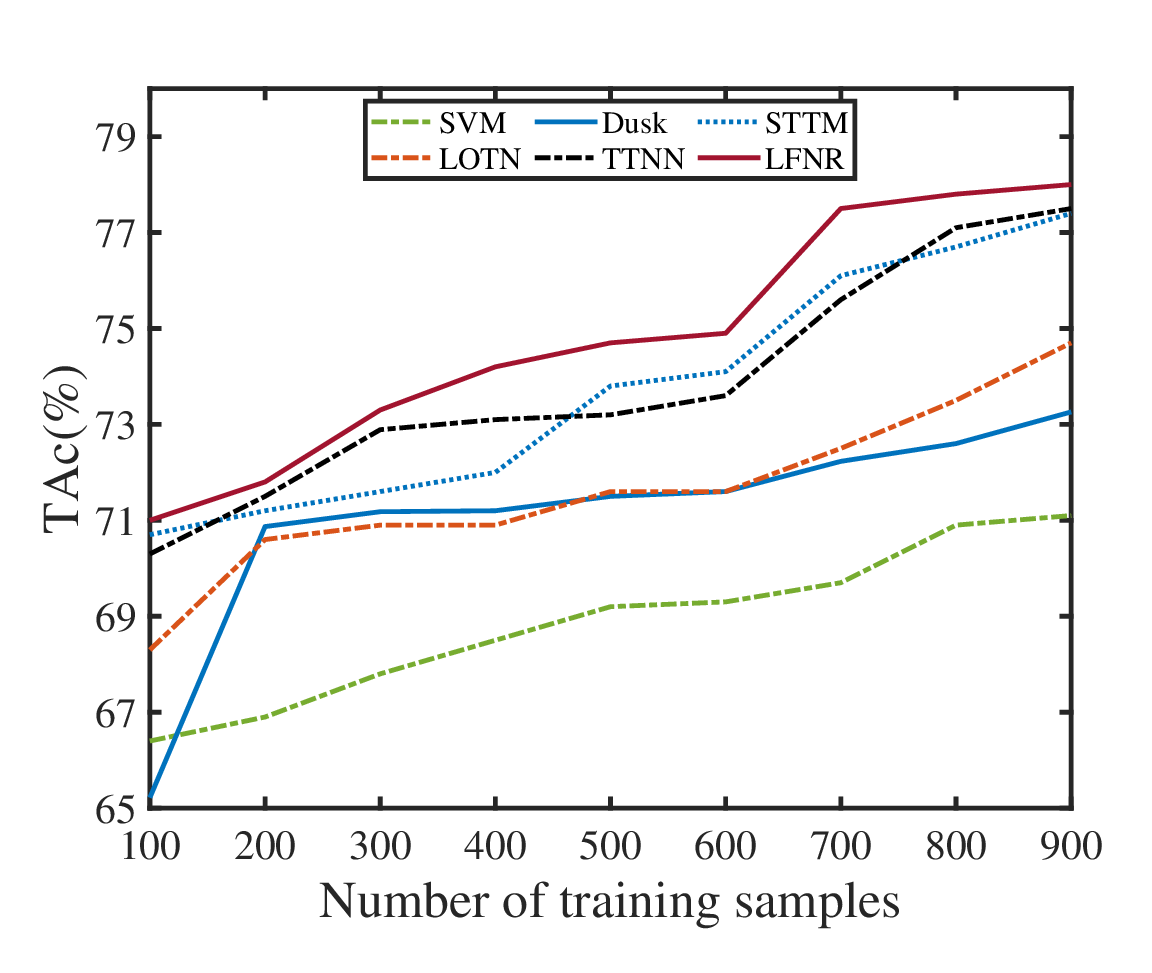}
	\caption{Classification  accuracy versus number of training samples of different methods for the  CIFAR-10 dataset.}
	\label{cifar}
\end{figure}

\subsubsection{Image Classification}

In this subsection, we demonstrate the effectiveness of the LFNR for binary image classification.
First, we select two out of the ten classes in CIFAR-$10$. Then
the first $250$ images of the two classes are used in our experiments.
More specifically, we randomly choose $200$ images from each class for training, and employ the remaining ones as the testing images.
Table \ref{Cifar-10} displays the classification accuracy of different methods for the CIFAR-10 dataset.
It can be observed from this table that the LFNR performs better than SVM, DuSK, STTM, LOTN, and TTNN in terms of classification accuracy for the testing cases.
Besides, for the classes `bird' and `deer', `bird' and `dog', `bird' and `house', `frog' and `automobile', the DuSK outperforms TTNN in terms of classification accuracy.
However, the TTNN performs better than other methods for other testing cases.
In particular, the TTNN can achieve higher classification accuracy  than LOTN for these images, which implies that the TTNN can explore the low-rankness better than overlapped trace norm.

Now we test the influence of different number of training samples of different methods for image classification.
Two categories of `deer' and `horse' are used in our experiments,
where $100$ images are randomly selected from
each class as the testing samples and the number of training samples varies from $100$ to $900$ with step size $100$.
In  Figure \ref{cifar}, we show the classification accuracy of different methods versus number of training samples for the `deer' and `horse' classes.
We can see that the classification accuracy  of LFNR is higher than those of SVM, DuSK, STTM, LOTN, and TTNN.
Furthermore, the classification accuracy  of these methods increases as the number of training samples increases.

\section{Concluding Remarks}\label{Conc}

In this paper, we have studied the problem of low-rank tensor learning and proposed the LFNR model based on TTNN.
Specifically, in order to explore the low-rankness of the underlying tensor,
a family of nonconvex functions were employed onto the singular values of
all frontal slices of a tensor in the transformed domain.
Besides, a general loss function was considered in the proposed model,
which could be specified to the least squares loss and logistic regression loss, respectively.
And the corresponding models were suitable for tensor completion and binary classification.
Then we established the error bound between the stationary point of
the LFNR model and the underlying tensor under the RSC condition on the loss function
and  some regularity conditions on the nonconvex penalty function.
Furthermore, based on the DC structure of the nonconvex regularization, a PMM algorithm was designed to solve the proposed model,
where the loss function and one convex function in DC structure were linearized with a proximal term.
Besides, we  showed that the sequence generated by  PMM converges globally to a stationary point of LFNR under very mild conditions.
Numerical experiments on tensor completion and binary classification were reported to demonstrate the effectiveness of LFNR compared with other methods.

In the future work, we are going to extend our model to more general loss functions,
where the RSC conditions are utilized to more general loss functions.
Another possible extension is to incorporate the random method instead of using SVD directly to accelerate the PMM.
It would be also of great interest to analyze the KL exponent of the objective function of the LFNR model.

\section*{Acknowledgments}
The authors would like to thank Dr. Silvia Gandy for providing the code of SNN in \cite{Gandy_2011}.

\section*{Appendix A. Least Squares Loss for Tensor Completion}

In the following, we show that the loss function in (\ref{eq30}) satisfies the RSC condition.
The least squares loss function in (\ref{eq30}) can  be also rewritten as

$$
f_{n,\mathcal{Y}}(\mathcal{X})=\frac{1}{2p}\|\mathcal{P}_{\Omega}(\mathcal{X}-\mathcal{Y})\|_F^2
=\frac{1}{2p}\sum_{(i,j,k)\in\Omega}(\langle\mathcal{A}^{ijk},\mathcal{X}\rangle-\mathcal{Y}_{ijk})^2,
$$
where $\mathcal{A}^{ijk}\in\mathbb{R}^{n_1\times n_2\times n_3}$ denotes a basic tensor whose $(i,j,k)$-th element is 1 and other elements are 0.
Note that
the gradient of the loss function $f_{n,\mathcal{Y}}(\mathcal{X})$ is given by
\begin{equation}\label{eq53}
	\nabla f_{n,\mathcal{Y}}(\mathcal{X})
	=\frac{1}{p}\sum_{(i,j,k)\in\Omega}(\langle\mathcal{A}^{ijk},\mathcal{X}\rangle-\mathcal{Y}_{ijk})\mathcal{A}^{ijk}.
\end{equation}

Now we show that $f_{n,\mathcal{Y}}(\mathcal{X})$ satisfies the RSC condition in (\ref{test0}) with high probability, which is stated in the following lemma.

\begin{lemma}\label{lem20}
	For any  $\tilde{\mathcal V}\in\mathbb{R}^{n_1\times n_2\times n_3}$,
	there exists a constant  $\eta_1 \in (0,1)$ such that
	\begin{equation}\label{eq50}
		\begin{aligned}
			\langle\nabla f_{n,\mathcal{Y}}(\mathcal X^*+\tilde{\mathcal V})
			-\nabla f_{n,\mathcal{Y}}(\mathcal X^*), \tilde{\mathcal V}\rangle\geq(1-\eta_1)\|\tilde{\mathcal V}\|_F^2,
		\end{aligned}
	\end{equation}
	with probability at least $1-\textup{exp}(- \frac{2\eta_1^2 n^2}{d^3})$.
\end{lemma}
\textbf{Proof. }
For any $\tilde{\mathcal V}\in\mathbb{R}^{n_1\times n_2\times n_3}$, we let $b:=\|\tilde{\mathcal V}\|_\infty$.
By (\ref{eq53}), one can get
\begin{equation}\label{eq74}
	\begin{aligned}
		&~~~~~\langle\nabla f_{n,\mathcal{Y}}(\mathcal X^*+\tilde{\mathcal V})
		-\nabla f_{n,\mathcal{Y}}(\mathcal X^*), \tilde{\mathcal V}\rangle\\
		&=\left\langle\frac{1}{p}\sum_{(i,j,k)\in\Omega}(\langle\mathcal{A}^{ijk},\mathcal X^*+\tilde{\mathcal V}\rangle-\mathcal{Y}_{ijk})\mathcal{A}^{ijk}-(\langle\mathcal{A}^{ijk},\mathcal X^*\rangle-\mathcal{Y}_{ijk})\mathcal{A}^{ijk},\tilde{\mathcal V} \right\rangle\\
		&=\left\langle\frac{1}{p}\sum_{(i,j,k)\in\Omega}\langle\mathcal{A}^{ijk},\tilde{\mathcal V}\rangle\mathcal{A}^{ijk},\tilde{\mathcal V} \right\rangle\\
		&=\frac{1}{p}\sum_{(i,j,k)\in\Omega}(\langle\mathcal{A}^{ijk},\tilde{\mathcal V}\rangle)^2.
	\end{aligned}
\end{equation}
Denote $\xi_{ijk}=(\langle\mathcal{A}^{ijk}, \tilde{\mathcal V}\rangle)^2$  if $(i,j,k)\in\Omega$,  $0$ otherwise, $1\leq i\leq n_1,1\leq j \leq n_2, 1\leq k \leq n_3$.
Notice that the observations are sampled at random, we know that
\begin{equation}\label{eq73}
	\xi_{ijk}=\left\{
	\begin{aligned}
		\tilde{\mathcal V}_{ijk}^2, &\quad \text{with probability}\ p,\\
		0, &\quad \text{with probability}\ 1-p.
	\end{aligned}
	\right.
\end{equation}
Note that $\{\xi_{ijk}\}$ are independent random variables
and the expectation of $\xi_{ijk}$
is $\mathbb{E}(\xi_{ijk})=p\tilde{\mathcal V}_{ijk}^2$.
Let
$$
\alpha=\sum_{i=1}^{n_1}\sum_{j=1}^{n_2}\sum_{k=1}^{n_3}\xi_{ijk},
$$
then $\mathbb{E}(\alpha)=p\sum_{i=1}^{n_1}\sum_{j=1}^{n_2}\sum_{k=1}^{n_3}\tilde{\mathcal V}_{ijk}^2=p\|\tilde{\mathcal V}\|_F^2$.
Notice that $0\leq\xi_{ijk}\leq b^2$.
For any $\epsilon>0$, it follows from the Hoeffding's inequality \cite[Proposition 2.5]{Wainwright_2019} that
$$
\mathbb{P}(\alpha-\mathbb{E}(\alpha)<-\epsilon)< \text{exp}\left(-\frac{2\epsilon^2}{d b^4}\right).
$$
By taking $\epsilon:=\eta_1 p \|\tilde{\mathcal V}\|_F^2$, where  $\eta_1\in(0,1)$ is a constant, we deduce that
\begin{equation}
	\begin{aligned}
		\mathbb{P}\left(\alpha<\mathbb{E}(\alpha)-\eta_1 p \|\tilde{\mathcal V}\|_F^2\right)
		&=\mathbb{P}\left(\frac{1}{p}\alpha<(1-\eta_1)\|\tilde{\mathcal V}\|_F^2\right)\\
		&< \text{exp}\left(-\frac{2\eta_1^2 p^2 \|\tilde{\mathcal V}\|_F^4}{d b^4}\right),
	\end{aligned}
\end{equation}
which implies
\begin{equation}\label{eq43}
	\begin{aligned}
		\mathbb{P}\left(\frac{1}{p}\alpha\geq(1-\eta_1)\|\tilde{\mathcal V}\|_F^2\right)
		&\geq 1-\text{exp}\left(-\frac{2\eta_1^2 p^2 \|\tilde{\mathcal V}\|_F^4}{d b^4}\right).
	\end{aligned}
\end{equation}
Since $1\leq \sqrt{d}\frac{\|\tilde{\mathcal V}\|_\infty}{\|\tilde{\mathcal V}\|_F}\leq \sqrt{d}$, we obtain that $\frac{1}{d^2}\leq \frac{\|\tilde{\mathcal V}\|_F^4}{d^2\|\tilde{\mathcal V}\|_\infty^4}\leq 1,$ which implies that
$$
0<\frac{1}{d}\leq \frac{\|\tilde{\mathcal V}\|_{F}^4}{d\|\tilde{\mathcal V}\|_{\infty}^4}=\frac{\|\tilde{\mathcal V}\|_{F}^4}{db^4}\leq {d}.
$$
As a consequence, (\ref{eq43})  further implies that
\begin{equation}\label{XV1}
	\begin{aligned}
		\mathbb{P}\left(\frac{1}{p}\alpha\geq(1-\eta_1)\|\tilde{\mathcal V}\|_F^2\right)
		&=\mathbb{P}\left(\frac{1}{p}\sum_{i=1}^{n_1}\sum_{j=1}^{n_2}\sum_{k=1}^{n_3}\xi_{ijk}\geq(1-\eta_1)\|\tilde{\mathcal V}\|_F^2\right)\\
		&\geq 1-\text{exp}\left(- \frac{2\eta_1^2 n^2}{d^3}\right),
	\end{aligned}
\end{equation}
Combining (\ref{XV1}) with (\ref{eq74}) yields
$$
\begin{aligned}
	\mathbb{P}\left( \langle\nabla f_{n,\mathcal{Y}}(\mathcal X^*+\tilde{\mathcal V})
	-\nabla f_{n,\mathcal{Y}}\left(\mathcal X^* \right), \tilde{\mathcal V}\rangle\geq(1-\eta_1)\|\tilde{\mathcal V}\|_F^2\right)
	&\geq 1-\text{exp}\left(- \frac{2\eta_1^2 n^2}{d^3}\right).
\end{aligned}
$$
This concludes the proof.
\qed

From Lemma \ref{lem20}, we know that $f_{n,\mathcal{Y}}(\mathcal{X})$ defined in (\ref{eq30}) satisfies the RSC condition in (\ref{test0}) by taking
$\alpha_1=\alpha_2=1-\eta_1$, $\tau_1=\tau_2=0$.

\section*{Appendix B. Logistic Regression Loss}

First, we give the definition and property of a sub-Gaussian random variable, which plays a vital  role in the proof of Lemma \ref{lem21}.

\begin{definition}\label{defin51}\cite{Foucart2013}
	A random variable $x$ is called sub-Gaussian if there exist constants $a_1, \kappa>0$ such that
	$
	\mathbb{P}(|x| \geq t) \leq a_1 e^{-\kappa t^2}$ for any $t>0$.
	
\end{definition}
\begin{Prop}\label{Prop1}\cite[Proposition 7.24]{Foucart2013}
	If $x$ is sub-Gaussian with $\mathbb{E}(x)=0$, then there exists a constant $\tilde b$ (depending only on $a_1$ and $\kappa)$ such that
	\begin{equation}\label{Pro1}
		\mathbb{E}[\exp (\theta x)] \leq \exp \left(\tilde b \theta^2\right) \quad \text { for all } \theta \in \mathbb{R},
	\end{equation}
	where the constant $\tilde b$ is called a sub-Gaussian parameter of $x$.
\end{Prop}

Now we show that	the logistic  loss function $f_{n,y}$ in  (\ref{eq61}) satisfies the RSC condition (\ref{test0})
with high probability in the following lemma.

\begin{lemma}\label{lem21}
	Assume that $\{(\mathcal{Z}_i,y_i)\}_{i=1}^n$
	are drawn i.i.d., where $\mathcal{Z}_i, i=1,2,\ldots,n$, are sub-Gaussian
	with independent, mean-zero, and sub-Gaussian entries with the same sub-Gaussian parameter $\tilde b$ in (\ref{Pro1}),
	and the response variables $y_i\in\{0,1\}$.
	Suppose that the number of samples $n\geq 4 c_2^4  t^2\log d$,
	where $c_2>0$ is a given constant and $t$ is defined in Section \ref{StaGua}.
	Then there exists some positive constant $c_3$ such that
	the loss function $f_{n,y}$ in  (\ref{eq61}) satisfies the RSC condition (\ref{test0})
	with probability at least $1-\exp \left(-c_3 \log d\right)$.
\end{lemma}
\textbf{Proof.}
Note that
the gradient of $f_{n,y}(\mathcal{X})$ in (\ref{eq61}) is
\begin{equation}\label{eq70}
	\nabla f_{n,y}(\mathcal{X})=\frac{1}{n} \sum_{i=1}^n
	\left[
	\left(\frac{\exp (\langle \mathcal{Z}_i, \mathcal X\rangle)}{1+\exp \left(\left\langle \mathcal{Z}_i, \mathcal X\right\rangle\right)}-y_i\right)\mathcal{Z}_i
	\right].
\end{equation}
For any third-order tensor $\tilde{\mathcal V}\in\mathbb{R}^{n_1\times n_2\times n_3}$,
we have
$$
\begin{aligned}
	&~~~~\nabla f_{n,y}(\mathcal X^*+\tilde{\mathcal V})
	-\nabla f_{n,y}(\mathcal X^* )\\
	&=\frac{1}{n} \sum_{i=1}^n
	\left[
	\left(\frac{\exp (\langle \mathcal{Z}_i, \mathcal X^*+\tilde{\mathcal V}\rangle)}{1+\exp (\langle \mathcal{Z}_i, \mathcal X^*+\tilde{\mathcal V}\rangle)}-y_i\right)\mathcal{Z}_i\right]
	-\frac{1}{n} \sum_{i=1}^n
	\left[
	\left(\frac{\exp (\langle \mathcal{Z}_i, \mathcal X^*\rangle)}{1+\exp \left(\left\langle \mathcal{Z}_i, \mathcal X^*\right\rangle\right)}-y_i\right)\mathcal{Z}_i\right]\\
	&=\frac{1}{n} \sum_{i=1}^n
	\left[
	\left(\frac{\exp (\langle \mathcal{Z}_i, \mathcal X^*+\tilde{\mathcal V}\rangle)}{1+\exp (\langle \mathcal{Z}_i, \mathcal X^*+\tilde{\mathcal V}\rangle)}-\frac{\exp (\langle \mathcal{Z}_i, \mathcal X^*\rangle)}{1+\exp \left(\left\langle \mathcal{Z}_i, \mathcal X^*\right\rangle\right)}\right)\mathcal{Z}_i\right].
\end{aligned}
$$
Therefore, one can deduce that
\begin{equation}\label{eq64}
	\begin{aligned}
		&~~~~\langle\nabla f_{n,y}(\mathcal X^*+\tilde{\mathcal V})
		-\nabla f_{n,y}(\mathcal X^* ), \tilde{\mathcal V}\rangle\\
		&=\frac{1}{n} \sum_{i=1}^n
		\left[
		\left(\frac{\exp (\langle \mathcal{Z}_i, \mathcal X^*+\tilde{\mathcal V}\rangle)}{1+\exp (\langle \mathcal{Z}_i, \mathcal X^*+\tilde{\mathcal V}\rangle)}-\frac{\exp (\langle \mathcal{Z}_i, \mathcal X^*\rangle)}{1+\exp \left(\left\langle \mathcal{Z}_i, \mathcal X^*\right\rangle\right)}\right)\right]\langle\mathcal{Z}_i, \tilde{\mathcal V}\rangle.
	\end{aligned}
\end{equation}
Define the  function $\psi(u):=\log \left(1+\exp(u)\right)$.
Note that
the first-order  derivative of $\psi(u)$ is
$
\psi{'}(u)=\frac{\exp (u)}{1+\exp \left(u\right)}
$.
Then (\ref{eq64}) can be rewritten as
\begin{equation}\label{graeq70}
	\begin{aligned}
		\langle\nabla f_{n,y}(\mathcal X^*+\tilde{\mathcal V})
		-\nabla f_{n,y}(\mathcal X^* ), \tilde{\mathcal V}\rangle
		&=\frac{1}{n} \sum_{i=1}^n
		\left(
		\psi{'}(\langle\mathcal{Z}_i, \mathcal X^*+\tilde{\mathcal V} \rangle)
		-\psi{'}(\langle\mathcal{Z}_i, \mathcal X^* \rangle)
		\right)\langle\mathcal{Z}_i, \tilde{\mathcal V}\rangle\\
		&=\frac{1}{n} \sum_{i=1}^n
		\psi{''}(\langle\mathcal{Z}_i, \mathcal X^* \rangle+t_i\langle\mathcal{Z}_i, \tilde{\mathcal V} \rangle)
		\langle\mathcal{Z}_i, \tilde{\mathcal V}\rangle^2,
	\end{aligned}
\end{equation}
where the second equality holds by the mean value theorem and  $t_i\in[0,1]$.
Now we consider two cases $\|\tilde{\mathcal V}\|_F\leq 1$ and $\|\tilde{\mathcal V}\|_F>1$.

{\bf Case I}. Suppose that  $\|\tilde{\mathcal V}\|_F=\delta\in(0,1]$.
First, similar to \cite[Theorem 9.36]{Wainwright_2019}, we  introduce the following  assumptions
\begin{equation}\label{ineq1}
	\mathbb{E}(\langle\mathcal{Z}_i, \tilde{\mathcal V}\rangle^2)\geq \tau_u
	\text{ and }
	\mathbb{E}(\langle\mathcal{Z}_i, \tilde{\mathcal V}\rangle^4)\leq \tau_l, \quad
	\text{for any $\tilde{\mathcal V}\in\mathbb{R}^{n_1\times n_2\times n_3}$ with $\|\tilde{\mathcal V}\|_F=1$},
\end{equation}
where $\tau_u, \tau_l>0$ are given constants.
In (\ref{ineq1}), the first inequality  is used to control the lower bound of covariance of  the entries of $\mathcal{Z}_i$
and the second inequality is used to simplify the upper bound of $\mathbb{E}(\langle\mathcal{Z}_i, \tilde{\mathcal V}\rangle^4)$.
Indeed, by simple calculations, we can get $\mathbb{E}(\langle\mathcal{Z}_i, \tilde{\mathcal V}\rangle^4)\leq(4\tilde{b})^2\times 2=64\tilde{b}^2$ \cite[Proposition 3.2]{Rivasplata2012}.
Similar to \cite{Banerjee2014,loh15a},
we use a truncated version to  the right-hand side of (\ref{graeq70}).
Denote a truncation level $\tau:=K\delta$ with some constant $K>0$.
For any tensor $\tilde{\mathcal V}\in\mathbb{R}^{n_1\times n_2\times n_3}$ with $\|\tilde{\mathcal V}\|_F=\delta\in(0,1]$, we define
$$
\mathbb{I}[|\langle \mathcal{Z}_i, \tilde{\mathcal V}\rangle|\leq 2\tau]:=\left\{
\begin{aligned}
	1,  \quad & \text{if } |\langle \mathcal{Z}_i, \tilde{\mathcal V}\rangle|\leq 2\tau, \\
	0,  \quad  &\text{otherwise},
\end{aligned}
\right.
\quad ~~~~ \\
\mathbb I[|\langle \mathcal{Z}_i, \tilde{\mathcal V}\rangle|> 2\tau]:=\left\{
\begin{aligned}
	1,  \quad & \text{if } |\langle \mathcal{Z}_i, \tilde{\mathcal V}\rangle| > 2\tau, \\
	0,  \quad  &\text{otherwise}.
\end{aligned}
\right.
$$
In addition, for any given truncation level $T>0$, define the functions
$$
\mathbb{I}[|\langle \mathcal{Z}_i, \mathcal X^*\rangle|\leq T]:=\left\{
\begin{aligned}
	1,  \quad & \text{if } |\langle \mathcal{Z}_i, \mathcal X^*\rangle|\leq T, \\
	0,  \quad  &\text{otherwise},
\end{aligned}
\right.
\quad ~~~~ \\
\mathbb I[|\langle \mathcal{Z}_i, \mathcal X^*\rangle|> T]:=\left\{
\begin{aligned}
	1,  \quad & \text{if } |\langle \mathcal{Z}_i, \mathcal X^*\rangle| > T, \\
	0,  \quad  &\text{otherwise}.
\end{aligned}
\right.
$$
Now we can represent $\psi{''}(\langle\mathcal{Z}_i, \mathcal X^* \rangle+t_i\langle\mathcal{Z}_i, \tilde{\mathcal V} \rangle)
\langle\mathcal{Z}_i, \tilde{\mathcal V}\rangle^2$ in (\ref{graeq70}) via above functions as follows
\begin{equation}\label{eq72}
	\begin{aligned}
		&~\langle\nabla f_{n,y}(\mathcal X^*+\tilde{\mathcal V})
		-\nabla f_{n,y}(\mathcal X^* ), \tilde{\mathcal V}\rangle\\
		=\ &\frac{1}{n} \sum_{i=1}^n
		\psi{''}(\langle\mathcal{Z}_i, \mathcal X^* \rangle+t_i\langle\mathcal{Z}_i, \tilde{\mathcal V} \rangle)
		\langle\mathcal{Z}_i, \tilde{\mathcal V}\rangle^2
		\mathbb{I}[|\langle \mathcal{Z}_i, \tilde{\mathcal V}\rangle|\leq 2\tau]
		\mathbb I\left[\left|\langle \mathcal{Z}_i, \mathcal X^*\rangle\right|\leq T\right]
		\\
		&+\frac{1}{n} \sum_{i=1}^n
		\psi{''}(\langle\mathcal{Z}_i, \mathcal X^* \rangle+t_i\langle\mathcal{Z}_i, \tilde{\mathcal V} \rangle)
		\langle\mathcal{Z}_i, \tilde{\mathcal V}\rangle^2
		\mathbb{I}[|\langle \mathcal{Z}_i, \tilde{\mathcal V}\rangle|\leq 2\tau]
		\mathbb I\left[\left|\langle \mathcal{Z}_i, \mathcal X^*\rangle\right|> T\right]
		\\
		&+\frac{1}{n} \sum_{i=1}^n
		\psi{''}(\langle\mathcal{Z}_i, \mathcal X^* \rangle+t_i\langle\mathcal{Z}_i, \tilde{\mathcal V} \rangle)\langle\mathcal{Z}_i, \tilde{\mathcal V}\rangle^2
		\mathbb I[|\langle \mathcal{Z}_i, \tilde{\mathcal V}\rangle|> 2\tau]
		\mathbb{I}\left[\left|\langle \mathcal{Z}_i, \mathcal X^*\rangle\right| \leq T\right]
		\\
		&+\frac{1}{n} \sum_{i=1}^n
		\psi{''}(\langle\mathcal{Z}_i, \mathcal X^* \rangle+t_i\langle\mathcal{Z}_i, \tilde{\mathcal V} \rangle)
		\langle\mathcal{Z}_i, \tilde{\mathcal V}\rangle^2
		\mathbb I[|\langle \mathcal{Z}_i, \tilde{\mathcal V}\rangle|> 2\tau]
		\mathbb I\left[\left|\langle \mathcal{Z}_i, \mathcal X^*\rangle\right| > T\right]\\
		\geq \	& \frac{1}{n} \sum_{i=1}^n
	\psi{''}(\langle\mathcal{Z}_i, \mathcal X^* \rangle+t_i\langle\mathcal{Z}_i, \tilde{\mathcal V} \rangle)
	\langle\mathcal{Z}_i, \tilde{\mathcal V}\rangle^2
	\mathbb{I}[|\langle \mathcal{Z}_i, \tilde{\mathcal V}\rangle|\leq 2\tau]
	\mathbb I\left[\left|\langle \mathcal{Z}_i, \mathcal X^*\rangle\right|\leq T\right],
	\end{aligned}
\end{equation}
where the inequality holds by the fact that the second derivative $\psi{''}$ is always positive.

Note that $0<\tau=K\delta\leq K$.
Define $\ell_\psi(T):=\min _{|u| \leq (T+2K)} \psi{''}(u)$.
An immediate consequence is that
$
\psi{''}(\langle\mathcal{Z}_i, \mathcal X^* \rangle+t_i\langle\mathcal{Z}_i, \tilde{\mathcal V} \rangle)
\mathbb{I}[|\langle \mathcal{Z}_i, \tilde{\mathcal V}\rangle|\leq 2\tau]
\mathbb{I}\left[\left|\langle \mathcal{Z}_i, \mathcal X^*\rangle\right|\leq T\right]
\geq \ell_\psi(T)
\mathbb{I}[|\langle \mathcal{Z}_i, \tilde{\mathcal V}\rangle|\leq 2\tau]
\mathbb{I}\left[\left|\langle \mathcal{Z}_i, \mathcal X^*\rangle\right|\leq T\right]
$.
This together with (\ref{eq72}) leads to
\begin{equation}\label{eq66}
	\langle\nabla f_{n,y}(\mathcal X^*+\tilde{\mathcal V})
	-\nabla f_{n,y}(\mathcal X^* ), \tilde{\mathcal V}\rangle
	\geq
	\ell_\psi(T) \frac{1}{n} \sum_{i=1}^n
	\langle\mathcal{Z}_i, \tilde{\mathcal V}\rangle^2
	\mathbb{I}[|\langle \mathcal{Z}_i, \tilde{\mathcal V}\rangle|\leq 2\tau]
	\mathbb{I}\left[\left|\langle \mathcal{Z}_i, \mathcal X^*\rangle\right|\leq T\right]
	.
\end{equation}

Next, we will demonstrate that, for $\varepsilon_0\geq0$,
\begin{equation}\label{ineq11}
	\begin{aligned}
		\frac{1}{n} \sum_{i=1}^n
		\langle\mathcal{Z}_i, \tilde{\mathcal V}\rangle^2
		\mathbb{I}[|\langle \mathcal{Z}_i, \tilde{\mathcal V}\rangle|\leq 2\tau]
		\mathbb{I}\left[\left|\langle \mathcal{Z}_i, \mathcal X^*\rangle\right|\leq T\right]
		\geq
		\frac{\tau_u}{2} \|\tilde{\mathcal V}\|_F^2-\varepsilon_0
	\end{aligned}
\end{equation}
with high probability, where  $\tau_u$ is defined in (\ref{ineq1}).
Note that $\tau=K\delta$.
For notational convenience, we define
\begin{equation}\label{ineq2}
	\begin{aligned}
		\varpi_{\tau(\delta)}(\langle\mathcal{Z}_i, \tilde{\mathcal V}\rangle)&:=\langle\mathcal{Z}_i, \tilde{\mathcal V}\rangle^2\mathbb{I}[|\langle \mathcal{Z}_i, \tilde{\mathcal V}\rangle|\leq 2\tau].
	\end{aligned}
\end{equation}
In particular, if $\delta=1$,
then $\varpi_{\tau(1)}(\langle\mathcal{Z}_i, \tilde{\mathcal V}\rangle)=\langle\mathcal{Z}_i, \tilde{\mathcal V}\rangle^2\mathbb{I}[|\langle \mathcal{Z}_i, \tilde{\mathcal V}\rangle|\leq 2K]$.

By using the notation in (\ref{ineq2}), (\ref{ineq11}) can be rewritten as
\begin{equation}\label{ineq16}
	\begin{aligned}
		\frac{1}{n} \sum_{i=1}^n
		\varpi_{\tau(\delta)}(\langle\mathcal{Z}_i, \tilde{\mathcal V}\rangle)
		\mathbb{I}\left[\left|\langle \mathcal{Z}_i, \mathcal X^*\rangle\right|\leq T\right]
		\geq
		\frac{\tau_u}{2} \|\tilde{\mathcal V}\|_F^2-\varepsilon_0,\quad
		\text{for any $\varepsilon_0\geq0$}.
	\end{aligned}
\end{equation}
Furthermore, we claim that the inequality in ($\ref{ineq16}$) can be reduced to the case  for $\delta = 1$.
Let $\tilde{\mathcal V}_1=\frac{\tilde{\mathcal V}}{\|\tilde{\mathcal V}\|_F}=\frac{\tilde{\mathcal V}}{\delta}$, then $\|\tilde{\mathcal V}_1\|_F=1$.
Note that
\begin{equation}\label{ineq13}
	\begin{aligned}
		\varpi_{\tau(\delta)}(\langle\mathcal{Z}_i, \tilde{\mathcal V}\rangle)
		=\langle\mathcal{Z}_i, \tilde{\mathcal V}\rangle^2\mathbb{I}[|\langle \mathcal{Z}_i, \tilde{\mathcal V}\rangle|\leq 2K\delta]
		=\delta^2\langle\mathcal{Z}_i, \tilde{\mathcal V}_1\rangle^2\mathbb{I}[|\langle \mathcal{Z}_i, \tilde{\mathcal V}_1\rangle|\leq 2K]
		=\delta^2\varpi_{\tau(1)}(\langle\mathcal{Z}_i, \tilde{\mathcal V}_1\rangle).
	\end{aligned}
\end{equation}
Then (\ref{ineq16}) is equivalent to the following form
\begin{equation}\label{ineq12}
	\begin{aligned}
		\frac{1}{n} \sum_{i=1}^n
		\varpi_{\tau(1)}(\langle\mathcal{Z}_i, \tilde{\mathcal V}_1\rangle)
		\mathbb{I}\left[\left|\langle \mathcal{Z}_i, \mathcal X^*\rangle\right|\leq T\right]
		\geq
		\frac{\tau_u}{2} -\frac{\varepsilon_0}{\delta^2},\quad
		\text{for any $\varepsilon_0\geq0$}.
	\end{aligned}
\end{equation}
In the following, we begin to prove that $(\ref{ineq12})$.
Define $x_i:=
\varpi_{\tau(\delta)}(\langle\mathcal{Z}_i, \tilde{\mathcal V}\rangle) \mathbb{I}\left[\left|\langle \mathcal{Z}_i, \mathcal X^*\rangle\right|\leq T\right]$.
For any $i$,
one can easily check the random variable $x_i\in[0, 4K^2]$.
Note that the bounded random variables are sub-Gaussian \cite[Example 2.4]{Wainwright_2019}, which
implies that
$x_i$ is sub-Gaussian.
Furthermore, we can obtain that $x_i$ is also sub-exponential \cite[Definition 2.7]{Wainwright_2019}.
For any $\varepsilon_0\geq 0$,
\begin{equation}\label{ineq4}
	\begin{aligned}
		&\mathbb{P}\left(\frac{1}{n} \sum_{i=1}^n
		\varpi_{\tau(1)}(\langle\mathcal{Z}_i, \tilde{\mathcal V}_1\rangle) \mathbb{I}\left[\left|\langle \mathcal{Z}_i, \mathcal X^*\rangle\right|\leq T\right]
		\geq
		\frac{1}{n} \sum_{i=1}^n
		\mathbb{E}\left(
		\varpi_{\tau(1)}(\langle\mathcal{Z}_i, \tilde{\mathcal V}_1\rangle) \mathbb{I}\left[\left|\langle \mathcal{Z}_i, \mathcal X^*\rangle\right|\leq T\right]\right)-\frac{\varepsilon_0}{\delta^2}\right)\\
		&=\mathbb{P}\left(\frac{1}{n} \sum_{i=1}^n
		\varpi_{\tau(\delta)}(\langle\mathcal{Z}_i, \tilde{\mathcal V}\rangle) \mathbb{I}\left[\left|\langle \mathcal{Z}_i, \mathcal X^*\rangle\right|\leq T\right]
		\geq
		\frac{1}{n} \sum_{i=1}^n
		\mathbb{E}\left(
		\varpi_{\tau(\delta)}(\langle\mathcal{Z}_i, \tilde{\mathcal V}\rangle) \mathbb{I}\left[\left|\langle \mathcal{Z}_i, \mathcal X^*\rangle\right|\leq T\right]\right)-\varepsilon_0\right)\\
		&=\mathbb{P}\left(\frac{1}{n} \sum_{i=1}^n x_i\geq \frac{1}{n} \sum_{i=1}^n \mathbb{E}(x_i)-\varepsilon_0\right)\\
		&\geq
		1-\exp\left(-\frac{n\varepsilon_0^2}{\frac{2}{n} \sum_{i=1}^n \mathbb{E}(x_i^2)}\right),
	\end{aligned}
\end{equation}
where the first equality holds by (\ref{ineq13}), the second equality holds by the definition of $x_i$,
and the inequality holds by the Bernstein's inequality \cite[Propostion 2.14]{Wainwright_2019}.
Suppose that we can prove that
\begin{equation}\label{ineq6}
	\mathbb{E}\left(
	x_i\right)=\mathbb{E}\left(
	\varpi_{\tau(\delta)}(\langle\mathcal{Z}_i, \tilde{\mathcal V}\rangle)
	\mathbb{I}\left[\left|\langle \mathcal{Z}_i, \mathcal X^*\rangle\right|\leq T\right]\right)\geq
	\frac{\tau_u}{2} \|\tilde{\mathcal V}\|_F^2,
\end{equation}
which taken collectively with (\ref{ineq2}), (\ref{ineq16}), (\ref{ineq4}) yields the
desired claim (\ref{ineq11}).

Note that (\ref{ineq6}) can be written equivalently in the following form
\begin{equation}\label{ineq17}
	\mathbb{E}\left(
	\varpi_{\tau(1)}(\langle\mathcal{Z}_i, \tilde{\mathcal V}_1\rangle)
	\mathbb{I}\left[\left|\langle \mathcal{Z}_i, \mathcal X^*\rangle\right|\leq T\right]\right)\geq
	\frac{\tau_u}{2}.
\end{equation}
Observe that
\begin{equation}\label{fin1}
	\begin{aligned}
	~	&\mathbb{E}\left(
		\varpi_{\tau(1)}(\langle\mathcal{Z}_i, \tilde{\mathcal V}_1\rangle)
		\mathbb{I}\left[\left|\langle \mathcal{Z}_i, \mathcal X^*\rangle\right|\leq T\right]\right)\\
		= \ &\mathbb{E}\left(\langle\mathcal{Z}_i, \tilde{\mathcal V}_1\rangle^2\mathbb{I}[|\langle \mathcal{Z}_i, \tilde{\mathcal V}_1\rangle|\leq 2K]
		\mathbb{I}\left[\left|\langle \mathcal{Z}_i, \mathcal X^*\rangle\right|\leq T\right]\right)\\
		= \ &\mathbb{E}\left(\langle\mathcal{Z}_i, \tilde{\mathcal V}_1\rangle^2\mathbb{I}[|\langle \mathcal{Z}_i, \tilde{\mathcal V}_1\rangle|\leq 2K]\right)
		-\mathbb{E}\left(\langle\mathcal{Z}_i, \tilde{\mathcal V}_1\rangle^2\mathbb{I}[|\langle \mathcal{Z}_i, \tilde{\mathcal V}_1\rangle|\leq 2K]
		\mathbb{I}\left[\left|\langle \mathcal{Z}_i, \mathcal X^*\rangle\right|> T\right]\right)\\
		= \ &\mathbb{E}\left(\langle\mathcal{Z}_i, \tilde{\mathcal V}_1\rangle^2\right)-
		\mathbb{E}\left(\langle\mathcal{Z}_i, \tilde{\mathcal V}_1\rangle^2 \mathbb{I}[|\langle \mathcal{Z}_i, \tilde{\mathcal V}_1\rangle|>2K]\right)\\
		\ & -\mathbb{E}\left(\langle\mathcal{Z}_i, \tilde{\mathcal V}_1\rangle^2\mathbb{I}[|\langle \mathcal{Z}_i, \tilde{\mathcal V}_1\rangle|\leq 2K]
		\mathbb{I}\left[\left|\langle \mathcal{Z}_i, \mathcal X^*\rangle\right|> T\right]\right)\\
		\geq \  &\tau_u-
		\mathbb{E}\left(\langle\mathcal{Z}_i, \tilde{\mathcal V}_1\rangle^2 \mathbb{I}[|\langle \mathcal{Z}_i, \tilde{\mathcal V}_1\rangle|>2K]\right)\\
		\ & -\mathbb{E}\left(\langle\mathcal{Z}_i, \tilde{\mathcal V}_1\rangle^2\mathbb{I}[|\langle \mathcal{Z}_i, \tilde{\mathcal V}_1\rangle|\leq 2K]
		\mathbb{I}\left[\left|\langle \mathcal{Z}_i, \mathcal X^*\rangle\right|> T\right]\right),
	\end{aligned}
\end{equation}
where the first equality follows from (\ref{ineq13}) and
the inequality holds by the condition $\mathbb{E}\left(\langle\mathcal{Z}_i, \tilde{\mathcal V}_1\rangle^2\right)\geq \tau_u$.
By Cauchy-Schwarz inequality, we get that
\begin{equation}\label{fin3}
	\begin{aligned}
		\mathbb{E}\left(\langle\mathcal{Z}_i, \tilde{\mathcal V}_1\rangle^2 \mathbb{I}[|\langle \mathcal{Z}_i, \tilde{\mathcal V}_1\rangle|>2K]\right)
		& \leq \sqrt{\mathbb{E}\left(\langle\mathcal{Z}_i, \tilde{\mathcal V}_1\rangle^4\right)}\sqrt{\mathbb{P}\left(|\langle \mathcal{Z}_i, \tilde{\mathcal V}_1\rangle|>2K\right)} \\
		& \leq  \sqrt{\tau_l} \sqrt{\mathbb{P}\left(|\langle \mathcal{Z}_i, \tilde{\mathcal V}_1\rangle|>2K\right)},
	\end{aligned}
\end{equation}
where the last inequality follows from (\ref{ineq1}).
On the other hand, we can deduce
\begin{equation}\label{ProZV}
\begin{aligned}
	\mathbb{P}\left(|\langle \mathcal{Z}_i, \tilde{\mathcal V}_1\rangle|>2K\right)                 =\mathbb{P}\left(|\langle \mathcal{Z}_i, \tilde{\mathcal V}_1\rangle|^4>16K^4\right)
	\leq \frac{\mathbb{E}(\langle\mathcal{Z}, \tilde{\mathcal V}_1\rangle^4)}{16K^4}
	\leq \frac{\tau_l}{16K^4},
\end{aligned}
\end{equation}
where the first inequality follows from Markov inequality \cite[Theorem 7.3]{Foucart2013}
and the second inequality follows from (\ref{ineq1}).
Set   $K^2\geq\frac{\tau_l}{\tau_u}$. Combining (\ref{ProZV})
 with (\ref{fin3}) yields
\begin{equation}\label{EZVK}
\begin{aligned}
	\mathbb{E}\left(\langle\mathcal{Z}_i, \tilde{\mathcal V}_1\rangle^2 \mathbb{I}[|\langle \mathcal{Z}_i, \tilde{\mathcal V}_1\rangle|>2K]\right)
	\leq \frac{\tau_l}{4K^2}\leq \frac{\tau_u}{4}.
\end{aligned}
\end{equation}
By a similar discussion, we obtain
\begin{equation}\label{ineq8}
	\begin{aligned}
		\mathbb{P}\left(|\langle \mathcal{Z}_i, \mathcal X^*\rangle|>T\right)
		\leq \frac{\mathbb{E}\left(\langle\mathcal{Z}_i, \mathcal X^*\rangle^4\right)}{T^4}
		= \frac{\|\mathcal X^*\|_F^4 \mathbb{E}\left(\langle\mathcal{Z}_i, \frac{\mathcal X^*}{\|\mathcal X^*\|_F}\rangle^4\right)}{T^4}
		\leq  \frac{\|\mathcal X^*\|_F^4 \tau_l}{T^4}.
	\end{aligned}
\end{equation}
By Cauchy-Schwarz inequality, one can easily get
$$
\begin{aligned}
	&\mathbb{E}\left(\langle\mathcal{Z}_i, \tilde{\mathcal V}_1\rangle^2\mathbb{I}[|\langle \mathcal{Z}_i, \tilde{\mathcal V}_1\rangle|\leq 2K]
	\mathbb{I}\left[\left|\langle \mathcal{Z}_i, \mathcal X^*\rangle\right|> T\right]\right)\\
	\leq\  &\sqrt{\mathbb{E}\left(\left(\langle\mathcal{Z}_i, \tilde{\mathcal V}_1\rangle\right)^2\mathbb{I}[|\langle \mathcal{Z}_i, \tilde{\mathcal V}_1\rangle|\leq 2K]\right)^2}
	\sqrt{\mathbb{P}\left(\left|\langle \mathcal{Z}_i, \mathcal X^*\rangle\right|> T\right)}\\
	\leq \  &\sqrt{\mathbb{E}\left(\langle\mathcal{Z}_i, \tilde{\mathcal V}_1\rangle\right)^4}
	\sqrt{\mathbb{P}\left(\left|\langle \mathcal{Z}_i, \mathcal X^*\rangle\right|> T\right)}
	\leq   \frac{\tau_l \|\mathcal X^*\|_F^2}{T^2},
\end{aligned}
$$
where the last inequality follows from (\ref{ineq8}) and (\ref{ineq1}).
Taking $T^2\geq\frac{4\|\mathcal X^*\|_F^2 \tau_l}{\tau_u}$ yields
\begin{equation}\label{ineq9}
	\begin{aligned}
		\mathbb{E}\left(\langle\mathcal{Z}_i, \tilde{\mathcal V}_1\rangle^2\mathbb{I}[|\langle \mathcal{Z}_i, \tilde{\mathcal V}_1\rangle|\leq 2K]
		\mathbb{I}\left[\left|\langle \mathcal{Z}_i, \mathcal X^*\rangle\right|> T\right]\right)
		\leq \frac{\tau_u}{4}.
	\end{aligned}
\end{equation}
Combining (\ref{fin1}), (\ref{EZVK}) and (\ref{ineq9}) then gives
\begin{equation}\label{ineq3}
	\mathbb{E}\left(
	\varpi_{\tau(1)}(\langle\mathcal{Z}_i, \tilde{\mathcal V}_1\rangle)
	\mathbb{I}\left[\left|\langle \mathcal{Z}_i, \mathcal X^*\rangle\right|\leq T\right]\right)\geq \tau_u-\frac{\tau_u}{4}-\frac{\tau_u}{4}=\frac{\tau_u}{2}.
\end{equation}
With the above analysis,
we can establish the desired result (\ref{ineq17}).
From (\ref{ineq4}), (\ref{ineq6}) and (\ref{ineq17}),
for any $\|\tilde{\mathcal V}\|_F\in(0,1]$, we get
\begin{equation}\label{ineq15}
	\begin{aligned}
		&\mathbb{P}\left(\frac{1}{n} \sum_{i=1}^n
		\varpi_{\tau(\delta)}(\langle\mathcal{Z}_i, \tilde{\mathcal V}\rangle) \mathbb{I}\left[\left|\langle \mathcal{Z}_i, \mathcal X^*\rangle\right|\leq T\right]
		\geq \frac{\tau_u}{2} \|\tilde{\mathcal V}\|_F^2-\varepsilon_0\right)
		\geq
		1-\exp\left(-\frac{n\varepsilon_0^2}{\frac{2}{n} \sum_{i=1}^n \mathbb{E}(x_i^2)}\right).
	\end{aligned}
\end{equation}
Combining this with (\ref{ineq2}) yields the claim (\ref{ineq11}).
Taking this collectively with (\ref{eq66}), we arrive at
\begin{equation}\label{ineq14}
	\begin{aligned}
		&\mathbb{P}\left(\langle\nabla f_{n,y}(\mathcal X^*+\tilde{\mathcal V})
		-\nabla f_{n,y}(\mathcal X^* ), \tilde{\mathcal V}\rangle
		\geq\frac{ \ell_\psi(T)\tau_u}{2} \|\tilde{\mathcal V}\|_F^2-\ell_\psi(T)\varepsilon_0\right) \\
		\geq \ &
		1-\exp\left(-\frac{n\varepsilon_0^2}{\frac{2}{n} \sum_{i=1}^n \mathbb{E}(x_i^2)}\right).
	\end{aligned}
\end{equation}
Note that
$$
x_i^2=\langle\mathcal{Z}_i, \tilde{\mathcal V}\rangle^4
\mathbb{I}[|\langle \mathcal{Z}_i, \tilde{\mathcal V}\rangle|\leq 2\tau] \mathbb{I}\left[\left|\langle \mathcal{Z}_i, \mathcal X^*\rangle\right|\leq T\right].
$$
Denote $z_i:=\langle \mathcal{Z}_i, \tilde{\mathcal V}\rangle=\langle \operatorname{vec}(\mathcal{Z}_i), \operatorname{vec}(\tilde{\mathcal V})\rangle, i=1,2,\ldots,n$.
Note that $\mathcal{Z}_i$ are sub-Gaussian with i.i.d. zero-mean entries
with same sub-Gaussian parameter $\tilde b$ in (\ref{Pro1}), which  implies that the random variable
$z_i=\langle \operatorname{vec}(\mathcal{Z}_i), \operatorname{vec}(\tilde{\mathcal V})\rangle$ is sub-Gaussian with parameter $\tilde{b}\|\tilde{\mathcal V}\|_F^2$ \cite[Theorem 7.27]{Foucart2013}.
Then we have
$$
\begin{aligned}
	\mathbb{E}(x_i^2)
	&=\mathbb{E}\left(\langle\mathcal{Z}_i, \tilde{\mathcal V}\rangle^4
	\mathbb{I}[|\langle \mathcal{Z}_i, \tilde{\mathcal V}\rangle|\leq 2\tau] \mathbb{I}\left[\left|\langle \mathcal{Z}_i, \mathcal X^*\rangle\right|\leq T\right])\right)\\
	&\leq  \sqrt{\mathbb{E}\left(\langle\mathcal{Z}_i, \tilde{\mathcal V}\rangle^4\mathbb{I}[|\langle \mathcal{Z}_i, \tilde{\mathcal V}\rangle|\leq 2\tau]\right)^2}
	\sqrt{\mathbb{P}(\left|\langle \mathcal{Z}_i, \mathcal X^*\rangle\right|\leq T)}\\
	&\leq \sqrt{\mathbb{E}\left(\langle\mathcal{Z}_i, \tilde{\mathcal V}\rangle^4\mathbb{I}[|\langle \mathcal{Z}_i, \tilde{\mathcal V}\rangle|\leq 2\tau]\right)^2}\\
	&\leq \sqrt{\mathbb{E}( \langle\mathcal{Z}_i, \tilde{\mathcal V}\rangle^8)}
	\leq 256 \tilde{b}^2\|\tilde{\mathcal V}\|_F^4,
\end{aligned}
$$
where the first inequality follows from the Cauchy-Schwarz inequality and
the last inequality follows from \cite[Proposition 3.2]{Rivasplata2012}.
Consequently,
\begin{equation}\label{ineq10}
	-\frac{n\varepsilon_0^2}{\frac{2}{n} \sum_{i=1}^n \mathbb{E}(x_i^2)}\leq
	-\frac{n\varepsilon_0^2}{\frac{2}{n} \sum_{i=1}^n 256 \tilde{b}^2   \|\tilde{\mathcal V}\|_F^4}=-\frac{n\varepsilon_0^2}{512\tilde{b}^2  \|\tilde{\mathcal V}\|_F^4}=-\frac{n\varepsilon_0^2}{\kappa  \|\tilde{\mathcal V}\|_F^4},
\end{equation}
where  $\kappa:=512\tilde{b}^2$.
Let
$$
\varepsilon_0 = c_2\sqrt{\frac{\log d}{ n}}\|\tilde{\mathcal V}\|_{\text{TTNN}}\|\tilde{\mathcal V}\|_{F},
$$
where $c_2> 0$ is a given constant.
By using the arithmetic mean-geometric mean inequality,
we have
$$
\begin{aligned}
	\varepsilon_0=c_2 \sqrt{\frac{\log d}{ n}}\|\tilde{\mathcal V}\|_{\text{TTNN}}\|\tilde{\mathcal V}\|_{F}
	=
	\sqrt{\frac{2c_2^2}{\tau_u  } \frac{\log d}{ n}\|\tilde{\mathcal V}\|_{\text{TTNN}}^2 \cdot \frac{\tau_u }{2}\|\tilde{\mathcal V}\|_{F}^2}
	\leq \frac{c_2^2}{\tau_u  } \frac{\log d}{ n}\|\tilde{\mathcal V}\|_{\text{TTNN}}^2 + \frac{\tau_u }{4}\|\tilde{\mathcal V}\|_{F}^2.
\end{aligned}
$$
Consequently, we deduce
\begin{equation}\label{LTtau}
\begin{aligned}
	\frac{\ell_\psi(T) \tau_u}{2}  \|\tilde{\mathcal V}\|_F^2 - \ell_\psi(T)\varepsilon_0
	&\geq \frac{\ell_\psi(T) \tau_u}{2}  \|\tilde{\mathcal V}\|_F^2-\frac{c_2^2 \ell_\psi(T)}{\tau_u} \frac{\log d}{ n}\|\tilde{\mathcal V}\|_{\text{TTNN}}^2 -  \frac{ \ell_\psi(T)\tau_u}{4}\|\tilde{\mathcal V}\|_{F}^2\\
	&= \frac{\ell_\psi(T) \tau_u}{4}  \|\tilde{\mathcal V}\|_F^2-\frac{c_2^2 \ell_\psi(T) }{\tau_u  } \frac{\log d}{ n}\|\tilde{\mathcal V}\|_{\text{TTNN}}^2.
\end{aligned}
\end{equation}
Combining (\ref{LTtau}), (\ref{ineq10}), and (\ref{ineq14}) yields
\begin{equation}\label{Ca1Re}
\begin{aligned}
	&\mathbb{P}\left(\langle\nabla f_{n,y}(\mathcal X^*+\tilde{\mathcal V})
	-\nabla f_{n,y}(\mathcal X^* ), \tilde{\mathcal V}\rangle
	\geq \frac{\ell_\psi(T) \tau_u}{4}  \|\tilde{\mathcal V}\|_F^2-\frac{c_2^2 \ell_\psi(T)}{\tau_u  } \frac{\log d}{ n}\|\tilde{\mathcal V}\|_{\text{TTNN}}^2\right)\\
	\geq\ & 1-\exp \left(-\frac{n\varepsilon_0^2}{\kappa\|\tilde{\mathcal V}\|_F^4}\right)\\
	=\ & 1-\exp \left(-\frac{ c_2^2 \log d\|\tilde{\mathcal V}\|_{\text{TTNN}}^2}{\kappa\|\tilde{\mathcal V}\|_F^2}\right)\\
	\geq \ & 1-\exp \left(-\frac{c_2 ^2  \log d }{\kappa }\right)
	= 1-\exp \left(-c_3 \log d \right),
\end{aligned}
\end{equation}
where the second inequality holds by $\|\tilde{\mathcal V}\|_{\text{TTNN}}^2\geq \|\tilde{\mathcal V}\|_F^2$ and the last equality holds by letting $c_3:=\frac{c_2 ^2}{\kappa}$.

{\bf Case II}. Suppose that $\|\tilde{\mathcal V}\|_{F}>1$.
Similar to \cite[Lemma 8]{loh15a}, we assume  $\|\tilde{\mathcal V}\|_{\text{TTNN}}\leq 2t$.
Define $L(\gamma):[0,1]\rightarrow \mathbb{R}$ as
$L(\gamma):=f(\mathcal X^*+\gamma\tilde{\mathcal V})$.
Notice that $L(\gamma)$ is  convex by the convexity of $f$, which implies that $L'$ is monotonously non-decreasing.
Consequently, for any $\gamma\in[0,1]$, one has
$
L'(1)-L'(0)\geq L'(\gamma)-L'(0).
$
Then we have
$$
\langle\nabla f_{n,y}(\mathcal X^*+\tilde{\mathcal V})
-\nabla f_{n,y}(\mathcal X^* ), \tilde{\mathcal V}\rangle\geq
\frac{1}{\gamma}\langle\nabla f_{n,y}(\mathcal X^*+\gamma\tilde{\mathcal V})
-\nabla f_{n,y}(\mathcal X^* ), \gamma\tilde{\mathcal V}\rangle.
$$
Taking $\gamma=\frac{1}{\|\tilde{\mathcal V}\|_{F}}\in(0,1]$,
we conclude that
\begin{equation}\label{Ca2Reh}
\begin{aligned}
	\langle\nabla f_{n,y}(\mathcal X^*+\tilde{\mathcal V})
	-\nabla f_{n,y}(\mathcal X^* ), \tilde{\mathcal V}\rangle
	&\geq \|\tilde{\mathcal V}\|_F\left(\frac{\ell_\psi(T) \tau_u}{4}  -\frac{c_2^2\ell_\psi(T)}{\tau_u  } \frac{\log d}{ n}\frac{\|\tilde{\mathcal V}\|_{\text{TTNN}}^2}{\|\tilde{\mathcal V}\|_{F}^2}\right)\\
	&\geq \|\tilde{\mathcal V}\|_F\left(\frac{\ell_\psi(T) \tau_u}{4} -
	\frac{2tc_2^2 \ell_\psi(T) }{\tau_u  } \frac{\log d}{ n}\frac{\|\tilde{\mathcal V}\|_{\text{TTNN}}}{\|\tilde{\mathcal V}\|_{F}^2}\right)\\
	&\geq \|\tilde{\mathcal V}\|_F\left(\frac{\ell_\psi(T) \tau_u}{4}-
	\frac{\ell_\psi(T)}{\tau_u } \sqrt{\frac{\log d}{ n}}\frac{\|\tilde{\mathcal V}\|_{\text{TTNN}}}{\|\tilde{\mathcal V}\|_{F}^2}\right)\\
	&\geq \|\tilde{\mathcal V}\|_F\left(\frac{\ell_\psi(T) \tau_u}{4}-
	\frac{\ell_\psi(T)}{\tau_u  } \sqrt{\frac{\log d}{ n}}\frac{\|\tilde{\mathcal V}\|_{\text{TTNN}}}{\|\tilde{\mathcal V}\|_{F}}\right)\\
	&=\frac{\ell_\psi(T) \tau_u}{4}\|\tilde{\mathcal V}\|_F-
	\frac{\ell_\psi(T)}{\tau_u } \sqrt{\frac{\log d}{ n}}\|\tilde{\mathcal V}\|_{\text{TTNN}},
\end{aligned}
\end{equation}
where the second inequality holds since the assumption $\|\tilde{\mathcal V}\|_{\text{TTNN}}\leq 2t$, the third inequality follows from
$n\geq 4t^2c_2^4\log d$, and the last inequality holds since
$\|\tilde{\mathcal V}\|_{F}^2\geq \|\tilde{\mathcal V}\|_{F}>1$.

Therefore, by combining (\ref{Ca1Re}) with (\ref{Ca2Reh}), and taking $\alpha_1=\alpha_2=\frac{\ell_\psi(T)\tau_u}{4}, \tau_1=\frac{c_2^2 \ell_\psi(T)}{\tau_u }, \tau_2=\frac{\ell_\psi(T)}{\tau_u}$, we get that
the loss function (\ref{eq61}) satisfies the RSC condition (\ref{test0})
 with probability at least $1-\exp (-c_3 \log d )$ .
\qed

\section*{Appendix C. Auxiliary Lemmas}\label{Appx}

\begin{lemma}\label{lemgra}
	Denote the function $h(x):=\frac{1}{1+\exp (-x)}$, where $x\in\mathbb{R}$.
	Then
	$$
	|h(x_1)-h(x_2)|\leq \frac{1}{4}|x_1-x_2|, \ \  \forall  x_1,x_2\in\mathbb{R}.
	$$
\end{lemma}
\textbf{Proof.}
For simplicity, we assume $x_1<x_2$.
Then there exists $x_3\in(x_1,x_2)$ such that
\begin{equation}\label{lip3}
	|h(x_1)-h(x_2)|=| h'(x_3)(x_1-x_2)|.
\end{equation}
Note that for any $x\in\mathbb{R}$, we have
\begin{equation}\label{lip4}
	h'(x)=\frac{\exp (x)}{(1+\exp (x))^2}=\frac{1}{2+\exp (x)+\exp (-x)}\leq \frac{1}{4},
\end{equation}
where the   inequality follows from the fact that $\exp (x)+\exp (-x)\geq 2\sqrt{\exp (x)\cdot \exp (-x)}=2$.
In addition, it is worth noting that $h'(x)>0$.
Taking (\ref{lip3}) collectively with (\ref{lip4}) shows that
$$
\begin{aligned}
	|h(x_1)-h(x_2)|\leq \frac{1}{4}|x_1-x_2|.
\end{aligned}
$$
This completes the proof.
\qed

Note that the gradient of $f_{n,y}(\mathcal{X})$ in (\ref{eq61}) is given by
\begin{equation}\label{eq70}
	\nabla f_{n,y}(\mathcal{X})=\frac{1}{n} \sum_{i=1}^n
	\left[
	\left(\frac{\exp (\langle \mathcal{Z}_i, \mathcal X\rangle)}{1+\exp \left(\left\langle \mathcal{Z}_i, \mathcal X\right\rangle\right)}-y_i\right)\mathcal{Z}_i
	\right].
\end{equation}
Now we show that $	\nabla f_{n,y}$ is Lipschitz continuous in the following lemma.

\begin{lemma}\label{lem22}
	The gradient $\nabla f_{n,y}$ in (\ref{eq70}) is Lipschitz continuous with Lipschitz constant $\frac{1}{4n} \sum_{i=1}^n\|\mathcal{Z}_i\|_F^2$, i.e.,
$$
\begin{aligned}
	\|\nabla f_{n,y}(\mathcal X_1)
	-\nabla f_{n,y}(\mathcal X_2 )\|_F
	&\leq \frac{1}{4n} \sum_{i=1}^n\|\mathcal{Z}_i\|_F^2\|\mathcal X_1-\mathcal X_2\|_F, \ \forall \ \mathcal X_1, \mathcal X_2\in\mathbb{R}^{n_1\times n_2\times n_3}.
\end{aligned}
$$
\end{lemma}
\textbf{Proof.}
For any tensor $\mathcal X_1, \mathcal X_2\in\mathbb{R}^{n_1\times n_2\times n_3}$,
we obtain
\begin{equation}\label{lip1}
	\nabla f_{n,y}(\mathcal X_1)
	-\nabla f_{n,y}(\mathcal X_2 )
	=\frac{1}{n} \sum_{i=1}^n
	\left[
	\left(\frac{\exp (\langle \mathcal{Z}_i, \mathcal X_1\rangle)}{1+\exp (\langle \mathcal{Z}_i, \mathcal X_1\rangle)}-\frac{\exp (\langle \mathcal{Z}_i, \mathcal X_2\rangle)}{1+\exp \left(\left\langle \mathcal{Z}_i, \mathcal X_2\right\rangle\right)}\right)\mathcal{Z}_i\right].
\end{equation}
Let
$
h(\langle \mathcal{Z}_i, \mathcal X\rangle)=\frac{1}{1+\exp (-\langle \mathcal{Z}_i, \mathcal X\rangle)}.
$
Then (\ref{lip1}) can be equivalently expressed as
$$
\nabla f_{n,y}(\mathcal X_1)
-\nabla f_{n,y}(\mathcal X_2 )
=\frac{1}{n} \sum_{i=1}^n
\left[
\left(h(\langle \mathcal{Z}_i, \mathcal X_1\rangle)-h(\langle \mathcal{Z}_i, \mathcal X_2\rangle)\right)\mathcal{Z}_i\right].
$$
Thus, we get 
\begin{equation}\label{lip2}
	\begin{aligned}
		\|\nabla f_{n,y}(\mathcal X_1)
		-\nabla f_{n,y}(\mathcal X_2 )\|_F
		&=\frac{1}{n} \bigg\|\sum_{i=1}^n
		\left[
		\left(h(\langle \mathcal{Z}_i, \mathcal X_1\rangle)-h(\langle \mathcal{Z}_i, \mathcal X_2\rangle)\right)\mathcal{Z}_i\right]\bigg\|_F\\
		&\leq\frac{1}{n} \sum_{i=1}^n \bigg\|
		\left(h(\langle \mathcal{Z}_i, \mathcal X_1\rangle)-h(\langle \mathcal{Z}_i, \mathcal X_2\rangle)\right)\mathcal{Z}_i\bigg\|_F\\
		&=\frac{1}{n} \sum_{i=1}^n
		|h(\langle \mathcal{Z}_i, \mathcal X_1\rangle)-h(\langle \mathcal{Z}_i, \mathcal X_2\rangle)|
		\|\mathcal{Z}_i\|_F,
	\end{aligned}
\end{equation}
where the inequality holds by the Minkowski's inequality.
Using Lemma \ref{lemgra} further leads to
\begin{equation}\label{hhd}
\begin{aligned}
	|h(\langle \mathcal{Z}_i, \mathcal X_1\rangle)-h(\langle \mathcal{Z}_i, \mathcal X_2\rangle)|
	\|\mathcal{Z}_i\|_F
	&\leq \frac{1}{4}|\langle \mathcal{Z}_i, \mathcal X_1\rangle-\langle \mathcal{Z}_i, \mathcal X_2\rangle|
	\|\mathcal{Z}_i\|_F\\
	&= \frac{1}{4}|\langle \mathcal{Z}_i, \mathcal X_1-\mathcal X_2\rangle|
	\|\mathcal{Z}_i\|_F\\
	&\leq \frac{1}{4}\|\mathcal{Z}_i\|_F\|\mathcal X_1-\mathcal X_2\|_F
	\|\mathcal{Z}_i\|_F\\
	&= \frac{1}{4}\|\mathcal{Z}_i\|_F^2\|\mathcal X_1-\mathcal X_2\|_F,
\end{aligned}
\end{equation}
where the second inequality follows from the Cauchy-Schwarz inequality.
Plugging (\ref{hhd}) into (\ref{lip2}) immediately yields
$$
\begin{aligned}
	\|\nabla f_{n,y}(\mathcal X_1)
	-\nabla f_{n,y}(\mathcal X_2 )\|_F
	&\leq \frac{1}{4n} \sum_{i=1}^n\|\mathcal{Z}_i\|_F^2\|\mathcal X_1-\mathcal X_2\|_F.
\end{aligned}
$$
This concludes the proof.
\qed

\begin{lemma}\label{lem4}
	For any matrix  $\mathbf{X}\in \mathbb{R}^{n_1\times n_2}$,
	define $
	h(\mathbf X):=\sum_{j=1}^{\min\{n_1,n_2\}}g_\lambda(\sigma_j(\mathbf{X}))$, where $g_\lambda(\cdot)$ satisfies
	Assumption \ref{assum1}. Then
	\begin{equation*}\label{test21}
		\begin{aligned}
			\|\mathbf{X}\|_*
			&\leq
			\frac{h(\mathbf X)}{\lambda k_0}
			+\frac{\mu}{2 \lambda k_0}\|\mathbf{X}\|_F^2.
		\end{aligned}
	\end{equation*}
\end{lemma}
\textbf{Proof.}
Firstly, we prove
\begin{equation}\label{test22}
\begin{aligned}
\lambda k_0x
&\leq
g_\lambda(x)
+\frac{\mu}{2}x^2, \ \textup{for  any} \ x\geq0,
\end{aligned}
\end{equation}
where $k_0,\mu$ are the parameters in Assumption \ref{assum1}.
The above inequality is trivial for $x=0$. For $x>0$, 
it follows from  Assumption \ref{assum1}  that there exists $\mu>0$ such that
$g_\lambda\left(x\right)
+\frac{\mu}{2}x^2$ is convex, which yields
\begin{equation}\label{Gdrt}
g_\lambda\left(x\right)+\frac{\mu}{2}x^2
\geq g_\lambda(x')+\frac{\mu}{2}(x')^2+
\langle g_\lambda^{\prime}(x')+\mu x', x-x' \rangle,
 \  \text{for any } x,x'>0.
\end{equation}
Here $g_\lambda^{\prime}(\cdot)$ represents the derivative of $g_\lambda(\cdot)$.
By taking  $x'\rightarrow0^{+}$ in (\ref{Gdrt}), under Assumption \ref{assum1},
we can get that  (\ref{test22}) holds.

Secondly,
 one has
\begin{equation}\label{test23}
\begin{aligned}
\lambda k_0\|\mathbf{X}\|_*
=
\sum_{j=1}^{\min\{n_1,n_2\}}\lambda k_0\sigma_j(\mathbf{X})
&\leq
\sum_{j=1}^{\min\{n_1,n_2\}}\left(g_\lambda(\sigma_j(\mathbf{X})\right)
+\frac{\mu}{2}(\sigma_j(\mathbf{X}))^2)\\
&=h(\mathbf{X})
+\frac{\mu}{2}\|\mathbf{X}\|_F^2,
\end{aligned}
\end{equation}
where the last equation follows from the fact that $\|\mathbf{X}\|_F^2=\sum_{j=1}^{\min\{n_1,n_2\}}(\sigma_j(\mathbf{X}))^2$. This completes the proof.
\qed

\begin{lemma}\label{lem5}
For any $\mathcal X\in\mathbb {R}^{n_1\times n_2\times n_3}$, define
$\gamma(\mathcal X):=G_\lambda(\mathcal X)+\frac{\mu}{2}\left\|\mathcal X\right\|_F^2$,
where $G_\lambda(\mathcal X)$ is defined in (\ref{test1}) and the parameter $\mu$ is the same  in Assumption \ref{assum1}(iv).
Then  $\gamma(\mathcal X)$ is convex.
\end{lemma}
\textbf{Proof.}
Let $\rho(x):=g_\lambda(x)+\frac{\mu}{2}x^2$ and $m:=\min\{n_1,n_2\}$.
Under Assumption \ref{assum1}(iv),
for any $\mathcal X,\mathcal Y\in\mathbb {R}^{n_1\times n_2\times n_3}$, $i=1,\ldots, n_3, j=1,\ldots, m$,
the convexity of $\rho(x)$ leads to
$$
\begin{aligned}
	&g_\lambda\left(\theta \sigma_j(\widehat{\mathcal X}_\mathbf U^{\langle i\rangle})
	+(1-\theta)\sigma_j(\widehat{\mathcal Y}_\mathbf U^{\langle i\rangle})\right)
	+\frac{\mu}{2}\left(\theta \sigma_j(\widehat{\mathcal X}_\mathbf U^{\langle i\rangle})
	+(1-\theta)\sigma_j(\widehat{\mathcal Y}_\mathbf U^{\langle i\rangle})\right)^2
	\\&\leq
	\theta g_\lambda\left(\sigma_j(\widehat{\mathcal X}_\mathbf U^{\langle i\rangle})\right)
	+\frac{\theta\mu}{2}\left(\sigma_j(\widehat{\mathcal X}_\mathbf U^{\langle i\rangle})\right)^2
	+
	(1-\theta) g_\lambda\left(\sigma_j{(\widehat{\mathcal Y}_\mathbf U^{\langle i\rangle})}\right)
	+\frac{(1-\theta)\mu}{2}\left(\sigma_j(\widehat{\mathcal Y}_\mathbf U^{\langle i\rangle})\right)^2, \forall\theta\in[0,1].
\end{aligned}
$$
Summing up the above inequalities from $i=1,\ldots, n_3, j=1,\ldots, m$ yields
\begin{equation}\label{test64}
	\begin{aligned}
		&\sum_{i=1}^{n_3}\sum_{j=1}^{m}
		g_\lambda\left(\theta \sigma_j(\widehat{\mathcal X}_\mathbf U^{\langle i\rangle})
		+(1-\theta)\sigma_j(\widehat{\mathcal Y}_\mathbf U^{\langle i\rangle})\right)
		+\frac{\mu}{2}\left(\theta \sigma_j(\widehat{\mathcal X}_\mathbf U^{\langle i\rangle})
		+(1-\theta)\sigma_j(\widehat{\mathcal Y}_\mathbf U^{\langle i\rangle})\right)^2\\
		&\leq
		\sum_{i=1}^{n_3}\sum_{j=1}^{m}
		\theta g_\lambda\left(\sigma_j(\widehat{\mathcal X}_\mathbf U^{\langle i\rangle})\right)
		+\frac{\theta\mu}{2}\left(\sigma_j(\widehat{\mathcal X}_\mathbf U^{\langle i\rangle})\right)^2
		+
		(1-\theta) g_\lambda\left(\sigma_j(\widehat{\mathcal Y}_\mathbf U^{\langle i\rangle})\right)
		+\frac{(1-\theta)\mu}{2}\left(\sigma_j(\widehat{\mathcal Y}_\mathbf U^{\langle i\rangle})\right)^2.
	\end{aligned}
\end{equation}
Moreover, by \cite[Corollary 3.4.3]{Roger1991}, we have
$$
\begin{aligned}
	& \sum_{j=1}^m
	\sigma_j\left(\theta\widehat{\mathcal X}_\mathbf U^{\langle i\rangle}+(1-\theta)\widehat{\mathcal Y}_\mathbf U^{\langle i\rangle}\right)
	\leq
	\sum_{j=1}^m
	\sigma_j\left(\theta\widehat{\mathcal X}_\mathbf U^{\langle i\rangle}\right)+
	\sigma_j\left((1-\theta)\widehat{\mathcal Y}_\mathbf U^{\langle i\rangle}\right), \  i=1,2,\ldots, n_3.
\end{aligned}
$$
Note that
$\rho(x)=g_\lambda(x)+\frac{\mu}{2}x^2$ is
a real-valued convex function on $[0,+\infty)$.
Then by \cite[Lemma 3.3.8]{Roger1991}, we obtain
$$
\begin{aligned}
	& \sum_{i=1}^{n_3} \sum_{j=1}^m
	\rho\left(\sigma_j\left(\theta\widehat{\mathcal X}_\mathbf U^{\langle i\rangle}+(1-\theta)\widehat{\mathcal Y}_\mathbf U^{\langle i\rangle}\right)\right)
	\leq
	\sum_{i=1}^{n_3} \sum_{j=1}^m
	\rho\left(\sigma_j\left(\theta\widehat{\mathcal X}_\mathbf U^{\langle i\rangle}\right)+
	\sigma_j\left((1-\theta)\widehat{\mathcal Y}_\mathbf U^{\langle i\rangle}\right)\right),
\end{aligned}
$$
which yields
\begin{equation}\label{TbCDIne}
\begin{aligned}
	& \sum_{i=1}^{n_3} \sum_{j=1}^m
	g_\lambda\left(\sigma_j\left(\theta\widehat{\mathcal X}_\mathbf U^{\langle i\rangle}+(1-\theta)\widehat{\mathcal Y}_\mathbf U^{\langle i\rangle}\right)\right)
	+\frac{\mu}{2}
	\left(\sigma_j\left(\theta\widehat{\mathcal X}_\mathbf U^{\langle i\rangle}+(1-\theta)\widehat{\mathcal Y}_\mathbf U^{\langle i\rangle}\right)\right)^2\\
	& \leq
	\sum_{i=1}^{n_3} \sum_{j=1}^m
	g_\lambda\left(\sigma_j\left(\theta\widehat{\mathcal X}_\mathbf U^{\langle i\rangle}\right)+
	\sigma_j\left((1-\theta)\widehat{\mathcal Y}_\mathbf U^{\langle i\rangle}\right)\right)
	+\frac{\mu}{2}
	\left(\sigma_j\left(\theta\widehat{\mathcal X}_\mathbf U^{\langle i\rangle}\right)+
	\sigma_j\left((1-\theta)\widehat{\mathcal Y}_\mathbf U^{\langle i\rangle}\right)\right)^2\\
	&=
	\sum_{i=1}^{n_3} \sum_{j=1}^m
	g_\lambda\left(\theta\sigma_j(\widehat{\mathcal X}_\mathbf U^{\langle i\rangle})+
	(1-\theta)\sigma_j(\widehat{\mathcal Y}_\mathbf U^{\langle i\rangle})\right)
	+\frac{\mu}{2}
	\left(\theta\sigma_j(\widehat{\mathcal X}_\mathbf U^{\langle i\rangle})+
	(1-\theta)\sigma_j(\widehat{\mathcal Y}_\mathbf U^{\langle i\rangle})\right)^2.
\end{aligned}
\end{equation}
Plugging (\ref{TbCDIne}) into (\ref{test64}) yields
\begin{equation}\label{test63}
	\begin{aligned}
		&\sum_{i=1}^{n_3} \sum_{j=1}^m
		g_\lambda\left(\sigma_j\left(\theta\widehat{\mathcal X}_\mathbf U^{\langle i\rangle}+(1-\theta)\widehat{\mathcal Y}_\mathbf U^{\langle i\rangle}\right)\right)
		+\frac{\mu}{2}
		\left(\sigma_j\left(\theta\widehat{\mathcal X}_\mathbf U^{\langle i\rangle}+(1-\theta)\widehat{\mathcal Y}_\mathbf U^{\langle i\rangle}\right)\right)^2  \\
		&\leq
		\sum_{i=1}^{n_3}\sum_{j=1}^{m}
		\theta g_\lambda\left(\sigma_j(\widehat{\mathcal X}_\mathbf U^{\langle i\rangle})\right)
		+
		\frac{\theta\mu}{2}\left(\sigma_j(\widehat{\mathcal X}_\mathbf U^{\langle i\rangle})\right)^2
		+
		(1-\theta)g_\lambda\left(\sigma_j(\widehat{\mathcal Y}_\mathbf U^{\langle i\rangle})\right)
		+
		\frac{(1-\theta)\mu}{2}\left(\sigma_j(\widehat{\mathcal Y}_\mathbf U^{\langle i\rangle})\right)^2.
	\end{aligned}
\end{equation}
By the definition of $G_\lambda$ and the fact that $\|\mathcal{X}\|_F^2=\sum_{i=1}^{n_3}\sum_{j=1}^m(\sigma_j(\widehat{\mathcal X}_\mathbf U^{\langle i\rangle}))^2$,   (\ref{test63}) implies that
$$
\begin{aligned}
	&~G_\lambda(\theta\mathcal X+(1-\theta)\mathcal Y)
+
	\frac{\mu}{2}\|\theta\mathcal X+(1-\theta)\mathcal Y\|_F^2\\
\leq&~
	\theta G_\lambda(\mathcal X)+
	\frac{\theta\mu}{2}\|\mathcal X\|_F^2+
	(1-\theta)G_\lambda(\mathcal Y)+
	\frac{(1-\theta)\mu}{2}\|\mathcal Y\|_F^2,
\end{aligned}
$$
which is equivalent to
\begin{equation*}\label{ConGm}
	\gamma(\theta\mathcal X+(1-\theta)\mathcal Y)
	\leq
	\theta\gamma(\mathcal X)+(1-\theta)\gamma(\mathcal Y), \quad \forall\theta\in[0,1].
\end{equation*}
Therefore,  $\gamma(\mathcal X)$ is convex.
\qed

\begin{lemma}\label{lem1}
For any tensors $\mathcal X_1,\mathcal X_2\in \mathbb {R}^{n_1\times n_2\times n_3}$, the following inequality holds:
\begin{equation*}\label{test6}
\begin{aligned}
\langle\tilde{\mathcal Z},
\mathcal X_1-\mathcal X_2\rangle
\leq
\beta G_\lambda(\mathcal X_1)-\beta G_\lambda(\mathcal X_2)
+\frac{\beta\mu}{2}\|\mathcal X_1-\mathcal X_2\|_F^2, \ \
\forall\tilde{\mathcal Z}\in\partial (\beta G_\lambda(\mathcal X_2)),\\
\end{aligned}
\end{equation*}
where $\beta$ is defined in (\ref{test1}) and $\mu$ is defined in Assumption \ref{assum1}(iv).
\end{lemma}
\textbf{Proof.}
As a consequence of Lemma \ref{lem5}, the function $\beta G_\lambda(\mathcal X)
+\frac{\beta\mu}{2}\|\mathcal X\|_F^2$ is convex with $\beta>0$,
which  yields
$$
\beta G_\lambda(\mathcal X_1)
+\frac{\beta\mu}{2}\|\mathcal X_1\|_F^2
\geq
\beta G_\lambda(\mathcal X_2)
+\frac{\beta \mu}{2}\|\mathcal X_2\|_F^2
+\langle\tilde{\mathcal Z}+\beta\mu\mathcal X_2,
\mathcal X_1-\mathcal X_2\rangle,
$$
for any $\tilde{\mathcal Z}\in \partial (\beta G_\lambda(\mathcal X_2))$.
The above inequality can be rewritten as
\begin{equation*}\label{test39}
\begin{aligned}
\langle\tilde{\mathcal Z},
\mathcal X_1-\mathcal X_2\rangle
&\leq
\beta G_\lambda(\mathcal X_1)-\beta G_\lambda(\mathcal X_2)
+\frac{\beta\mu}{2}\|\mathcal X_1\|_F^2-\frac{\beta\mu}{2}\|\mathcal X_2\|_F^2
-\beta\mu\langle\mathcal X_2,
\mathcal X_1\rangle
+\beta\mu\|\mathcal X_2\|_F^2\\
&=\beta G_\lambda(\mathcal X_1)-\beta G_\lambda(\mathcal X_2)
+\frac{\beta \mu}{2}\|\mathcal X_1-\mathcal X_2\|_F^2.
\end{aligned}
\end{equation*}
The proof is completed. \qed

For any matrix $\mathbf{A}\in \mathbb {R}^{n_1\times n_2}$, we define $\Phi_r(\mathbf{A})$ to be the best rank $r$ approximation of $\mathbf{A}$. Denote
\begin{equation}\label{Phir}
	\Psi_r(\mathbf{A})=\mathbf{A}-\Phi_r(\mathbf{A})
\end{equation}
and
\begin{equation}\label{phiA}
	\phi(\mathbf{A})=\sum_{j=1}^{m}g_\lambda(\sigma_j(\mathbf{A})),
\end{equation}
where $g_\lambda(\cdot)$ satisfies Assumption \ref{assum1}.

\begin{lemma}\label{lem11}
	For any $\mathbf{X}, \mathbf{Y}\in \mathbb{R}^{n_1\times n_2}$ with $\textup{rank}(\mathbf{X})=r$,
	one has
	\begin{equation}\label{test80}
		\phi(\Psi_{r}(\mathbf{Y})) \geq
		\phi(\Psi_{r}(\mathbf{X-Y})),
	\end{equation}
	where $\Psi_{r}(\cdot)$ and $\phi(\cdot)$  are defined as in (\ref{Phir}) and (\ref{phiA}).
\end{lemma}
\textbf{Proof.}
For any $\mathbf{X}, \mathbf{Y}\in \mathbb{R}^{n_1\times n_2}$,
it follows from  \cite[Theorem 3.3.16]{Roger1991}  that
$$
\sigma_j(\mathbf{Y})
\leq
\sigma_1(\mathbf{X})+\sigma_j(\mathbf{Y-X}), \quad j=r+1,\ldots,m,
$$
where $m=\min\{n_1,n_2\}$.
Summing up the above inequality over $j=r+1,\ldots,m$,
 we obtain
$$
\sum_{j=r+1}^m \sigma_j(\mathbf{Y})
\leq
\sum_{j=r+1}^m (\sigma_j(\mathbf{Y-X})+\sigma_1(\mathbf{X})).
$$
Note that  $-g_\lambda$ is convex and non-increasing on $[0,+\infty)$ by 
Assumption \ref{assum1}(i),
which together with \cite[Lemma 3.3.8]{Roger1991} leads to
\begin{equation}\label{test81}
	\begin{aligned}
		\sum_{j=r+1}^m -g_\lambda(\sigma_j(\mathbf{Y}))
		&\leq
		\sum_{j=r+1}^m -g_\lambda(\sigma_j(\mathbf{Y-X})+\sigma_1(\mathbf{X})) \\
		&\leq
		\sum_{j=r+1}^m -g_\lambda(\sigma_j(\mathbf{Y-X}))=\sum_{j=r+1}^m -g_\lambda(\sigma_j(\mathbf{X-Y})),
	\end{aligned}
\end{equation}
where the second inequality holds since
$\sigma_j(\mathbf{Y-X})+\sigma_1(\mathbf{X})\geq\sigma_j(\mathbf{Y-X})\geq0$ for any
$j=r+1,\ldots,m$
and the fact
$-g_\lambda$ is non-increasing.
By definitions of $\Psi_{r}(\cdot)$ and $\phi(\cdot)$   in (\ref{Phir}) and (\ref{phiA}), we know that
(\ref{test81}) is equivalent to
$$
\phi(\Psi_{r}(\mathbf{Y})) \geq
\phi(\Psi_{r}(\mathbf{X-Y})).
$$
This completes the proof.
\qed

\begin{lemma}\label{lem2}
For any $\mathbf{X}, \mathbf{Y}\in \mathbb{R}^{n_1\times n_2}$ with $\textup{rank}(\mathbf{X})=r$,
the following inequality holds
\begin{equation*}\label{test8}
\begin{aligned}
\phi(\mathbf{X})
-\phi(\mathbf{Y})
\leq
\phi(\Phi_{r}(\mathbf{E}))
-\phi(\Psi_{r}(\mathbf{E})),
\end{aligned}
\end{equation*}
where $\mathbf{E}:=\mathbf{X}-\mathbf{Y}$, $\Phi_{r}(\cdot)$, $\Psi_{r}(\cdot)$, and $\phi(\cdot)$  are defined as in (\ref{Phir}) and (\ref{phiA}).
\end{lemma}
\textbf{Proof.}
By the definitions of $\phi, \Phi_{r}$ and $\Psi_{r}$  in (\ref{Phir}) and (\ref{phiA}), we can obtain that
\begin{equation}\label{9}
\begin{aligned}
\phi(\mathbf{X})
-\phi(\mathbf{Y})
&=\phi(\Phi_{r}(\mathbf{X}))
+\phi(\Psi_{r}(\mathbf{X}))
-\phi(\Phi_{r}(\mathbf{Y}))
-\phi(\Psi_{r}(\mathbf{Y}))\\
&=\phi(\Phi_{r}(\mathbf{X}))
-\phi(\Phi_{r}(\mathbf{Y}))
-\phi(\Psi_{r}(\mathbf{Y}))\\
&=\sum_{j=1}^rg_\lambda(\sigma_j(\mathbf{X})) -\sum_{j=1}^rg_\lambda(\sigma_j(\mathbf{Y}))-\phi(\Psi_{r}(\mathbf{Y}))\\
&=\sum_{j=1}^rg_\lambda(\sigma_j(\mathbf{X}-\mathbf{Y}+\mathbf{Y})) -\sum_{j=1}^rg_\lambda(\sigma_j(\mathbf{Y}))-\phi(\Psi_{r}(\mathbf{Y}))\\
&\leq\sum_{j=1}^rg_\lambda(\sigma_j(\mathbf{X}-\mathbf{Y})) -\phi(\Psi_{r}(\mathbf{Y}))\\
&=\phi(\Phi_{r}(\mathbf{E}))
-\phi(\Psi_{r}(\mathbf{Y}))\\
&\leq\phi(\Phi_{r}(\mathbf{E}))
-\phi(\Psi_{r}(\mathbf{E})),
\end{aligned}
\end{equation}
where the first inequality follows from \cite[Theorem 2.6]{MIAO2013} and the second inequality follows from Lemma \ref{lem11}.
\qed

\section*{Appendix D. Subdifferential of 	$G_\lambda\left(\mathcal X\right)$ in (\ref{Glmbade})}

\begin{lemma}\label{lem7}
	For any matrix $ {\bf X}\in \mathbb {R}^{n_1\times n_2}$ and any tensor $\mathcal X\in \mathbb {R}^{n_1\times n_2\times n_3}$,
	define
	$(g_\lambda\circ\sigma)({\bf X}):=
	\sum_{j=1}^{\min\{n_1,n_2\}} g_\lambda(\sigma_j({\bf X}))$
	and
	$H_\lambda(\widehat{\mathcal X}_\mathbf U):=
	\sum_{i=1}^{n_3}\sum_{j=1}^{\min\{n_1,n_2\}} g_\lambda(\sigma_j({\widehat{\mathcal X}}_\mathbf U^{\langle i\rangle}))$
	.
	Then
	$$
	\overline{\mathcal Z}\in \partial (g_\lambda\circ\sigma)(\overline{\mathcal X})
	\iff
	\widehat{\mathcal Z}_\mathbf U\in \partial H_\lambda(\widehat{\mathcal X}_\mathbf U).
	$$
\end{lemma}
\textbf{Proof.}
By the definition of the subdifferential of a function, we know that for any
$\overline{\mathcal Z}\in \partial (g_\lambda\circ\sigma)(\overline{\mathcal X})$,
there exist a sequence
$\{\overline{\mathcal X}^{k}\}$ such that  $\overline{\mathcal X}^{k}$ converges to $\overline{\mathcal X}$
with
$(g_\lambda\circ\sigma)(\overline{\mathcal X}^{k})\rightarrow(g_\lambda\circ\sigma)(\overline{\mathcal X})$
and
a sequence of regular subdifferential $\overline{\mathcal Z}^k\in\widehat\partial (g_\lambda\circ\sigma)(\overline{\mathcal X}^k)$ such that  $\overline{\mathcal Z}^k\rightarrow\overline{\mathcal Z}$.
Notice that
$(g_\lambda\circ\sigma)(\overline{\mathcal X}^{k})=H_\lambda(\widehat{\mathcal X}_\mathbf U^{k})$.
For any  $\widehat{\mathcal Y}_\mathbf U^{k}\in \mathbb {C}^{n_1\times n_2\times n_3}$,
where
$\|\widehat{\mathcal Y}_\mathbf U^{k}\|_F\rightarrow 0$ as $k\rightarrow +\infty$,
the following inequality holds

$$
\begin{aligned}
	H_\lambda(\widehat{\mathcal X}_\mathbf U^{k}+\widehat{\mathcal Y}_\mathbf U^{k})
	&=
	(g_\lambda\circ\sigma)(\overline{\mathcal X}^{k}+\overline{\mathcal Y}^{k})\\
	&\geq
	(g_\lambda\circ\sigma)(\overline{\mathcal X}^{k})+\langle \overline{\mathcal Z}^{k},\overline{\mathcal Y}^{k}\rangle
	+o(\|\overline{\mathcal Y}^{k}\|_F) \\
	&=H_\lambda(\widehat{\mathcal X}_\mathbf U^{k})+ \langle\widehat{\mathcal Z}_\mathbf U^{k}, \widehat{\mathcal Y}_\mathbf U^{k}\rangle + o(\|\widehat{\mathcal Y}_\mathbf U^{k}\|_F), \\
\end{aligned}
$$
where
the first inequality follows from the definition of the regular subdifferential of $g_\lambda\circ\sigma$ at  $\overline{\mathcal X}^{k}$
and
the second equation holds since
$\langle \overline{\mathcal Z}^{k},\overline{\mathcal Y}^{k}\rangle=\langle\widehat{\mathcal Z}_\mathbf U^{k}, \widehat{\mathcal Y}_\mathbf U^{k}\rangle$ and $\|\overline{\mathcal Y}^{k}\|_F=\|\widehat{\mathcal Y}_\mathbf U^{k}\|_F$  \cite[Definition 5]{Song_2020}. Here $o(\cdot)$ denotes a higher-order infinitesimal.
As a consequence, we can deduce that
$\widehat{\mathcal Z}_\mathbf U^{k}\in \widehat\partial H_\lambda(\widehat{\mathcal X}_\mathbf U^{k})$
for each $k$.

Since $\overline{\mathcal Z}^k\rightarrow\overline{\mathcal Z}$, we get that
$\widehat{\mathcal Z}_\mathbf U^{k}\rightarrow\widehat{\mathcal Z}_\mathbf U$.
It can be easily shown that
the sequence $\{\widehat{\mathcal X}_\mathbf U^k\}$ converges to $\widehat{\mathcal X}_\mathbf U$  and $\{H_\lambda(\widehat{\mathcal X}_\mathbf U^k)\}$
converges to
$H_\lambda(\widehat{\mathcal X}_\mathbf U)$.
Since $g_\lambda(\cdot)$ is continuous, we know that  $\widehat\partial H_\lambda(\widehat{\mathcal X}_\mathbf U^k)$ and  $\partial H_\lambda(\widehat{\mathcal X}_\mathbf U^k)$ are closed
with $\widehat\partial H_\lambda(\widehat{\mathcal X}_\mathbf U^k)\subset\partial H_\lambda(\widehat{\mathcal X}_\mathbf U^k)$ \cite[Theorem 8.6]{R.TyrrellRockafellar1998}, which yields
$\widehat{\mathcal Z}_\mathbf U\in \widehat\partial H_\lambda(\widehat{\mathcal X}_\mathbf U)\subset\partial H_\lambda(\widehat{\mathcal X}_\mathbf U)$.

By using similar arguments to the proof of the opposite inclusion, we can easily obtain that
for any $\widehat{\mathcal Z}_\mathbf U\in \partial H_\lambda(\widehat{\mathcal X}_\mathbf U)$,
$\overline{\mathcal Z}\in \partial (g_\lambda\circ\sigma)(\overline{\mathcal X})$ holds.
\qed

\begin{lemma}\label{lem6}
	For any tensor $ \mathcal X\in \mathbb {R}^{n_1\times n_2\times n_3}$, one has
	$$
	\partial H_\lambda(\widehat{\mathcal X}_{\mathbf U})=
	\textup{fold}_{3}(\partial (g_\lambda\circ\sigma)(\overline{\mathcal X})),
	$$
	where  $H_\lambda$ and $g_\lambda\circ\sigma$ are defined as the same  in Lemma \ref{lem7}, and $\operatorname{fold}_{3}(\cdot)$ operating on a set means to operate each element of the set.
\end{lemma}
\textbf{Proof.}
For any $\widehat{\mathcal Y}_{\mathbf U}\in\partial H_\lambda(\widehat{\mathcal X}_{\mathbf U})$,
it follows from Lemma \ref{lem7} that $
\text{bdiag}(\widehat{\mathcal Y}_{\mathbf U})=\overline{\mathcal Y}
\in\partial(g_\lambda\circ\sigma)(\overline{\mathcal X})
$.
By the definition of $\textup{fold}_{3}(\cdot)$ in (\ref{Fold3}), we get that
$
\widehat{\mathcal Y}_{\mathbf U}=
\textup{fold}_{3}(\overline{\mathcal Y})
\in\textup{fold}_{3}(\partial(g_\lambda\circ\sigma)(\overline{\mathcal X}))
$, which implies that
$
\partial H_\lambda(\widehat{\mathcal X}_{\mathbf U})
\subseteq
\textup{fold}_{3}(\partial (g_\lambda\circ\sigma)(\overline{\mathcal X}))
$.

A similar argument leads to the opposite direction, i.e., $\textup{fold}_{3}(\partial (g_\lambda\circ\sigma)(\overline{\mathcal X}))
\subseteq
\partial H_\lambda(\widehat{\mathcal X}_{\mathbf U})
$.
This completes the proof.
\qed
\begin{lemma}\label{lem3}
	For any  $\mathcal X\in \mathbb {R}^{n_1\times n_2\times n_3}$,
	the transformed tensor SVD of $\mathcal X$ is denoted by  $\mathcal X= \mathcal U \diamond_{\mathbf U}\Sigma\diamond_{\mathbf U}\mathcal V^T$.
	Let $G_\lambda(\mathcal{X})$ be defined in (\ref{test1}).
	Then the  subdifferential of
	$G_\lambda\left(\mathcal X\right)$ at
	$\mathcal X$ is given by
	$$
	\partial G_\lambda(\mathcal X)
	=\left\{\mathcal U \diamond_{\mathbf U}\mathcal D\diamond_{\mathbf U}\mathcal V^T\mid
	\widehat{\mathcal D}_\mathbf U^{\langle i\rangle}\in \mathfrak{B}_i,  i=1,2,\ldots,n_3
	\right\},
	$$
	where
$\mathfrak{B}_i$ is defined as
\begin{equation}\label{Bid}
	\mathfrak{B}_i
	=\left\{\operatorname{Diag}(\mathbf{d}^{i})
	\mid \mathbf{d}^{i}=({d}_1^{i}, {d}^{i}_2,\ldots, {d}^{i}_m)^T \in \mathbb{R}^m,
	{d}^{i}_j\in \partial g_\lambda(\sigma_j(\widehat{\mathcal X}_\mathbf U^{\langle i\rangle})),
	j=1,2,\ldots,m
	\right\}
\end{equation}
and $m:=\min\{n_1,n_2\}$.
\end{lemma}
\textbf{Proof.}
Note that  $(g_\lambda\circ\sigma)({\bf X}):=\sum_{j=1}^{m} g_\lambda(\sigma_j({\bf X}))$ for any  matrix $ {\bf X}\in \mathbb {R}^{n_1\times n_2}$.
For any  $\mathcal X\in \mathbb {R}^{n_1\times n_2\times n_3}$, by \cite[Theorem 7.1]{Lewis2005a}, we get that
\begin{equation}\label{test65}
	\partial (g_\lambda\circ\sigma)(\overline{\mathcal X})
	=
	\left\{\text{bdiag}({\widehat{\mathcal U}}_{\mathbf U})\cdot
	\text{bdiag}({\widehat{\mathcal D}}_{\mathbf U})\cdot
	\text{bdiag}({\widehat{\mathcal V}}_{\mathbf U})\mid
	\widehat{\mathcal D}_\mathbf U^{\langle i\rangle}\in \mathfrak{B}_i,  i=1,2,\ldots,n_3
	\right\},
\end{equation}
where $\mathfrak{B}_i$ is defined in (\ref{Bid}).
%
It follows from Lemma \ref{lem6} that
\begin{equation}\label{test62}
	\partial (g_\lambda\circ\sigma)(\overline{\mathcal X})=
	\text{bdiag}(\partial H_\lambda(\widehat{\mathcal X}_{\mathbf U})),
\end{equation}
where $H_\lambda$ is defined as the same in Lemma \ref{lem7} and $\text{bdiag}(\cdot)$ operates a set means that $\text{bdiag}(\cdot)$ operates each element of  the corresponding set.
In light of the definitions of  $G_\lambda$ and $H_\lambda$, it is obvious that
\begin{equation}\label{testG8}
	G_\lambda(\mathcal X)=H_\lambda(\widehat{\mathcal X}_{\mathbf U})
	=H_\lambda(\mathbf U[\mathcal X]).
\end{equation}
Under Assumptions \ref{assum1}, one can easily get that $H_\lambda(\cdot)$ is a proper and lower semicontinuous function.
On the other hand, the tensor $\mathbf U[\mathcal X]$ is  defined as multiplying by a unitary matrix $\mathbf U$ onto each tube of  $\mathcal X$,
which is an invertible linear transformation.
Applying \cite[Exercise 10.7]{R.TyrrellRockafellar1998} to (\ref{testG8}) yields
\begin{equation}\label{test61}
	\partial G_\lambda(\mathcal X)
	=\mathbf U^{T}[\partial H_\lambda(\mathbf U[\mathcal X])]
	=\mathbf U^{T}[\partial H_\lambda(\widehat{\mathcal X}_{\mathbf U})].
\end{equation}
Here for abuse of notation, we use $\mathbf U[E]$ to denote a set with  each element of $E$ being operated by $\mathbf U$.
Taking (\ref{test61})  with (\ref{test62}), we obtain that
$$
\text{bdiag}(\mathbf U[\partial G_\lambda(\mathcal X)])
=
\text{bdiag}(\partial H_\lambda(\widehat{\mathcal X}_{\mathbf U}))
=\partial (g_\lambda\circ\sigma)(\overline{\mathcal X}),
$$
which together with (\ref{test65}) and  the definition of $\mathbf U$-product further yields
$$
\begin{aligned}
	\partial G_\lambda(\mathcal X)
	&=\mathbf U^T[\textup{fold}_{3}(\partial (g_\lambda\circ\sigma)(\overline{\mathcal X}))]\\
	&=\left\{\mathcal U \diamond_{\mathbf U}\mathcal D\diamond_{\mathbf U}\mathcal V^T\mid
	\widehat{\mathcal D}_\mathbf U^{\langle i\rangle}\in \mathfrak{B}_i, i=1,2,\ldots,n_3
	\right\},
\end{aligned}
$$
where $\mathfrak{B}_i$ is defined in (\ref{Bid}).
The proof is completed.
\qed

\section*{Appendix E. Proof of Theorem \ref{theo1}}

\textbf{Step 1}. Let $\Delta=\tilde{\mathcal X}-\mathcal X^*$.
Suppose that $\|\Delta\|_F > 1$,
it can be easily seen from  (\ref{test0}) that
\begin{equation}\label{test40}
	\begin{aligned}
		\alpha_2\|\Delta\|_F-\tau_2 \sqrt{{\log (d)}/{n}}\|\Delta\|_{\text {TTNN}}
		&\leq
		\langle\nabla f_{n,\mathcal{Y}}\left(\mathcal X^*+\Delta\right)-\nabla f_{n,\mathcal{Y}}(\mathcal X^*), \Delta\rangle \\
		&=\langle\nabla f_{n,\mathcal{Y}}(\tilde{\mathcal X})-\nabla f_{n,\mathcal{Y}}\left(\mathcal X^*\right), \tilde{\mathcal X}-\mathcal X^*\rangle \\
		&=\langle\nabla f_{n,\mathcal{Y}}(\tilde{\mathcal X})+\mathcal Z-\mathcal Z-\nabla f_{n,\mathcal{Y}}(\mathcal X^*),\tilde{\mathcal X}-\mathcal X^*\rangle \\
		&=\langle\nabla f_{n,\mathcal{Y}}(\tilde{\mathcal X})+\mathcal Z, \tilde{\mathcal X}-\mathcal X^*\rangle+\langle-\mathcal Z-\nabla f_{n,\mathcal{Y}}(\mathcal X^*), \tilde{\mathcal X}-\mathcal X^*\rangle,
	\end{aligned}
\end{equation}
where $\mathcal Z\in\partial (\beta G_\lambda(\tilde{\mathcal X}))$ will be given in detail later.
Since $\tilde{\mathcal X}$ is a stationary point of (\ref{test1}),
we get that
$$
0\in\nabla f_{n,\mathcal{Y}}(\tilde{\mathcal X})+\partial (\beta G_\lambda(\tilde{\mathcal X}))+\partial \delta_{D}(\tilde{\mathcal X}),
$$
where $\delta_D({\mathcal X})$ is the indicator function of the set $D:=\{\mathcal X: \|\mathcal X\|_\infty\leq c\}$.
Then there exists a tensor  $\mathcal Z\in\partial (\beta G_\lambda(\tilde{\mathcal X}))$ such that $-\nabla f_{n,\mathcal{Y}}(\tilde{\mathcal X})-\mathcal Z\in \partial \delta_D(\tilde{\mathcal X})=N_D(\tilde{\mathcal X})$, where $N_D(\tilde{\mathcal X})$ denotes the normal cone of $D$ at $\tilde{\mathcal X}$ and the equality holds by \cite[Example 3.5]{Amir2017}.
Since $\mathcal X^*\in D$ is feasible,
by the definition of normal cone (e.g., see \cite{Amir2017}), we have
\begin{equation}\label{Inseq9}
	\langle\nabla f_{n,\mathcal{Y}}(\tilde{\mathcal X})+\mathcal Z, \tilde{\mathcal X}-\mathcal X^*\rangle \leq 0.
\end{equation}
Taking (\ref{Inseq9}) together with (\ref{test40}) yields
\begin{equation}\label{test2}
	\begin{aligned}
		\alpha_2\|\Delta\|_F-\tau_2 \sqrt{\frac{\log d}{n}}\|\Delta\|_{\text {TTNN}}
		& \leq
		\langle-\mathcal Z-\nabla f_{n,\mathcal{Y}}(\mathcal X^*), \tilde{\mathcal X}-\mathcal X^*\rangle\\
		&=\underbrace{\langle-\mathcal Z, \tilde{\mathcal X}-\mathcal X^*\rangle}_{I_1}
		+\underbrace{\langle-\nabla f_{n,\mathcal{Y}}(\mathcal X^*), \tilde{\mathcal X}-\mathcal X^*\rangle}_{I_2}.
	\end{aligned}
\end{equation}

\textbf{Upper bound of  $I_1$.}
By virtue of the definition of the inner product of two tensors \cite[Definition 5]{Song_2020},
we immediately obtain
\begin{equation}\label{test66}
	\begin{aligned}
		\langle-\mathcal Z, \tilde{\mathcal X}-\mathcal X^*\rangle
		&=\langle \overline{-\mathcal Z},
		\overline{\Delta}
		\rangle\leq\left\|\overline{\mathcal Z}\right\|
		\left\|\overline{\Delta}\right\|_*=\left\|\mathcal Z\right\|_{\mathbf U}
		\left\|\Delta\right\|_{\text{TTNN}},
	\end{aligned}
\end{equation}
where the first inequality holds by the H{\"o}lder's inequality.
In addition, by Lemma \ref{lem3} and Assumption \ref{assum1}(iii), we obtain
$\left\|\mathcal Z\right\|_{\mathbf U}\leq\beta\lambda k_0$,
which together with (\ref{test66}) yields
\begin{equation}\label{test4}
	\begin{aligned}
		\langle-\mathcal Z, \tilde{\mathcal X}-\mathcal X^*\rangle
		\leq\beta\lambda k_0\left\|\Delta\right\|_{\text{TTNN}}.
	\end{aligned}
\end{equation}

\textbf{Upper bound of  $I_2$.}
Similar to the analysis of $I_1$, we have
\begin{equation}\label{test3}
	\begin{aligned}
		\langle-\nabla f_{n,\mathcal{Y}}(\mathcal X^*), \tilde{\mathcal X}-\mathcal X^*\rangle
		&=\langle \overline{-\nabla f_{n,\mathcal{Y}}(\mathcal X^*)},
		\overline{\Delta}
		\rangle\\
		&\leq\left\|\overline{\nabla f_{n,\mathcal{Y}}\left(\mathcal X^*\right)}\right\|
		\left\|\overline{\Delta}\right\|_*\\
		&=\left\|\nabla f_{n,\mathcal{Y}}\left(\mathcal X^*\right)\right\|_{\mathbf U}
		\left\|\Delta\right\|_{\text{TTNN}}\\
		&\leq\frac{\beta\lambda k_0}{4}\left\|\Delta\right\|_{\text{TTNN}},
	\end{aligned}
\end{equation}
where
the last inequality follows from (\ref{test20}).

Substituting (\ref{test3}) and (\ref{test4}) into (\ref{test2}) yields
\begin{equation}\label{test33}
	\begin{aligned}
		\alpha_2\|\Delta\|_F-\tau_2 \sqrt{\frac{\log d}{n}}\|\Delta\|_{\text{TTNN}}
		& \leq
		\left(\beta\lambda k_0+\frac{\beta\lambda k_0}{4}\right)\left\|\Delta\right\|_{\text{TTNN}}=\frac{5\beta\lambda k_0}{4}\left\|\Delta\right\|_{\text{TTNN}}.
	\end{aligned}
\end{equation}
Since
$\|\mathcal X\|_{\text{TTNN}}\leq t$,
we can deduce that
\begin{equation}\label{test34}
	\left\|\Delta\right\|_{\text{TTNN}}
	=
	\|\tilde{\mathcal X}-\mathcal X^*\|_{\text{TTNN}}
	\leq \|\tilde{\mathcal X}\|_{\text{TTNN}}+\|\mathcal X^*\|_{\text{TTNN}}
	\leq 2t,
\end{equation}
where the first inequality follows from \cite[Remark 2]{Qiu_2021}.
This taken together with (\ref{test33}) indicates that
\begin{equation}\label{test5}
	\begin{aligned}
		\alpha_2\|\Delta\|_F-\tau_2 \sqrt{\frac{\log d}{n}}\|\Delta\|_{\text{TTNN}}
		\leq \frac{5\beta\lambda k_0 t}{2}.
	\end{aligned}
\end{equation}

Note that
\begin{equation}\label{test24}
	\begin{aligned}
		\tau_2 \sqrt{\frac{\log d}{n}}\|\Delta\|_{\text{TTNN}}
		&\leq
		\tau_2 \sqrt{\log d \frac{\alpha_2^2}{16 t^2 \tau_2^2\log d} }\|\Delta\|_{\text{TTNN}}\\
		&=\frac{\alpha_2}{4t}\|\Delta\|_{\text{TTNN}}\\
		&\leq \frac{\alpha_2}{4t}\cdot 2t
		=\frac{\alpha_2}{2}.
	\end{aligned}
\end{equation}
where the first inequality holds by (\ref{test27}) and the second inequality holds by (\ref{test34}).
By
taking (\ref{test24})  with (\ref{test5}), we obtain that
\begin{equation}\label{test25}
	\begin{aligned}
		\alpha_2\|\Delta\|_F
		&\leq
		\frac{\alpha_2}{2}
		+\frac{5\beta\lambda k_0 t}{2}\\
		&\leq
		\frac{\alpha_2}{2}+\frac{5\alpha_2}{12}=\frac{11\alpha_2}{12},
	\end{aligned}
\end{equation}
where the second inequality holds by the assumption $\lambda\leq\frac{\alpha_2}{6  t\beta k_0}$ in (\ref{test20}).
Then we can derive that
$
\|\Delta\|_F
\leq \frac{11}{12}< 1,
$
which leads to a contradiction with
the
assumption $\|\Delta\|_F \textgreater 1$.

\textbf{Step 2}. We know that  $\|\Delta\|_F \leq 1$ from Step 1. Invoking the RSC condition in (\ref{test0}) gives
\begin{equation}\label{test38}
	\alpha_1\|\Delta\|_F^2-\tau_1 \frac{\log d}{n}\|\Delta\|_{\text{TTNN}}^2
	\leq
	\langle\nabla f_{n,\mathcal{Y}}(\mathcal X^*+\Delta)-\nabla f_{n,\mathcal{Y}}(\mathcal X^*), \Delta\rangle,
\end{equation}
which together with  (\ref{test40}) and (\ref{Inseq9}) yields that
\begin{equation}\label{test16}
	\begin{aligned}
		\alpha_1\|\Delta\|_F^2-\tau_1 \frac{\log d}{n}\|\Delta\|_{\text{TTNN}}^2
		&\leq
		\langle-\mathcal Z-\nabla f_{n,\mathcal{Y}}(\mathcal X^*), \tilde{\mathcal X}-\mathcal X^*\rangle\\
		&=\langle-\mathcal Z, \tilde{\mathcal X}-\mathcal X^*\rangle
		+\langle-\nabla f_{n,\mathcal{Y}}(\mathcal X^*), \tilde{\mathcal X}-\mathcal X^*\rangle.
	\end{aligned}
\end{equation}
It follows from Lemma \ref{lem1} that
\begin{equation}\label{test15}
	\begin{aligned}
		\langle-\mathcal Z, \tilde{\mathcal X}-\mathcal X^*\rangle=
		\langle\mathcal Z,\mathcal X^*-\tilde{\mathcal X}\rangle
		\leq
		\beta G_\lambda(\mathcal X^*)-\beta G_\lambda(\tilde{\mathcal X})
		+\frac{\beta\mu}{2}\|\mathcal X^*-\tilde{\mathcal X}\|_F^2,
	\end{aligned}
\end{equation}
where $G_\lambda$ is defined in (\ref{test1}).
This together with (\ref{test3}) and (\ref{test16}) yields
$$
\begin{aligned}
	\alpha_1\|\Delta\|_F^2-
	\tau_1 \frac{\log d}{n}\|\Delta\|_{\text{TTNN}}^2
	&\leq
	\beta G_\lambda(\mathcal X^*)-\beta G_\lambda(\tilde{\mathcal X})
	+\frac{\beta\mu}{2}\|\mathcal X^*-\tilde{\mathcal X}\|_F^2
	+\frac{\beta\lambda k_0}{4}\|\Delta\|_{\text{TTNN}}\\
	&=
	\beta G_\lambda(\mathcal X^*)-\beta G_\lambda(\tilde{\mathcal X})
	+\frac{\beta\mu}{2}\|\Delta\|_F^2
	+\frac{\beta\lambda k_0}{4}\|\Delta\|_{\text{TTNN}}.
\end{aligned}
$$
Rearranging the terms of the above inequality yields
\begin{equation}\label{test17}
	\begin{aligned}
		\left(\alpha_1-\frac{\beta\mu}{2}\right)\|\Delta\|_F^2
		&\leq
		\beta G_\lambda\left(\mathcal X^*\right)-\beta G_\lambda(\tilde{\mathcal X})
		+\frac{\beta \lambda k_0}{4}\left\|\Delta\right\|_{\text{TTNN}}
		+\tau_1 \frac{\log d}{n}\|\Delta\|_{\text{TTNN}}^2.
	\end{aligned}
\end{equation}
Notice that
\begin{equation}\label{test52}
	\begin{aligned}
		\tau_1\frac{\log d}{n}
		\|\Delta\|_{\text{TTNN}}^2
		&=\frac{\tau_1}{\alpha_2} \sqrt{\frac{\log d}{n}}
		\cdot\alpha_2 \sqrt{\frac{\log d}{n}}
		\cdot\|\Delta\|_{\text{TTNN}}
		\cdot\|\Delta\|_{\text{TTNN}}\\
		&\leq
		\frac{\tau_1}{\alpha_2} \sqrt{\frac{\log d}{n}}
		\cdot\alpha_2 \sqrt{\frac{\log d}{n}}
		\cdot2t
		\cdot\|\Delta\|_{\text{TTNN}},
	\end{aligned}
\end{equation}
where the inequality  follows from (\ref{test34}).
In view of (\ref{test20}) and (\ref{test27}), one has
\begin{equation}\label{34Condu}
	\frac{\tau_1}{\alpha_2} \sqrt{\frac{\log d}{n}}
	\leq
	\frac{\tau_1}{\alpha_2} \sqrt{\log d \frac{\alpha_2^2}{16 t^2 \tau_1^2\log d} }
	=
	\frac{1}{4t} ,\quad \
	\alpha_2 \sqrt{\frac{\log d}{n}}\leq\frac{\beta \lambda k_0}{4}.
\end{equation}
Plugging (\ref{34Condu}) into (\ref{test52}) then gives
$$
\begin{aligned}
	\tau_1\frac{\log d}{n}
	\|\Delta\|_{\text{TTNN}}^2
	&\leq\frac{1}{4t}
	\cdot\frac{\beta\lambda k_0}{4}
	\cdot2t
	\cdot\|\Delta\|_{\text{TTNN}} =
	\frac{\beta\lambda k_0}{8}\|\Delta\|_{\text{TTNN}},
\end{aligned}
$$
which taken together  with (\ref{test17}) gives
\begin{equation}\label{test13}
	\begin{aligned}
		\left(\alpha_1-\frac{\beta\mu}{2}\right)\|\Delta\|_F^2
		&\leq
		\beta G_\lambda(\mathcal X^*)-\beta G_\lambda(\tilde{\mathcal X})
		+\frac{\beta\lambda k_0}{4}\|\Delta\|_{\text{TTNN}}
		+\frac{\beta\lambda k_0}{8}\|\Delta\|_{\text{TTNN}}\\
		&\leq
		\beta G_\lambda(\mathcal X^*)-\beta G_\lambda(\tilde{\mathcal X})
		+\frac{\beta\lambda k_0}{2}\|\Delta\|_{\text{TTNN}}.
	\end{aligned}
\end{equation}

It follows from Lemma \ref{lem4} and Definition \ref{definition1} that
\begin{equation}\label{test51}
	\begin{aligned}
		\frac{\beta\lambda k_0}{2}\|\Delta\|_{\text{TTNN}}
		=\frac{\beta\lambda k_0}{2}\sum_{i=1}^{n_3}\left\|\widehat{\Delta}_\mathbf U^{\langle i\rangle}\right\|_*
		&\leq
		\frac{\beta\lambda k_0}{2}\sum_{i=1}^{n_3}\left(\frac{\phi(\widehat{\Delta}_\mathbf U^{\langle i\rangle})}{\lambda k_0}
		+\frac{\mu}{2 \lambda k_0}\left\|\widehat{\Delta}_\mathbf U^{\langle i\rangle}\right\|_F^2\right)\\
		&=\sum_{i=1}^{n_3}\left(\frac{\beta \phi(\widehat{\Delta}_\mathbf U^{\langle i\rangle})}{2}
		+\frac{\beta \mu}{4}\left\|\widehat{\Delta}_\mathbf U^{\langle i\rangle}\right\|_F^2\right),
	\end{aligned}
\end{equation}
where $\phi(\widehat{\Delta}_\mathbf U^{\langle i\rangle})=\sum_{j=1}^{m}g_\lambda(\sigma_j(\widehat{\Delta}_\mathbf U^{\langle i\rangle}))$.

For any matrix $\mathbf{A}\in \mathbb {R}^{n_1\times n_2}$, we define $\Phi_r(\mathbf{A})$ to be the best rank $r$ approximation of $\mathbf{A}$ and $\Psi_r(\mathbf{A})=\mathbf{A}-\Phi_r(\mathbf{A})$.
Thus, one can rewrite (\ref{test51}) equivalently as follows:
\begin{equation}\label{test55}
	\begin{aligned}
		\frac{\beta \lambda k_0}{2}\|\Delta\|_{\text{TTNN}}
		&\leq
		\sum_{i=1}^{n_3}\frac{\beta \phi(\widehat{\Delta}_\mathbf U^{\langle i\rangle})}{2}
		+\sum_{i=1}^{n_3}\frac{\beta \mu}{4}\left\|\widehat{\Delta}_\mathbf U^{\langle i\rangle}\right\|_F^2\\
		&=\sum_{i=1}^{n_3}\frac{1}{2}\left(\beta \phi\big(\Phi_{r_i}\big(\widehat{\Delta}_\mathbf U^{\langle i\rangle}\big)\big)
		+\beta \phi\big(\Psi_{r_i}\big(\widehat{\Delta}_\mathbf U^{\langle i\rangle}\big)\big)\right)+\frac{\beta \mu}{4}\|\widehat{\Delta}_\mathbf U\|_F^2\\
		&=\sum_{i=1}^{n_3}\frac{1}{2}\left(\beta \phi\big(\Phi_{r_i}\big(\widehat{\Delta}_\mathbf U^{\langle i\rangle}\big)\big)
		+\beta \phi\big(\Psi_{r_i}\big(\widehat{\Delta}_\mathbf U^{\langle i\rangle}\big)\big)\right)+\frac{\beta \mu}{4}\|\Delta\|_F^2,
	\end{aligned}
\end{equation}
where the last equation follows from the fact that $\|\widehat{\Delta}_\mathbf U\|_F=\|\Delta\|_F$.
Recalling the definition of $G_\lambda$ in (\ref{test1}), we know that $G_\lambda(\mathcal X)=\sum_{i=1}^{n_3}\phi(\widehat{\mathcal X}_\mathbf U^{\langle i\rangle})$ for any $\mathcal X\in \mathbb {R}^{n_1\times n_2\times n_3}$.
Suppose $r_i$ is the rank of matrix
$(\widehat{\mathcal X^*})_\mathbf U^{\langle i\rangle}, i=1,\ldots, n_3.$ 
Consequently, we obtain
\begin{equation}\label{Unsdg}
	\begin{aligned}
		\beta G_\lambda(\mathcal X^*)-\beta G_\lambda(\tilde{\mathcal X})
		&=\sum_{i=1}^{n_3}\beta \big(\phi\big((\widehat{\mathcal X^*})_\mathbf U^{\langle i\rangle}\big)
		-\phi\big(\widehat{\tilde{\mathcal X}}_{\mathbf U}^{\langle i\rangle}\big)\big)\\
		&\leq\sum_{i=1}^{n_3}\beta \big(\phi\big(\Phi_{r_i}\big(\widehat \Delta_\mathbf U^{\langle i\rangle}\big)\big)
		-\phi\big(\Psi_{r_ i}\big(\widehat \Delta_{\mathbf U}^{\langle i\rangle}\big)\big)\big),
	\end{aligned}
\end{equation}
where the inequality holds arises from Lemma \ref{lem2}.
Taking this together with (\ref{test13}) and (\ref{test55}), one has
\begin{equation}\label{test14}
	\begin{aligned}
		&\left(\alpha_1-\frac{\beta \mu}{2}\right)\|\Delta\|_F^2\\
		&\leq
		\sum_{i=1}^{n_3}\big(\beta\phi\big(\Phi_{r_i}\big(\widehat \Delta_\mathbf U^{\langle i\rangle}\big)\big)
		-\beta\phi\big(\Psi_{r_ i}\big(\widehat \Delta_{\mathbf U}^{\langle i\rangle}\big)\big)\big)
		+
		\sum_{i=1}^{n_3}\frac{1}{2}\left(\beta\phi\big(\Phi_{r_i}\big(\widehat{\Delta}_\mathbf U^{\langle i\rangle}\big)\big)
		+\beta\phi\big(\Psi_{r_i}\big(\widehat{\Delta}_\mathbf U^{\langle i\rangle}\big)\big)\right)+\frac{\beta\mu}{4}\|\Delta\|_F^2\\
		&=\sum_{i=1}^{n_3}\frac{3\beta}{2}\phi\big(\Phi_{r_i}\big(\widehat \Delta_\mathbf U^{\langle i\rangle}\big)\big)
		-\sum_{i=1}^{n_3}\frac{\beta}{2}\phi\big(\Psi_{r_ i}\big(\widehat \Delta_{\mathbf U}^{\langle i\rangle}\big)\big)+\frac{\beta\mu}{4}\|\Delta\|_F^2.
	\end{aligned}
\end{equation}
Rearranging the above terms yields
\begin{equation}\label{test12}
	\begin{aligned}
		\left(\alpha_1-\frac{3\beta\mu}{4}\right)\|\Delta\|_F^2
		&\leq
		\sum_{i=1}^{n_3}\frac{3\beta}{2}\phi\big(\Phi_{r_i}\big(\widehat \Delta_\mathbf U^{\langle i\rangle}\big)\big)
		-\sum_{i=1}^{n_3}\frac{\beta}{2}\phi\big(\Psi_{r_ i}\big(\widehat \Delta_{\mathbf U}^{\langle i\rangle}\big)\big)\\
		&\leq
		\sum_{i=1}^{n_3}\frac{3\beta}{2}\phi\big(\Phi_{r_i}\big(\widehat \Delta_\mathbf U^{\langle i\rangle}\big)\big).
	\end{aligned}
\end{equation}

\textbf{Step 3}.
It follows from \cite[Lemma 4]{loh15a} that
\begin{equation}\label{Unsdg}
	\begin{aligned}
		\sum_{i=1}^{n_3}\phi\big(\Phi_{r_i}\big(\widehat \Delta_\mathbf U^{\langle i\rangle}\big)\big)
		=\sum_{i=1}^{n_3}\sum_{j=1}^{r_i}g_\lambda(\sigma_j(\widehat \Delta_\mathbf U^{\langle i\rangle}))
		\leq\sum_{i=1}^{n_3}\sum_{j=1}^{r_i}\lambda k_0\sigma_j(\widehat \Delta_\mathbf U^{\langle i\rangle})
		=\sum_{i=1}^{n_3}\lambda k_0 \left\|\Phi_{r_i}\big(\widehat \Delta_\mathbf U^{\langle i\rangle}\big)\right\|_*.
	\end{aligned}
\end{equation}

Combining  (\ref{test12}) and (\ref{Unsdg}), one gets that
\begin{equation}\label{test18}
	\begin{aligned}
		2\left(\alpha_1-\frac{3\beta\mu}{4}\right)\|\Delta\|_F^2
		&\leq
		\sum_{i=1}^{n_3}
		3\beta\lambda k_0\left\|\Phi_{r_i}\left(\widehat{\Delta}_\mathbf U^{\langle i\rangle}\right)\right\|_*
		\leq
		\sum_{i=1}^{n_3}
		3\beta\lambda k_0\sqrt{r_i}\left\|\Phi_{r_i}\left(\widehat{\Delta}_\mathbf U^{\langle i\rangle}\right)\right\|_F\\
		&\leq
		\sum_{i=1}^{n_3}
		3\beta\lambda k_0\sqrt{r_i}\left\|\widehat{\Delta}_\mathbf U^{\langle i\rangle}\right\|_F\\
		&\leq
		3\beta\lambda k_0\sqrt{\sum_{i=1}^{n_3}r_i}\left\|\widehat{\Delta}_\mathbf U\right\|_F
		=
		3\beta\lambda k_0
		\sqrt{\sum_{i=1}^{n_3}r_i}
		\left\|\Delta\right\|_F.
	\end{aligned}
\end{equation}
where the second inequality holds by the fact that
$\|{\bf X}\|_*\leq\sqrt{r}\|{\bf X}\|_F$ for any ${\bf X}$ with rank at most $r$,
the third inequality holds since
$
\|\Phi_{r_i}(\widehat{\Delta}_\mathbf U^{\langle i\rangle})\|_F^2
=\sum_{j=1}^{r_i}(\sigma_j(\widehat{\Delta}_\mathbf U^{\langle i\rangle}))^2
\leq\sum_{j=1}^{\min\{n_1,n_2\}}(\sigma_j(\widehat{\Delta}_\mathbf U^{\langle i\rangle}))^2
=\|\widehat{\Delta}_\mathbf U^{\langle i\rangle}\|_F^2
$,
and the fourth inequality holds by the H{\"o}lder's inequality \cite{MUDHOLKAR1984435}.
Consequently, one can deduce that
\begin{equation}\label{test19}
	\begin{aligned}
		2\left(\alpha_1-\frac{3\beta\mu}{4}\right)\|\Delta\|_F
		&\leq
		3\beta\lambda k_0\sqrt{\sum_{i=1}^{n_3}r_i},
	\end{aligned}
\end{equation}
which together with $\alpha_1>\frac{3\beta\mu}{4}$ implies that
\begin{equation}\label{test26}
	\begin{aligned}
		\|\tilde{\mathcal X}-\mathcal X^*\|_F
		&\leq
		\frac{6\beta\lambda k_0\sqrt{\sum_{i=1}^{n_3}r_i}}{4\alpha_1-3\beta\mu}.
	\end{aligned}
\end{equation}
This completes the proof.

\section*{Appendix F. Locally Lipschitz Continuity of $S_2$ in (\ref{S2Def})}

First we give the locally Lipschitz continuity for the composition function between the singular value vector and a differentiable function.

\begin{lemma}\label{lem23}
	Suppose that $s_2$ satisfies Assumption \ref{assum5}.
	Define $f:\mathbb{R}^m\rightarrow \mathbb{R}$ as $f(\mathbf{x}):=\sum_{i=1}^ms_2(x_i)$.
	For any matrix ${\bf X}\in\mathbb{C}^{n_1\times n_2}$ with the singular value vector
	$\sigma({\bf X}):=(\sigma_1({\bf X}),\sigma_2({\bf X}),\ldots, \sigma_m({\bf X}))^T\in\mathbb{R}^{m}$,
	where $m=\operatorname{min}\{n_1, n_2\}$,
	define
	$(f \circ \sigma) ({\bf X}):\mathbb{C}^{n_1\times n_2}\rightarrow \mathbb{R}$
	as
	$(f \circ \sigma) ({\bf X}):=f(\sigma({\bf X}))$.
	Then  $f \circ \sigma$ is differentiable.
	Moreover, $\nabla( f \circ \sigma)$ is locally Lipschitz continuous,
	i.e.,
	for any given matrix $\tilde{{\bf X}}\in \mathbb{R}^{n_1\times n_2}$,
	there exist constants $\tilde{\delta}>0, L_1>0$ such that
	$$
	\begin{aligned}
		\|\nabla( f \circ \sigma)({\bf X}_1) - \nabla( f \circ \sigma)({\bf X}_2)\|_F
		\leq L_1\|{\bf X}_1-{\bf X}_2\|_F, \ \forall \ {\bf X}_1, {\bf X}_2\in B(\tilde{{\bf X}}, \tilde{\delta}/(2\sqrt{m})).
	\end{aligned}
	$$
\end{lemma}
\textbf{Proof.}
Since $s_2$ is convex, we know that $f(\mathbf{x})$ is convex.
Recall that a matrix is a signed permutation matrix if there is just only one nonzero entry taken $1$ or $-1$ in each row and column.
Since $s_2$ is symmetric, we get that $f({\bf P}\mathbf{x})= f (\mathbf{x})$
for any $\mathbf{x}\in\mathbb{R}$, which implies that $f$ is absolutely symmetric.
Since $s_2$ is differentiable and $f(\mathbf{x})=\sum_{i=1}^ms_2(x_i)$, we can deduce that $f$ is differentiable.
As a consequence,
we  get that $f \circ \sigma$ is differentiable at $\mathbf{X}$ \cite[Proposition 6.2]{Lewis2005a}.
Let the singular value decomposition of ${\bf X}\in\mathbb{C}^{n_1\times n_2}$ be ${\bf X}={\bf U}\Sigma {\bf V}^T$,
where ${\bf U}\in\mathbb{C}^{n_1\times n_1}, {\bf V}\in\mathbb{C}^{n_2\times n_2}$ are
unitary matrices.
It follows from \cite[Proposition 6.2]{Lewis2005a} that
$$
\nabla (f \circ \sigma) ({\bf X}) = {\bf U} \operatorname{\text{Diag}} (\nabla f(\sigma({\bf X}))){\bf V}^T.
$$

By the definition of $f$, we get that
\begin{equation}\label{lip22}
	g(\mathbf{x}):=\nabla f(\mathbf{x})= ( s_2'(x_1),  s_2'(x_2),\ldots, s_2'(x_m))^T\in\mathbb{R}^m,
\end{equation}
where $x_i$ is the $i$-th component of $\mathbf{x}$.
Denote
$$
M(\mathbf{X}):=	\nabla (f \circ \sigma) ({\bf X}) = {\bf U} \operatorname{\text{Diag}} (g(\sigma({\bf X}))){\bf V}^T.
$$
Since $s_2$ is symmetric and differentiable, we know that $s_2'(-x)=-s_2'(x)$.
Therefore, for any signed permutation matrix ${\bf P}\in\mathbb{R}^{m\times m}$, one has
$
g({\bf P}\mathbf{x})={\bf P} g (\mathbf{x}),
$
which shows that $g$ is mixed symmetric at $\mathbf{x}$ \cite[Definition 2.1]{Ding2020}.

Since the derivative $s_2'$  of $s_2$ is  locally Lipschitz continuous, we obtain that for any given $x_i$, there exist $\delta_i, L_0>0$ such that $|s_2'(y_i)-s_2'(z_i)|\leq L_0|y_i-z_i|$, $\forall y_i,z_i\in B(x_i,\delta_i):=\{w: |w-\textcolor{red}{x_i}|\leq \delta_i\}$.
Therefore, for any given $\mathbf{x}=(x_1,x_2,\ldots,x_m)^T\in\mathbb{R}^m$,
$$
\|g(\mathbf{y})-g(\mathbf{z})\|\leq L_0\|\mathbf{y}-\mathbf{z}\|, \ \  \forall \mathbf{y},\mathbf{z}\in B(\mathbf{x},\tilde{\delta}):=\{\mathbf{w}:\|\mathbf{w}-\mathbf{x}\|\leq\tilde{\delta}\},
$$
where $\tilde{\delta}=\sqrt{\sum_{i=1}^m \delta_i^2}$. Consequently, $g$ is locally Lipschitz continuous near $\mathbf{x}$.
For any given matrix $\tilde{{\bf X}}\in \mathbb{C}^{n_1\times n_2}$,
it follows from \cite[Theorem 3.3]{Ding2020} that $M(\cdot)$ is locally Lipschitz continuous on $B(\tilde{{\bf X}}, \tilde{\delta}/(2\sqrt{m}))$ with  modulus 
\begin{equation}\label{DefL1}
L_1=\max \{(2 {L}_0\tilde{\delta} +\tau_0) / \tilde{\delta}, \sqrt{2} {L}_0\}, 
\end{equation}
where
$
\tau_0:=\max _{i, j}\{| s_2'(\sigma_i({\tilde{\bf X}}))- s_2'(\sigma_j({\tilde{\bf X}}))|, | s_2'(\sigma_i({\tilde{\bf X}}))+ s_2'(\sigma_j({\tilde{\bf X}}))|\}.
$
\qed

Next we show the locally Lipschitz continuity of $S_2$  in (\ref{S2Def}), which is stated in the following lemma.

\begin{lemma}\label{lem24}
Define
	$
	S_2(\mathcal X):=\sum_{i=1}^{n_3} \sum_{j=1}^{\min\{n_1, n_2\}} s_2(\sigma_j(\widehat{\mathcal X}_\mathbf U^{\langle i\rangle}))
	$.
	Let ${\mathcal X}\in \mathbb{R}^{n_1\times n_2\times n_3}$ be a given tensor.
	Then there exists a constant $\tilde{\delta}>0$ such that
	$$
	\begin{aligned}
		\|\nabla S_2 (\mathcal Y) - \nabla S_2 (\mathcal Z)\|_F\leq L_1\|\mathcal Y-\mathcal Z\|_F, \ \forall \ \mathcal Y, \mathcal Z\in B(\mathcal X, \tilde{\delta}/(2\sqrt{mn_3}))
	\end{aligned}
	$$
	where $L_1$ is the same constant defined in (\ref{DefL1}).
\end{lemma}
\textbf{Proof.}
Define $f_1:\mathbb{R}^{mn_3}\rightarrow \mathbb{R}$ as $f_1(\mathbf{x})=\sum_{i=1}^{mn_3}s_2(x_i)$, where $m=\min\{n_1,n_2\}$.
For any given tensor ${\mathcal X}\in \mathbb{R}^{n_1\times n_2\times n_3}$.
it follows from Lemma \ref{lem23} that
\begin{equation}\label{lip15}
	\begin{aligned}
		\|\nabla( f_1 \circ \sigma)(\overline{\mathcal Y}) - \nabla( f_1 \circ \sigma)(\overline{\mathcal Z})\|_F
		\leq L_1\|\overline{\mathcal Y} -\overline{\mathcal Z} \|_F, \ \forall \ \overline{\mathcal Y} , \overline{\mathcal Z} \in B(\overline{\mathcal X} , \tilde{\delta}/(2\sqrt{mn_3})).
	\end{aligned}
\end{equation}
Define $P:\mathbb{C}^{n_1\times n_2\times n_3}\rightarrow \mathbb{R}$
as $P(\widehat{\mathcal X}_\mathbf U):=
\sum_{i=1}^{n_3}\sum_{j=1}^{\min\{n_1,n_2\}} s_2(\sigma_j({\widehat{\mathcal X}}_\mathbf U^{\langle i\rangle}))$.
By a similar argument as that in Lemma \ref{lem7}, we get that  $
	\overline{\mathcal A}= \nabla (f_1\circ\sigma)(\overline{\mathcal X})
$ is equivalent to $
\widehat{\mathcal A}_\mathbf U = \nabla P(\widehat{\mathcal X}_\mathbf U),
$
which implies that $\nabla P(\widehat{\mathcal X}_\mathbf U)=\text{fold}_{3}(\nabla( f_1 \circ \sigma)(\overline{\mathcal X}))$.
Note that  $\|\overline{\mathcal Y}-\overline{\mathcal X}\|_F=\|\widehat{\mathcal Y}_\mathbf U-\widehat{\mathcal X}_\mathbf U\|_F$, which yields $\widehat{\mathcal Y}_\mathbf U\in B(\widehat{\mathcal X}_\mathbf U, \tilde{\delta}/(2\sqrt{mn_3}))$.
Similarly, we get that  $\widehat{\mathcal Z}_\mathbf U\in B(\widehat{\mathcal X}_\mathbf U, \tilde{\delta}/(2\sqrt{mn_3}))$.
Consequently, for any tensors $\widehat{\mathcal Y}_\mathbf U, \widehat{\mathcal Z}_\mathbf U\in B(\widehat{\mathcal X}_\mathbf U, \tilde{\delta}/(2\sqrt{mn_3}))$,
we have
\begin{equation}\label{lip11}
	\begin{aligned}
		\|\nabla P(\widehat{\mathcal Y}_\mathbf U) -
		\nabla P(\widehat{\mathcal Z}_\mathbf U)\|_F
		&=\|\text{fold}_{3}(\nabla( f_1 \circ \sigma)(\overline{\mathcal Y})) -
		\text{fold}_{3}(\nabla( f_1 \circ \sigma)(\overline{\mathcal Z}))\|_F\\
		&=\|\nabla( f_1 \circ \sigma)(\overline{\mathcal Y}) -
		\nabla( f_1 \circ \sigma)(\overline{\mathcal Z})\|_F\\
		&\leq L_1\|\overline{\mathcal Y}-\overline{\mathcal Z}\|_F\\
		&=L_1\|\widehat{\mathcal Y}_\mathbf U-\widehat{\mathcal Z}_\mathbf U\|_F,
	\end{aligned}
\end{equation}
where the first inequality holds by (\ref{lip15}).

By the definitions of $S_2$ and $P$,
we get
$S_2 (\mathcal X)=P(\widehat{\mathcal X}_{{\mathbf U}})
=P({\mathbf U}[\mathcal X])$.
By  using a similar discussion in \cite[Exercise 10.7]{R.TyrrellRockafellar1998}, we can easily obtain that
\begin{equation}\label{lip19}
	\nabla S_2 (\mathcal X)={\mathbf U}^T[\nabla P({\mathbf U}[\mathcal X])]
	={\mathbf U}^T[\nabla P(\widehat{\mathcal X}_{\mathbf U})].
\end{equation}
Additionally, we can
easily obtain from $\widehat{\mathcal Y}_\mathbf U, \widehat{\mathcal Z}_\mathbf U\in B(\widehat{\mathcal X}_\mathbf U, \tilde{\delta}/(2\sqrt{mn_3}))$ that $\mathcal Y, \mathcal Z\in B(\mathcal X, \tilde{\delta}/(2\sqrt{mn_3}))$.
By (\ref{lip11}) and (\ref{lip19}), we obtain
\begin{equation}\label{lip12}
	\begin{aligned}
		\|\nabla S_2 (\mathcal Y) -
		\nabla S_2 (\mathcal Z)\|_F
		&=\|{\mathbf U}^T[\nabla P(\widehat{\mathcal Y}_\mathbf U)]-
		{\mathbf U}^T[\nabla P(\widehat{\mathcal Z}_\mathbf U)]\|_F\\
		&=\|{\mathbf U}^T[\nabla P(\widehat{\mathcal Y}_\mathbf U)-
		\nabla P(\widehat{\mathcal Z}_\mathbf U)]\|_F\\
		&=\|\nabla P(\widehat{\mathcal Y}_\mathbf U)-
		\nabla P(\widehat{\mathcal Z}_\mathbf U)\|_F\\
		&\leq L_1\|\widehat{\mathcal Y}_\mathbf U-\widehat{\mathcal Z}_\mathbf U\|_F\\
		&= L_1\|\mathcal Y-\mathcal Z\|_F,
	\end{aligned}
\end{equation}
where the third equality holds since the tensor Frobenius norm is unitarily invariant
\cite[Definition 2.1]{Spectral2021}.
This concludes the proof.
\qed

\section*{Appendix G. Proof of Theorem \ref{ConvRe}}

First,
we give the sufficiently descent property of the objective function $H(\mathcal{X})$ in (\ref{eq3}).

\begin{lemma}\label{DesPr}
	Let $\{\mathcal X^t\}$ be the sequence generated by Algorithm \ref{alg:algorithm1}. Suppose that $\nabla f_{n,\mathcal{Y}}$ is Lipschitz continuous with Lipschitz constant $L$.  Then for any  $\rho>\frac{L}{1-2\xi}$ with $\xi\in(0,\frac{1}{2})$,
	\begin{equation}\label{eq19}
		\begin{aligned}
			H(\mathcal{X}^{t+1})+a\|\mathcal{X}^{t+1}-\mathcal{X}^{t}\|_{F}^{2}
			&\leq
			H(\mathcal{X}^{t}),
		\end{aligned}
	\end{equation}
	where $a:=\frac{1-2\xi}{2}\rho-\frac{L}{2}>0$.
\end{lemma}
{\bf Proof.} By virtue of the definitions (\ref{eq3}) and (\ref{eq4}), we immediately obtain
\begin{equation}\label{eq5}
	\begin{aligned}
		&\ H(\mathcal{X}^{t+1})-Q(\mathcal{X}^{t+1},\mathcal X^t)\\
		=& \ f_{n,\mathcal{Y}}(\mathcal{X}^{t+1})+\beta S_1(\mathcal X^{t+1})-\beta S_2(\mathcal X^{t+1})+\delta_D(\mathcal X^{t+1})
		- f_{n,\mathcal{Y}} (\mathcal X^t)
		-\langle \nabla  f_{n,\mathcal{Y}}(\mathcal X^t),\mathcal X^{t+1}-\mathcal X^t \rangle\\
		& \ -\frac{\rho}{2}\|\mathcal{X}^{t+1}-\mathcal{X}^{t}\|_{F}^{2}
		-\beta S_1(\mathcal X^{t+1})+ \beta S_2(\mathcal X^t)
		+ \beta \langle \nabla S_2(\mathcal X^t),\mathcal X^{t+1}-\mathcal X^t \rangle
		-\delta_D(\mathcal X^{t+1})\\
		=& \ f_{n,\mathcal{Y}}(\mathcal{X}^{t+1})
		- f_{n,\mathcal{Y} }f (\mathcal X^t)
		-\langle \nabla  f_{n,\mathcal{Y} }(\mathcal X^t),\mathcal X^{t+1}-\mathcal X^t \rangle-\frac{\rho}{2}\|\mathcal{X}^{t+1}-\mathcal{X}^{t}\|_{F}^{2}\\
		& \ - \beta (S_2(\mathcal X^{t+1})
		-S_2(\mathcal X^t)
		-\langle \nabla S_2(\mathcal X^t),\mathcal X^{t+1}-\mathcal X^t \rangle).
	\end{aligned}
\end{equation}
Since $\nabla f_{n,\mathcal{Y}}$ is Lipschitz continuous with  Lipschitz constant $L$, we can deduce from \cite[Lemma 5.7]{Amir2017} that
\begin{equation}\label{eq17}
	f_{n,\mathcal{Y}}(\mathcal{X}^{t+1})- f_{n,\mathcal{Y}} (\mathcal X^t)-\langle \nabla f_{n,\mathcal{Y}}(\mathcal X^t),\mathcal X^{t+1}-\mathcal X^t \rangle
	\leq\frac{L}{2}\|\mathcal{X}^{t+1}-\mathcal{X}^{t}\|_{F}^{2}.
\end{equation}
Additionally,
it follows from the convexity of $S_2$ that
\begin{equation}\label{eq21}
	S_2(\mathcal X^{t+1})\geq S_2(\mathcal X^t)+\langle \nabla S_2(\mathcal X^t),\mathcal X^{t+1}-\mathcal X^t \rangle.
\end{equation}
Combining (\ref{eq5}), (\ref{eq17}) and (\ref{eq21}), we get that
\begin{equation}\label{eq8}
	\begin{aligned}
		H(\mathcal{X}^{t+1})-Q(\mathcal{X}^{t+1},\mathcal X^t)
		&\leq\frac{L}{2}\|\mathcal{X}^{t+1}-\mathcal{X}^{t}\|_{F}^{2}
		-\frac{\rho}{2}\|\mathcal{X}^{t+1}-\mathcal{X}^{t}\|_{F}^{2}\\
		&=\frac {L-\rho}{2}\|\mathcal{X}^{t+1}-\mathcal{X}^{t}\|_{F}^{2}.
	\end{aligned}
\end{equation}

Recalling the definition of  $Q(\mathcal{X},\mathcal X^t)$ in (\ref{eq4}), it can be easily verified that $Q(\mathcal{X},\mathcal X^t)$ is convex.
This together with the definition of $\mathcal{W}^{t+1}$ in (\ref{eq9}) leads to
\begin{equation}\label{eq18}
	\begin{aligned}
		Q(\mathcal{X}^{t},\mathcal X^t)
		&\geq
		Q(\mathcal{X}^{t+1},\mathcal X^t)+
		\langle \mathcal{W}^{t+1},\mathcal X^{t}-\mathcal X^{t+1} \rangle\\
		&\geq
		Q(\mathcal{X}^{t+1},\mathcal X^t)-
		\|\mathcal{W}^{t+1}\|_{F}\|\mathcal{X}^{t+1}-\mathcal{X}^{t}\|_{F}\\
		&\geq
		Q(\mathcal{X}^{t+1},\mathcal X^t)-
		\xi\rho\|\mathcal{X}^{t+1}-\mathcal{X}^{t}\|_{F}^2.
	\end{aligned}
\end{equation}
Taking  (\ref{eq18}) together with (\ref{eq8}) yields
\begin{equation}\label{eq7}
	\begin{aligned}
		H(\mathcal{X}^{t+1})
		&\leq
		Q(\mathcal{X}^{t+1},\mathcal X^t)+
		\frac{L-\rho}{2}\|\mathcal{X}^{t+1}-\mathcal{X}^{t}\|_{F}^{2}\\
		&\leq
		Q(\mathcal{X}^{t},\mathcal X^t)+\xi\rho\|\mathcal{X}^{t+1}-\mathcal{X}^{t}\|_{F}^2
		+\frac{L-\rho}{2}\|\mathcal{X}^{t+1}-\mathcal{X}^{t}\|_{F}^{2}\\
		&=
		Q(\mathcal{X}^{t},\mathcal X^t)+\left(\frac{L}{2}-\frac{1-2\xi}{2}\rho\right)\|\mathcal{X}^{t+1}-\mathcal{X}^{t}\|_{F}^2.
	\end{aligned}
\end{equation}
Therefore, we get that
\begin{equation}\label{eq19}
	\begin{aligned}
		H(\mathcal{X}^{t+1})+\left(\frac{1-2\xi}{2}\rho-\frac{L}{2}\right)\|\mathcal{X}^{t+1}-\mathcal{X}^{t}\|_{F}^{2}
		&\leq
		Q(\mathcal{X}^{t},\mathcal X^t)=H(\mathcal{X}^{t}),
	\end{aligned}
\end{equation}
where $\frac{1-2\xi}{2}\rho-\frac{L}{2}> 0$. This completes the proof.
\qed

Next, we give the relative error property of the iterations in Algorithm \ref{alg:algorithm1}, which is stated in the following lemma.

\begin{lemma}\label{SubBoun}
	Let $\{\mathcal X^t\}$ be the sequence generated by Algorithm \ref{alg:algorithm1}. Suppose that $\nabla f_{n,\mathcal{Y}}$ and $s_2'$ are Lipschitz continuous and locally Lipschitz continuous with Lipschitz constants $L$ and $L_0$, respectively. Then
	there exist $\mathcal{N}^{t+1}\in\partial H(\mathcal{X}^{t+1})$ and a positive integer $K$ such that
\begin{equation*}\label{SybUpubud}
	\|\mathcal{N}^{t+1}\|_F\leq \nu\|\mathcal{X}^{t+1}-\mathcal{X}^{t}\|_F, \  \forall \ t\geq K,
\end{equation*}
	where $\nu:=L+\beta L_1+(\xi+1)\rho>0$ and $L_1$ is defined in (\ref{DefL1}).
\end{lemma}
\textbf{Proof.}
By (\ref{eq9}), we obtain that there exists $
\mathcal{Y}^{t+1}\in\partial [\beta S_1(\mathcal X^{t+1})+\delta_D(\mathcal X^{t+1})]
$ such that
$$
\mathcal{W}^{t+1}=\nabla f_{n,\mathcal{Y}}(\mathcal X^t)+\rho(\mathcal{X}^{t+1}-\mathcal{X}^{t})-\beta\nabla S_2(\mathcal X^t)+\mathcal{Y}^{t+1}
\ \textup{and} \  \|\mathcal{W}^{t+1}\|_F\leq
\xi\rho\|\mathcal{X}^{t+1}-\mathcal{X}^{t}\|_{F}.
$$
Denote
$$
\mathcal{N}^{t+1}:=\nabla f_{n,\mathcal{Y}}(\mathcal X^{t+1})-
\beta\nabla S_2(\mathcal X^{t+1})+\mathcal{Y}^{t+1}.
$$
Note that
$$
\partial H(\mathcal{X})=\nabla f_{n,\mathcal{Y}}(\mathcal{X})-\beta \nabla S_2(\mathcal X)+\partial[\beta S_1(\mathcal X)+\delta_D(\mathcal X)].
$$
Consequently, we deduce that $\mathcal{N}^{t+1}\in\partial H(\mathcal{X}^{t+1})$.

By (\ref{eq19}), we know that $\|\mathcal{X}^{t+1}-\mathcal{X}^{t}\|_{F}$ tends to $0$ as $t$ tends to infinity, which implies that
there exists a constant ${\bar \delta}>0$ and positive integer $K$ such that $\|\mathcal{X}^{t+1}-\mathcal{X}^{t}\|_{F}\leq {\bar \delta}$, for any $t\geq K$.
Note that  the sequence $\{\mathcal X^t\}$ is an approximate solution of
$Q(\mathcal{X},\mathcal X^t)$ in (\ref{eq4}) at each iteration and $\delta_D(\cdot)$ is the indicator function of $D$. Therefore, $\mathcal{X}^t\in D$,
which implies that $\{\mathcal X^t\}$ is bounded.
Assume that $\tilde{\mathcal X}$ is a cluster point of $\{\mathcal X^{t}\}$,
there exists $\delta_0>0$ such that
$\mathcal X^{t}\in B(\tilde{\mathcal X}, \delta_0)$ holds for $t\geq K$.
We further obtain that
$$
\begin{aligned}
	\|\mathcal{X}^{t+1}-\tilde{\mathcal X}\|_{F}
	&=\|\mathcal{X}^{t+1}-\mathcal{X}^{t}+\mathcal{X}^{t}-\tilde{\mathcal X}\|_{F}\\
	&\leq \|\mathcal{X}^{t+1}-\mathcal{X}^{t}\|_{F}+\|\mathcal{X}^{t}-\tilde{\mathcal X}\|_{F}\\
	&\leq {\bar \delta}+\delta_0.
\end{aligned}
$$
Denote $\tilde{\delta}:={\bar \delta}+\delta_0$.
Then we have $\mathcal X^{t+1}\in B(\tilde{\mathcal X}, \tilde{\delta})$.
Note that  $\|\mathcal{X}^{t}-\tilde{\mathcal X}\|_{F}\leq \delta_0<\tilde{\delta}$, which
immediately yields
$\mathcal X^{t}, \mathcal X^{t+1}\in B(\tilde{\mathcal X}, \tilde{\delta})$.
It follows from Lemma \ref{lem24}  that
\begin{equation}\label{SSK1}
\|\nabla S_2 (\mathcal X^{t+1}) -
\nabla S_2 (\mathcal X^{t})\|_F \leq  L_1\|\mathcal X^{t+1}-\mathcal X^{t}\|_F.
\end{equation}
for any $t\geq K$.

Notice that
$$
\mathcal{N}^{t+1}=\nabla f_{n,\mathcal{Y}}(\mathcal X^{t+1})-
\beta\nabla S_2(\mathcal X^{t+1})+\mathcal{W}^{t+1}-\nabla f_{n,\mathcal{Y}}(\mathcal X^t)-\rho(\mathcal{X}^{t+1}-\mathcal{X}^{t})+\beta\nabla S_2(\mathcal X^t).
$$
Then we get that
\begin{equation}
	\begin{aligned}
		\|\mathcal{N}^{t+1}\|_F&\leq\|\nabla f_{n,\mathcal{Y}}(\mathcal X^{t+1})-\nabla f_{n,\mathcal{Y}}(\mathcal X^t)\|_F +\|\mathcal{W}^{t+1}\|_F+\rho\|\mathcal{X}^{t+1}-\mathcal{X}^{t}\|_F\\
		&~~~~~+\beta\|\nabla S_2(\mathcal X^{t+1})-\nabla S_2(\mathcal X^t)\|_F\\
		&\leq L\|\mathcal{X}^{t+1}-\mathcal{X}^{t}\|_F	+(\xi+1)\rho\|\mathcal{X}^{t+1}-\mathcal{X}^{t}\|_{F}+\beta L_1\|\mathcal{X}^{t+1}-\mathcal{X}^{t}\|_{F} \\
		&=(L+\beta L_1+(\xi+1)\rho)\|\mathcal{X}^{t+1}-\mathcal{X}^{t}\|_{F},
	\end{aligned}
\end{equation}
where the second inequality holds by (\ref{SSK1}) and the  Lipschitz continuity of $\nabla f_{n,\mathcal{Y}}$.
This completes the proof. \qed

Now we return to prove Theorem \ref{ConvRe} in detail. By the proof of Lemma \ref{SubBoun}, we know that
 $\{\mathcal X^t\}$ is bounded.
Consequently, there exists a subsequence
$\{\mathcal X^{t_i}\}$ such that $\mathcal X^{t_i}$ converges to $\tilde{\mathcal X}$ as $i$ tends to $\infty$, where $\tilde{\mathcal X}$ is a cluster point of the sequence $\{\mathcal X^t\}$.
Note that $D$ is a closed set and $\mathcal X^{t_i}\in D$, then $\tilde{\mathcal X}\in D$.
Therefore, $\delta_D({\mathcal X^{t_i}})$ converges to $\delta_D(\tilde{\mathcal X})$ as $i$ tends to $\infty$.
Since $f_{n,\mathcal{Y}} $ and $G_\lambda$ are continuous, we deduce that
$f_{n,\mathcal{Y}}(\mathcal X^{t_i}) +\beta G_\lambda(\mathcal X^{t_i})$ converges to
$f_{n,\mathcal{Y}}(\tilde{\mathcal X}) +\beta G_\lambda(\tilde{\mathcal X})$ as $i $ tends to infinity, which implies that
$H(\mathcal X^{t_i})$ converges to $ H(\tilde{\mathcal X})$ as $i$ tends to $\infty$.

Since $D$ is a closed convex set, we know that $\delta_D(\cdot)$ is closed \cite[Proposition 2.3]{Amir2017}, which implies that $\delta_D(\cdot)$ is lower semicontinouns \cite[Theorem 2.6]{Amir2017}. Since $f_{n,\mathcal{Y}} $ and $G_\lambda$ are continuous, we obtain that $H(\mathcal X)$ is lower semicontinouns.
It is known that $\delta_D(\mathcal X)$ is semialgebraic since $D$ is a semialgebraic set \cite{Bolte2013}.
Then  $\delta_D(\mathcal X)$ is a KL function  \cite[Theorem 3]{Bolte2013}.
Since $g_\lambda$ is a KL function, $G_\lambda(\mathcal{X})$ is a KL function \cite{qiu2021nonlocal, Bolte2013}.
Note that $f_{n,\mathcal{Y}}$ is a  KL function. As a result,
$H(\mathcal X)=f_{n,\mathcal{Y}}(\mathcal X)+\beta G_\lambda(\mathcal X)+\delta_D(\mathcal X)$ is a KL function.
According to \cite[Theorem 2.9]{Attouch_2013}, we
conclude that the sequence $\{\mathcal X^t\}$ converges to $\tilde{\mathcal X}$ as $t$ goes to infinity, and $\tilde{\mathcal X}$ is a stationary point of $H$.

\section*{Appendix H. Proof of Theorem \ref{thm3}}

First, we give a lemma about the sequence $\{\mathcal X^{t}\}$ generated by Algorithm \ref{alg:algorithm1}.

\begin{lemma}\label{lem30}
Suppose that the assumptions in Theorem \ref{ConvRe} hold and  $H({\mathcal X})$ defined in (\ref{eq3}) satisfies the KL property at
$\tilde{\mathcal X}$ with an exponent $\alpha\in[0,1)$.
Then the following statements hold:\par
(i) There exist constants $\rho_1, \mu_1>0$ and a KL exponent $\alpha\in(0,1)$, such that
\begin{equation}\label{test116}
\sum\limits_{j=t}^{\infty}\|\mathcal X^{j+1}-\mathcal X^{j}\|_{F}\leq \| \mathcal X^{t}-\mathcal {X}^{t-1}\|_F+ \frac{\mu_1}{\rho_1(1-\alpha)}(H(\mathcal X^{t})-H(\tilde{\mathcal X}))^{1-\alpha}.
\end{equation}

(ii) For any positive integer $t$, the following inequality holds:
\begin{equation}\label{test401}
\|\mathcal X^{t}-\tilde{\mathcal X}\|_{F}\leq \sum\limits_{j=t}^{\infty}\|\mathcal X^{j+1}-\mathcal X^{j}\|_{F}.
\end{equation}
 \end{lemma}
\textbf{Proof.}
(i)
Define a function $p:\mathbb{R}\rightarrow\mathbb{R}_+$ as
$$
p(s):=\frac{\mu_1}{1-\alpha}(s-H(\tilde{\mathcal X}))^{1-\alpha},\ \forall  s\geq H(\tilde{\mathcal X}),
$$
where $\mu_1>0$ is a constant and $\alpha\in(0,1)$.
It can be easily seen that
$p$ is a concave function,
whose derivative function is given by
$p'(s)=\frac{\mu_1}{(s-H(\tilde{\mathcal X}))^{\alpha}}$ for any $s>H(\tilde{\mathcal X}).$
From the concavity of $p(s)$, one has
\begin{equation}\label{test122}
\begin{aligned}
p(H(\mathcal X^{t}))-p(H(\mathcal X^{t+1}))
&\geq p{'}(H(\mathcal X^{t}))(H(\mathcal X^{t})-H(\mathcal X^{t+1}))\\
&=\frac{\mu_1}{(H(\mathcal X^{t})-H(\tilde{\mathcal X}))^{\alpha}}(H(\mathcal X^{t})-H(\mathcal X^{t+1})).
\end{aligned}
\end{equation}
Note that $H({\mathcal X})$ satisfies the KL property at
$\tilde{\mathcal X}$ with an exponent $\alpha\in[0,1)$.
Then the KL inequality (\ref{test316KL}) yields
\begin{align}\label{test126}
(H(\mathcal X^{t})-H(\tilde{\mathcal X}))^{\alpha}\leq\mu_1 (1-\alpha)\mathop{\mathrm{dist}}(0,\partial H(\mathcal X^{t}))\leq \mu_1 \nu\| \mathcal X^{t}-\mathcal {X}^{t-1}\|_F,  
\end{align}
where the last inequality follows from
$\alpha\in(0,1)$
and Lemma \ref{SubBoun}.
Here $\nu$ is defined in Lemma \ref{SubBoun}.
This taken together with (\ref{test122}) and Lemma \ref{DesPr} gives
\begin{equation*}\label{test123}
\begin{aligned}
p(H(\mathcal X^{t}))-p(H(\mathcal X^{t+1}))
\geq\frac{a \| \mathcal X^{t+1}-\mathcal {X}^{t}\|_F^2}{\nu \| \mathcal X^{t}-\mathcal {X}^{t-1}\|_F},
\end{aligned}
\end{equation*}
where $a$ is defined in Lemma \ref{DesPr}.
Furthermore, we can obtain that
\[
\begin{split}
2\| \mathcal X^{t+1}-\mathcal {X}^{t}\|_F
&\leq 2\Big(\frac{\nu}{a}\Big)^{\frac{1}{2}}\Big(p(H(\mathcal X^{t}))-p(H(\mathcal X^{t+1}))\Big)^{\frac{1}{2}}\| \mathcal X^{t}-\mathcal {X}^{t-1}\|_F^{\frac{1}{2}} \\
&\leq \frac{\nu}{a}\Big(p(H(\mathcal X^{t}))-p(H(\mathcal X^{t+1}))\Big)+\| \mathcal X^{t}-\mathcal {X}^{t-1}\|_F,
\end{split}
\]
where the second inequality follows from the fact that $ 2xy\leq x^2+y^2$.
Summing the aforementioned inequality from $t$ to infinity yields
$$
2\sum\limits_{j=t}^{\infty}\| \mathcal X^{j+1}-\mathcal {X}^{j}\|_F
\leq
\sum\limits_{j=t}^{\infty}\frac{\nu}{a}\Big(p(H(\mathcal X^{j}))-p(H(\mathcal X^{j+1}))\Big)+\sum\limits_{j=t}^{\infty}\| \mathcal X^{j}-\mathcal {X}^{j-1}\|_F
.
$$
Consequently,
\[
\begin{aligned}
\sum\limits_{j=t}^{\infty}{\big\|\mathcal X^{j+1}-\mathcal X^{j}\big\|}_{F}
&\leq \| \mathcal X^{t}-\mathcal {X}^{t-1}\|_F+\frac{\nu}{a}\sum\limits_{j=t}^{\infty}(p(H(\mathcal X^{j}))-p(H(\mathcal X^{j+1})))\\
&=\| \mathcal X^{t}-\mathcal {X}^{t-1}\|_F+\frac{\nu}{a}\Big(p(H(\mathcal X^{t}))-\lim_{j\rightarrow \infty}p(H(\mathcal X^{j}))\Big)\\
&\leq\| \mathcal X^{t}-\mathcal {X}^{t-1}\|_F+\frac{\nu}{a}p(H(\mathcal X^{t}))\\
&=\| \mathcal X^{t}-\mathcal {X}^{t-1}\|_F+\frac{\nu\mu_1}{a(1-\alpha)}(H(\mathcal X^{t})-H(\tilde{\mathcal X}))^{1-\alpha} \\
&= \| \mathcal X^{t}-\mathcal {X}^{t-1}\|_F+\frac{\mu_1}{\rho_1(1-\alpha)}(H(\mathcal X^{t})-H(\tilde{\mathcal X}))^{1-\alpha},
\end{aligned}
\]
where  $\rho_1:=\frac{a}{\nu}$.

(ii) Define $M_{t}(\mathcal{X}):=\mathcal{X}^{t+1}-\mathcal{X}^{t}$.
It follows from Theorem \ref{ConvRe}
that
$\{\mathcal{X}^{t+1}\}_{t\in \mathbb{N}}$ is bounded
and
$\|M_{t}(\mathcal{X})\|_{F}=\|\mathcal{X}^{t+1}-\mathcal{X}^{t}\|_{F}<\infty$ for any $t=1,2,\ldots$
Additionally,
we denote a bounded measurable set by $\mathcal{G}$, which
obeys $\{\mathcal{X}^{t+1}\}_{t\in \mathbb{N}}\subset \mathcal{G}$.
One can obtain that $M_{t}(\mathcal{X})$ is a measurable function on $\mathcal{G}$.
Note that,
the sequence $\{\mathcal{X}^{t}\}_{t\in \mathbb{N}}$ converges to $\tilde{\mathcal X}$, which implies that $\sum_{t=1}^{n}M_{t}(\mathcal{X})=\sum_{t=1}^{n}\left(\mathcal{X}^{t+1}-\mathcal{X}^{t}\right)=\mathcal{X}^{n+1}-\mathcal{X}^{1}$ converges to $\tilde{\mathcal X}-\mathcal{X}^{1}$ as $n$ tends to infinity.
Similar to the argument in \cite[Lemma 1(iv)]{Xia2023},
we can easily prove
$
\|\mathcal X^{t}-\tilde{\mathcal X}\|_{F}\leq \sum_{j=t}^{\infty}\|\mathcal{X}^{j+1}-\mathcal{X}^{j}\|_{F}.
$
This completes the proof.
\qed

We now provide the details of the proof of Theorem \ref{thm3}, which follows a similar argument of \cite{attouch2009convergence, Xia2023}.

(i)
We must have $H(\mathcal X^{t_0})=H(\tilde{\mathcal X})$ for some $t_0$ when $\alpha=0$.
Otherwise, for sufficiently large $t$, one has $H(\mathcal X^{t})>H(\tilde{\mathcal X})$.
Apply the KL inequality to obtain $\mu_1 \mathop{\mathrm{dist}}(0,\partial H(\mathcal X^{t}))\geq 1$ for all $t$,
it is impossible
since $\mathcal X^{t}\rightarrow \tilde{\mathcal X}$ and $0\in\partial H(\tilde{\mathcal X})$.
Then there exists some $t_0$ such that $H(\mathcal X^{t_0})=H(\tilde{\mathcal X})$.
Note that $H$ decreases monotonically, then $\mathcal X^{t}=\mathcal X^{t_0}=\tilde{\mathcal X}$ holds true for all $t>t_0$.

(ii)
Let $\Delta_t:=\sum_{j=t}^{\infty}\|\mathcal X^{j+1}-\mathcal X^{j}\|_{F}$.
According to (\ref{test401}),
it can be directly seen that $\Delta_t \geq \|\mathcal X^{t}-\tilde{\mathcal X}\|_{F}$.
Then, (\ref{test116}) implies that
\begin{equation}\label{test111}
\begin{split}
\Delta_t=\sum\limits_{j=t}^{\infty}\|\mathcal X^{j+1}-\mathcal X^{j}\|_{F}
&\leq \| \mathcal X^{t}-\mathcal {X}^{t-1}\|_F+\frac{\mu_1}{\rho_1(1-\alpha)}(H(\mathcal X^{t})-H(\tilde{\mathcal X}))^{1-\alpha}\\
&\leq \| \mathcal X^{t}-\mathcal {X}^{t-1}\|_F+\frac{\mu_1}{\rho_1(1-\alpha)}\Big(\mu_1 \nu \|\mathcal X^{t}-\mathcal X^{t-1}\|_{F}\Big)^{\frac{1-\alpha}{\alpha}}\\
&=(\Delta_{t-1}-\Delta_{t})+a_1(\Delta_{t-1}-\Delta_{t})^{\frac{1-\alpha}{\alpha}},
\end{split}
\end{equation}
where the second inequality follows from (\ref{test126})
and the last equality holds according to letting $a_1:=\frac{\mu_1}{\rho_1(1-\alpha)}(\mu_1 \nu)^{\frac{1-\alpha}{\alpha}}$.

If $0<\alpha\leq\frac{1}{2}$, then $\frac{1-\alpha}{\alpha}\geq 1$.
Similar to the proof in \cite[Theorem 2(ii)]{Xia2023},
by  (\ref{test111}),
we can easily show that $\Delta_t\leq \frac{a_2}{a_2+1}\Delta_{t-1}$ holds, where  $a_2:=a_1+1$.
Denote $w:=\Delta_0$ and $ \vartheta:=\frac{a_2}{a_2+1}\in(0,1)$.
It follows from Lemma \ref{lem30}(ii) that
$$
\|\mathcal X^t-\tilde{\mathcal X}\|_F \leq \Delta_t\leq \vartheta\Delta_{t-1}\leq\cdots\leq
\vartheta^t\Delta_{0}=w\vartheta^t.
$$

(iii) If $\frac{1}{2}<\alpha<1$, then $0<\frac{1-\alpha}{\alpha}<1$.
Let $b_t:=\sum_{j=0}^{t}\|\mathcal X^{j+1}-\mathcal X^{j}\|_{F}$.
By setting $t=1$ in (\ref{test116}), we can obtain that
\begin{equation}\label{cov7}
\begin{split}
\sum\limits_{j=0}^{\infty}{\|\mathcal X^{j+1}-\mathcal X^{j}\|}_{F}
&=\sum\limits_{j=1}^{\infty}{\|\mathcal X^{j+1}-\mathcal X^{j}\|}_{F}+\| \mathcal X^{1}-\mathcal {X}^{0}\|_F \\
&\leq 2\| \mathcal X^{1}-\mathcal {X}^{0}\|_F+ \frac{\mu_1}{\rho_1(1-\alpha)}(H(\mathcal X^{1})-H(\tilde{\mathcal X}))^{1-\alpha}<+\infty,
\end{split}
\end{equation}
which implies that
${\lim\limits_{t\to\infty}b_t=\sum_{j=0}^{\infty}{\|\mathcal X^{j+1}-\mathcal X^{j}\|}_{F}}$ exists.
Observe that
$$
\Delta_t=\sum_{j=t}^{\infty}\|\mathcal X^{j+1}-\mathcal X^{j}\|_{F}
=\sum_{j=0}^{\infty}\|\mathcal X^{j+1}-\mathcal X^{j}\|_{F}-b_{t-1}.
$$
Hence, we have
\begin{equation}\label{DelKlim}
\lim\limits_{t\to\infty}\Delta_t=\sum_{j=0}^{\infty}{\|\mathcal X^{j+1}-\mathcal X^{j}\|}_{F}-\lim\limits_{t\to\infty}b_{t-1}=0.
\end{equation}
Consequently, there exists some positive integer $T_0$ such that for any $t>T_0$,
$$
(\Delta_{t-1}-\Delta_{t})\leq (\Delta_{t-1}-\Delta_{t})^{\frac{1-\alpha}{\alpha}}.
$$
This together with (\ref{test111}) leads to
\begin{equation}\label{cov1}
\Delta_t^\frac{\alpha}{1-\alpha} \leq a_3(\Delta_{t-1}-\Delta_{t}),
\end{equation}
where $a_3=(1+a_1)^\frac{\alpha}{1-\alpha}$.

Given a constant $c$ with $c\in(1,\infty)$.
Define $\gamma(s):=s^{-\frac{\alpha}{1-\alpha}}, s>0$.

{\bf Case I}. Assume that $\gamma(\Delta_t)\leq c\gamma(\Delta_{t-1})$,
by (\ref{cov1}),
we obtain
\begin{equation}\label{cov2}
\begin{split}
a_3^{-1}
&\leq \Delta_t^{-\frac{\alpha}{1-\alpha}}(\Delta_{t-1}-\Delta_{t})=\gamma(\Delta_t)(\Delta_{t-1}-\Delta_{t})\\
&\leq\int_{\Delta_{t}}^{\Delta_{t-1}} c\gamma(\Delta_{t-1}) ds\leq \int_{\Delta_{t}}^{\Delta_{t-1}} c\gamma(s) ds\\
&=\frac{c(1-\alpha)}{1-2\alpha}\left(\Delta_{t-1}^{\frac{1-2\alpha}{1-\alpha}}-
\Delta_{t}^{\frac{1-2\alpha}{1-\alpha}}\right).
\end{split}
\end{equation}
Let $\tau:=\frac{1-2\alpha}{1-\alpha} \text{ and } e_1:=a_3^{-1}\frac{2\alpha-1}{c(1-\alpha)}$.
Note that $\tau<0$.
Observe from (\ref{cov2}) that
\begin{equation}\label{cov3}
\Delta_{t}^{\tau}-\Delta_{t-1}^{\tau}\geq e_1 > 0.
\end{equation}

{\bf Case II}. Assume that $\gamma(\Delta_t)> c\gamma(\Delta_{t-1})$, i.e.,
$
\Delta_t^{-\frac{\alpha}{1-\alpha}}>c\Delta_{t-1}^{-\frac{\alpha}{1-\alpha}},
$
which is equivalent to
$
{\Delta_t}< c^{-\frac{1-\alpha}{\alpha}}\Delta_{t-1}.
$
By the definition of   $\tau=\frac{1-2\alpha}{1-\alpha}<0$, we get that
$$
{\Delta_t^\tau}> (c^{-\frac{1-\alpha}{\alpha}})^\tau\Delta_{t-1}^\tau,
$$
which implies that
\begin{equation}\label{cov4}
{\Delta_t^\tau}-\Delta_{t-1}^\tau> ((c^{-\frac{1-\alpha}{\alpha}})^\tau-1)\Delta_{t-1}^\tau.
\end{equation}
Notice that $c^{-\frac{1-\alpha}{\alpha}}\in(0,1)$. Therefore,
we obtain $(c^{-\frac{1-\alpha}{\alpha}})^{\tau}-1>0$.

It follows from (\ref{DelKlim}) that there exist a constant $a_4>0$ and a positive integer $T_1$ such that $\Delta_{t-1}\leq a_4$
for any $t>T_1$.
Consequently, we can deduce that $((c^{-\frac{1-\alpha}{\alpha}})^\tau-1)\Delta_{t-1}^\tau>((c^{-\frac{1-\alpha}{\alpha}})^\tau-1)a_4^{\tau}$.
This taken together with (\ref{cov4}) leads to
\begin{equation}\label{cov5}
{\Delta_t^\tau}-\Delta_{t-1}^\tau > e_2>0,
\end{equation}
where $e_2:=((c^{-\frac{1-\alpha}{\alpha}})^\tau-1)a_4^{\tau}$.

Let $\tilde{e}:=\min\{e_1,e_2\}$.
Combining (\ref{cov3}) with (\ref{cov5}), we obtain
$
{\Delta_t^\tau}-\Delta_{t-1}^\tau \geq \tilde{e}.
$
By a similar argument as \cite[Theorem 2(iii)]{Xia2023}, we get that
$
\Delta_{t}^{\tau}\geq \frac{\tilde e t}{2},
$
where $t>2\max\{T_0,T_1\}$.
Then we can conclude that
$$
{\|\mathcal X^{t}-\tilde{\mathcal X}\|}_{F}\leq\Delta_t \leq w t^{\frac{1}{\tau}} = w t^{-\frac{1-\alpha}{2\alpha-1}},
$$
where $w:=(\frac{\tilde e }{2})^{-\frac{1-\alpha}{2\alpha-1}}$.

\bibliographystyle{abbrv}

\bibliography{Refalg2}

\end{document}